\documentclass{article}

\PassOptionsToPackage{numbers, compress}{natbib}

 \usepackage[preprint]{neurips_2026}

\usepackage[utf8]{inputenc} %
\usepackage[T1]{fontenc}    %
\usepackage{hyperref}       %
\usepackage{url}            %
\usepackage{booktabs}       %
\usepackage{amsfonts}       %
\usepackage{nicefrac}       %
\usepackage{microtype}      %
\usepackage{xcolor}         %

\usepackage{mathtools}
\usepackage{amsfonts,amsmath,amsthm}
\usepackage[capitalise,noabbrev]{cleveref}
\AddToHook{cmd/appendix/before}{\crefalias{section}{appendix}}
\AddToHook{cmd/appendix/before}{\crefalias{subsection}{appendix}}

\usepackage{tabularx}
\usepackage{tabularray}
\usepackage{multicol}
\usepackage{multirow}
\usepackage{makecell}
\usepackage{subcaption}
\usepackage{tcolorbox}
\usepackage{wrapfig}
\AtBeginEnvironment{tcolorbox}{\small}

\usepackage{enumitem}
\usepackage{adjustbox}
\newcommand{\SelTableStart}{%
  \begingroup
  \scriptsize
  \setlength{\tabcolsep}{3pt}%
  \renewcommand{\arraystretch}{1.15}%
  \begin{adjustbox}{max width=\linewidth}%
}
\newcommand{\SelTableEnd}{%
  \end{adjustbox}%
  \endgroup
}

\usepackage{cite}

\usepackage{graphicx}

\newcommand{\OSS}{\text{OSS-120B}}

\newtheorem{example}{Example}

\newtheorem{result}{Finding}

\usepackage{tikz}
\usepackage{pifont} %

\title{Preference Reasoning under Indeterminacy\\ in Large Language Models
}

\author{%
\textbf{Hadi Hosseini} \\ 
Penn State University, USA\\ 
\texttt{hadi@psu.edu}
\and 
\textbf{Samarth Khanna}\footnotemark[1] \\
Penn State University, USA\\ 
\texttt{samarth.khanna@psu.edu}
\and
\textbf{Xiyuan Wang} \\
Penn State University, USA\\ 
\texttt{xjw5253@psu.edu}
}

\begin{document}

\maketitle
\footnotetext[1]{Corresponding author.}

\begin{abstract}
  As large language models evolve into decision-making agents, the ability to reason over preferences becomes fundamental to alignment, coordination, and collective intelligence. Yet, unlike standard benchmarks, real-world preference reasoning is inherently \textit{indeterminate}: information may be incomplete, and valid solutions may not exist. We argue that indeterminacy, rather than correctness alone, is a central challenge for AI reasoning. We formalize this challenge along two axes, (i) \textit{epistemic indeterminacy}, arising from incomplete, partial, or expressive preferences, and (ii) \textit{structural indeterminacy}, arising from the non-existence of solutions under standard social choice concepts. Across a hierarchy of tasks, we show that state-of-the-art language models systematically fail to distinguish between determined and undetermined instances, exhibiting miscalibrated reasoning even in verification settings.
\end{abstract}

\section{Introduction}

Large Language Models (LLMs) have evolved from statistical language generators into systems capable of executing increasingly complex reasoning tasks, including logical inference and algorithmic problem solving.
This progression signals a shift toward models that operate over structured representations rather than purely linguistic patterns, enabling reasoning over preferences.

Preference reasoning is a foundational component of modern AI systems, spanning across alignment, fine-tuning, and recommender systems. 
In agentic settings, LLMs are tasked with acting on behalf of users, requiring either \textit{implicit} or \textit{explicit} inference, comparison, and aggregation of potentially conflicting preferences when interacting with environments and other agents. At scale, these mechanisms extend to collective decision-making, where models are used to elicit and aggregate preferences into social judgments \citep{fish2023generative,Conitzer+24}, as in platforms such as Pol.is, Remesh, and deliberative frameworks like the Habermas Machine \citep{doi:10.1126/science.adq2852}.

Despite these advances, current evaluation paradigms for LLM reasoning rely primarily (and often exclusively) on \textit{closed-world benchmarks} in which a ground-truth solution is often assumed to exist. However, real-world decision-making is inherently more complex, often characterized by undetermined scenarios in which preferences are incomplete or structural constraints preclude the existence of a valid solution under a given objective (aka a solution concept).
This motivates a shift in assessing whether a model can distinguish between what is determined (entailed by axioms or existing as a solution) and what is undetermined (not entailed or non-existent).
With this lens, a central axis of evaluation extends beyond correctness to whether models can identify the boundary between determinacy and indeterminacy: \textit{Can advanced reasoning models identify whether a preference query is not answerable or whether a solution is infeasible under given specification?}

We study preference reasoning under two distinct forms of \textit{indeterminacy}:
(i) \textbf{epistemic indeterminacy}, arising from incomplete preference information, and
(ii) \textbf{structural indeterminacy}, arising from the interaction between preference structure (e.g., ties) and solution concepts, which may render objectives infeasible even when preferences are fully specified.\footnote{We avoid using the term ``decidable/undecidable'' since decidability is often used to imply computational tractability. In this paper, we are concerned about epistemological determinacy when efficient algorithms exist.}
This view is formally aligned with the Open-World Assumption (OWA) and multi-valued semantic frameworks such as Kleene’s three-valued logic ($K3$), in which propositions may take an explicit “unknown” or “indeterminate” truth value \citep{kleene1952introduction}.

We ground our evaluation in classical economic problems over ranked preferences, which provide a principled testbed for reasoning under constraints. These settings require models to interpret incomplete preferences and ties, resolve conflicts across agents, and satisfy global objectives defined by axiomatic solution concepts such as stability and welfare. Crucially, these concepts distinguish between feasible and infeasible outcomes, making this domain ideal for evaluating not only reasoning accuracy but also indeterminacy. 
Consequently, we transform indeterminacy into an observable and measurable failure mode of LLMs. 

\begin{figure}
    \centering
    \includegraphics[width=\linewidth]{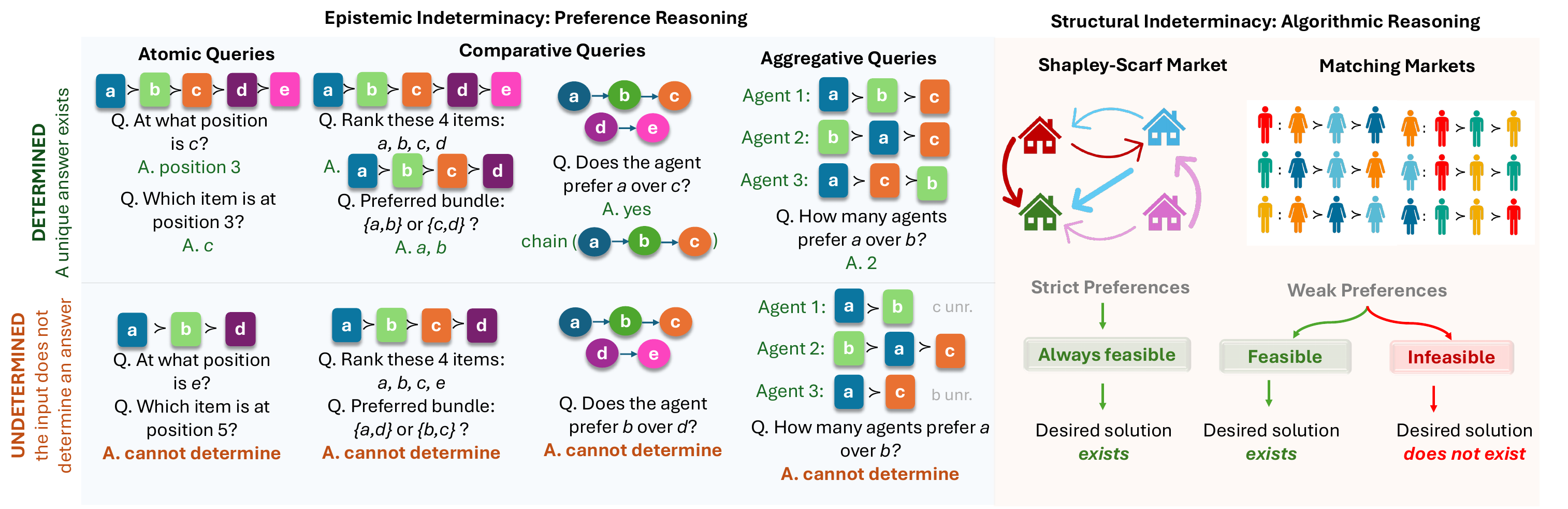}
    \caption{Overview of our taxonomy of evaluating LLMs' reasoning with indeterminacy.
    }
    \label{fig:pref_questions_teaser}
\end{figure}

\subsection{Main Results}
We study preference-based reasoning in LLMs and introduce a formal taxonomy of determined vs. undetermined reasoning across increasing levels of complexity: (i) \textit{atomic queries}, requiring retrieval from a single preference (e.g., what is the rank of alternative `$a$'?); (ii) \textit{comparative queries}, requiring entailment or refutation of relations (e.g., whether `$a$' is preferred to `$b$'); (iii) \textit{aggregative queries}, requiring aggregation across multiple preferences (e.g., how many agents prefer `$a$' over `$b$'?); and (iv) \textit{structural (algorithmic) queries}, requiring the construction of outcomes satisfying social choice solution concepts (e.g., finding a matching solution in the \textit{core}).

We leverage the nuanced interaction between \textit{preference expressivity} (e.g., partial orders, ties, and incomplete lists) and solution concepts from social choice
to generate a spectrum of complex reasoning tasks within an axiomatic framework of indeterminacy. Crucially, all tasks---both query resolution and feasibility determination---are computable in polynomial time using well-established combinatorial algorithms, isolating reasoning, rather than computational hardness, as the primary challenge for LLMs.
Our main findings are as follows.

\begin{enumerate}[leftmargin=.8cm]
    \item \textbf{Epistemic Indeterminacy}: LLMs perform significantly worse on undetermined questions compared to determined ones, owing to systematic assumptions (e.g. lexicographic ordering of preferences) made when inputs are under-specified. 
    Providing an explicit indeterminacy option helps on a subset of tasks, but models continue to exhibit systematic reasoning errors.

    \item \textbf{Structural Indeterminacy}: LLMs' performance on tasks requiring algorithmic reasoning  degrades rapidly with market size---even in settings where solutions are guaranteed to exist---and deteriorates further in the presence of structural infeasibility, with models failing both to correctly identify infeasible instances and to generate valid solutions when they do exist.
    Providing an \textbf{indeterminacy option} (aka ``return null when infeasible'')  improves infeasibility detection but induces systematic bias, leading models to incorrectly declare non-existence on feasible instances.

    \item \textbf{Verification of Solution Concepts}: Even in selection tasks where only verification is required (as opposed to generation), LLMs continue to select incorrect options even when valid solutions are present. Although NOTA improves average accuracy, models seldom use this option, rarely abstaining even when no valid option is present, which reflects poor calibration in distinguishing feasible from infeasible cases.
    In addition, LLMs exhibit a systematic \textbf{intention–action misalignment} in preference reasoning: even when they appear to target a specific solution concept, the outcomes they select frequently fail to satisfy that concept, including in selection tasks where only verification is required.

    \item \textbf{Assisted Reasoning}: We consider assisted reasoning under two settings: \textbf{(i) refinement via feedback}, where models are given violations of the target property and attempt to iteratively repair invalid solutions, and \textbf{(ii) reasoning with code execution}, where models generate and execute programs to solve instances. While both settings improve performance, the gains are largely driven by brute-force enumeration on small markets and heuristic search on larger instances, neither of which scales to deployment-relevant market sizes.

\end{enumerate}

\subsection{Related Work}

\textbf{LLM Reasoning and Abstention.}
A growing body of work evaluates LLMs as procedural reasoners. 
Recent benchmarks show that LLMs fail to reliably execute classical algorithms as instance size grows~\citep{hosseini2025matching, shojaee2025illusion}.
A separate but related literature studies whether models know when to abstain: frontier models systematically miscalculate their uncertainty~\citep{kapoor2024large, yona-etal-2024-large}, hallucinate confidently even on questions they could answer correctly~\citep{Simhi2025Hallucinate, abbasi-yadkori2024to}, and in many settings fail to abstain at all~\citep{kirichenko2025abstentionbench}. Our findings sit at the intersection of these threads. 
Our findings sit at their intersection, linking procedural failure with abstention failure in preference reasoning. 

Beyond the fact that these benchmarks pose questions in natural language and do not consider structured preferences, two differences matter. First, because the ways of resolving an underdetermined query are enumerable in our setting, we identify the specific assumption a model substitutes for the missing information (e.g., ordering bundles by their highest-ranked item) rather than only whether it failed to abstain. Second, and more importantly, our benchmark includes problems for which the requested solution does not exist at all, a case with no counterpart in the abstention setting.

\textbf{LLMs in Economic Settings.}
Adjacent literature evaluate LLMs as agents in economic and social-choice contexts. On strategic decision-making, recent reasoning models come closer to equilibrium play than earlier ones~\citep{Jia2024Strategic, Wang2026Strategic} but remain susceptible to anchoring effects~\citep{Rios2025Tacit, Qian2026Strategic} and Bayesian inconsistency~\citep{Yamin2026Rational, Huynh2025Game}. Computational social choice, an extensively studied domain on which we build~\citep{long2021balanced, Amanatidis2022Resource, echenique2023online}, has begun engaging with LLMs both as fairness-aligned allocators~\citep{hosseini2025distributive, Cookson2026Fairness} and as solvers of canonical solution concepts~\citep{hosseini2025matching, fish2025econevals}, alongside applications to voting and participatory budgeting~\citep{Yang2024Voting, Thach2026PB}. A parallel line of work uses LLMs as proxies in preference elicitation pipelines~\citep{Handa2024Preference, Huang2025Accelerated, Li2025Preferences}, to which our work also contributes by characterizing how reliably LLMs parse structured preferences. \Cref{app:related_work} contains an extended related work.

\section{Preference-Based Tasks and Methodology}

\subsection{Formalizing Problems and Solution Concepts}

\textbf{Problem Domains.}
We consider three economic problems requiring reasoning over preferences: \textbf{house (or object) allocation} \citep{svensson1999strategy}, \textbf{Shapley-Scarf housing markets} with endowments \citep{shapley1974cores}, and \textbf{two-sided matching markets} \citep{gale1962college}, each with increasing structural complexity and distinct axiomatic solution requirements.
Let $A$ denote a set of agents and $B$ a set of alternatives (objects). Each agent $i \in A$ is endowed with a preference relation $\succeq_i$ over $B$, which is a weak and potentially partial order. We write $b_1 \succ_i b_2$ if agent $i$ strictly prefers $b_1$ to $b_2$, and $b_1 \succeq_i b_2$ if $i$ weakly prefers $b_1$ to $b_2$, allowing indifference. In two-sided markets each $b \in B$ likewise holds a preference relation $\succeq_b$ over $A$. In settings with incomplete preferences, $\phi \succ_i b$ indicates that $b$ is not present in agent $i$'s preference list and is thus unranked or incomparable. A \textbf{preference profile} is the collection of all agents' preferences, denoted by $\succeq = (\succeq_{a_1}, \dots, \succeq_{a_m}, \succeq_{b_1}, \dots, \succeq_{b_n})$, where $m = |A|$ and $n = |B|$.

In a \textit{house allocation} problem, agents in $A$ are assigned objects in $B$ with no initial endowments; in the \textit{Shapley--Scarf housing market}, each agent $i \in A$ is initially endowed with an object $e_i \in B$, forming an exchange economy; and in \textit{two-sided matching markets}, agents in $A$ and $B$ both have preferences over each other, inducing bilateral matching constraints. A \textbf{matching} is a mapping $\mu: A \cup B \rightarrow A \cup B$ such that for all $a \in A$, $\mu(a) \in B \cup \{\emptyset\}$ and for all $b \in B$, $\mu(b) \in A \cup \{\emptyset\}$, with the properties that each agent and object is matched to at most one counterpart and $b = \mu(a)$ if and only if $a = \mu(b)$. 

\textbf{Solution Concepts.}
Preferences and market structure induce standard \textbf{solution concepts}: in house allocation, the primary objective is \textbf{Pareto optimality}; in Shapley--Scarf markets, the central solution concept is the \textbf{core}, where no coalition of agents can reallocate their endowments to make all members strictly better off; and in two-sided matching markets, the standard concept is \textbf{stability}, requiring that no blocking pair exists. All three settings admit polynomial-time algorithms for computing canonical solutions and verifying feasibility under standard assumptions.

\textbf{Stronger Notions under Ties.}
When preferences admit ties, the canonical solution concepts above split into refinements that differ in their robustness to indifference. In Shapley--Scarf markets, a \textbf{weak core} allocation is one that no coalition can strictly improve upon, while the \textbf{strict core} requires that no coalition can find a reallocation under which every member is weakly better off and at least one is strictly better off. In two-sided matching, an analogous hierarchy arises: a matching is \textbf{weakly stable} if no pair strictly prefers each other to their current partners; \textbf{strongly stable} if no pair contains an agent who weakly prefers a partner who strictly prefers them back; and \textbf{super stable} if no pair contains agents who weakly prefer each other. The stronger refinements (strict core, strong stability, super stability) may be infeasible for some instances, generating the structural indeterminacy we study in \Cref{sec:algo_reasoning}. \Cref{app:soln_concepts} provides detailed formalisms, and a review of the relevant algorithms.

\subsection{Methodology and Experimental Setup}\label{subsec:methodology}
\textbf{Preference Expressivity and Reasoning Tasks.}
Preference expressivity gives rise to nuanced query tasks and solution concepts; we consider a spectrum of structures ranging from strict complete orders (SO), strict but possibly incomplete orders (SI), complete orders with ties (TO), and incomplete orders with ties (TI), as well as general partial orders (See \Cref{tab:preference-structures}).

We categorize evaluation into a hierarchy of increasing complexity to isolate where LLMs fail: \textbf{atomic queries} test direct preference retrieval, \textbf{comparative queries} test relational entailment/refutation, \textbf{aggregative queries} test collective preference computation, and\textbf{ structural (algorithmic) queries} test the ability to construct or verify solution concepts given an instance.
\Cref{tab:query_taxonomy} provides representative examples of determined and undetermined tasks in each category.

\textbf{Queries.}
For preference queries, we distinguish between \textbf{determined} queries, whose answers are uniquely implied by the input preferences, and \textbf{undetermined} queries, where the available information is insufficient to resolve the query. Formally in logic, the former corresponds to whether the query is \textit{entailed} (or \textit{refuted}) by the preference instance, while the latter implies lack of information.
For example, preference $a \succ c$ is determined if it is a necessary consequence of the provided axioms (e.g., $a \succ b$ and $b \succ c$ via transitive closure).
Otherwise, it is undetermined.

An undetermined query is scored correct only if the response indicates that the answer cannot be determined. A response that commits to an answer counts as incorrect even when it names the assumption it relied on. \Cref{app:f2_detail} re-scores those responses as correct and shows that the gap is essentially unchanged.

For algorithmic queries, given an instance, a solution concept is \textbf{feasible} if there exists a solution satisfying the concept, and \textbf{infeasible} otherwise.
For example, an instance of a matching market may admit \textit{no} super stable matching, in which case it is infeasible with respect to super stability.

This taxonomy enables us to move beyond ``accuracy'' and instead evaluate the Invalid Rate, i.e. the frequency with which a model produces a ``determined'' response to an ``undetermined'' query. This measure captures speculative completion, or the \textit{hallucination of certainty} \citep{Simhi2025Hallucinate,abbasi-yadkori2024to}, providing a principled diagnostic of a key failure mode in LLM reasoning.

\begin{table}[t]
\centering
\footnotesize
\caption{
Taxonomy of preference reasoning tasks, each exhibiting determined or undetermined instances depending on preference expressivity and feasibility of solutions.
}
\begin{tabular}{p{2cm} p{4.3cm} p{6cm}}
\toprule
\textbf{Task Type} & \textbf{Subtype} & \textbf{Example (Determined / Undetermined)} \\
\midrule

\textbf{Atomic} 
& Rank retrieval 
& Determined: rank of $a$ in $\succeq_i$ \newline Undetermined: rank of $b$ under incomplete list \\

\textbf{Comparative} 
& Pairwise, RS-extension, ranking 
& Determined: $a \succ_i b$ holds \newline Undetermined: $a \succ_i b$ under partial information \\

\textbf{Aggregative} 
& Preference aggregation (counts) 
& Determined: $\#\{i : a \succ_i b\}$ \newline Undetermined: count under missing comparisons \\

\textbf{Structural} 
& Core, stability (weak/strong/super) 
& Determined: existence of stable matching \newline Undetermined: existence under ties / incompleteness \\
\bottomrule
\end{tabular}

\label{tab:query_taxonomy}
\end{table}

\textbf{Models.}
\label{sec:models}
We evaluate four large language models that achieve state-of-the-art results in reasoning and coding benchmarks (see, for example \citep{ReinGPQA2023,Jiminez2024SWEBench}): \textbf{GPT-5.2} \citep{singh2025openai},
\textbf{Gemini-2.5-Pro} \citep{deepmind_2025}, \textbf{Claude-4.5-Sonnet}
(Claude-4.5-S) \citep{Anthropic_2025}, and a frontier-class open-source model \textbf{OSS-120B} \citep{agarwal2025gpt}.

\textbf{Prompt Generation.}
Each model was queried $30$ times on a given question type (See Appendix~\ref{app:prompts} for the prompt-templates used for each type of task).
Each query is a zero-shot, single-turn setting at the default temperature. 
In addition, we evaluated the models through two critical variations: 
i) \textbf{multi-shot refinements with feedback}, enabling models to refine and evaluate responses, and
ii) \textbf{code-execution capability}, enabling multiple iterations and with the ability to write code and verify responses (\Cref{sec:improvements}). Throughout the paper, we report results at four reference sizes by the number of agents/items $n$: \textbf{Small} ($n{=}10$), \textbf{Modest} ($n{=}30$), \textbf{Medium} ($n{=}50$), and \textbf{Large} ($n \geq 100$).

\section{Epistemic Indeterminacy: Preference Reasoning}\label{sec:pref_results}

We begin with epistemic indeterminacy: cases where a question is not answerable based on the information provided in the input preferences. Across the three query families introduced in \Cref{subsec:methodology} (atomic, comparative, and aggregative), we vary preference expressivity (i.e. completeness, ties, partial orders) to construct analogous questions of \textit{determined} and \textit{undetermined} instances on the same profile \footnote{By default, we use structured JSON format to express preference profiles as inputs to the models. In \Cref{app:nat_lang_pref} we include an additional experiment of rendering preferences in natural language, and demonstrate that our findings persist in that setting as well.} (see \Cref{sec:pref_benchmark} for details on how each type of question is designed). \Cref{fig:pref_questions_teaser} illustrates a few examples per query type. Our experiments investigate three criteria, (i) how performance depends on whether a query is determined, (ii) how it scales with input size, and (iii) how it depends on the way the question is framed. \footnote{We report model accuracy, aggregated by problem types in the plots. We include statistical analysis of each single experiment and paired comparisons in the corresponding tables in the appendices. An overview of our statistical methods is in \Cref{app:statistics}}.

\begin{figure}
    \centering
    \includegraphics[width=\linewidth]{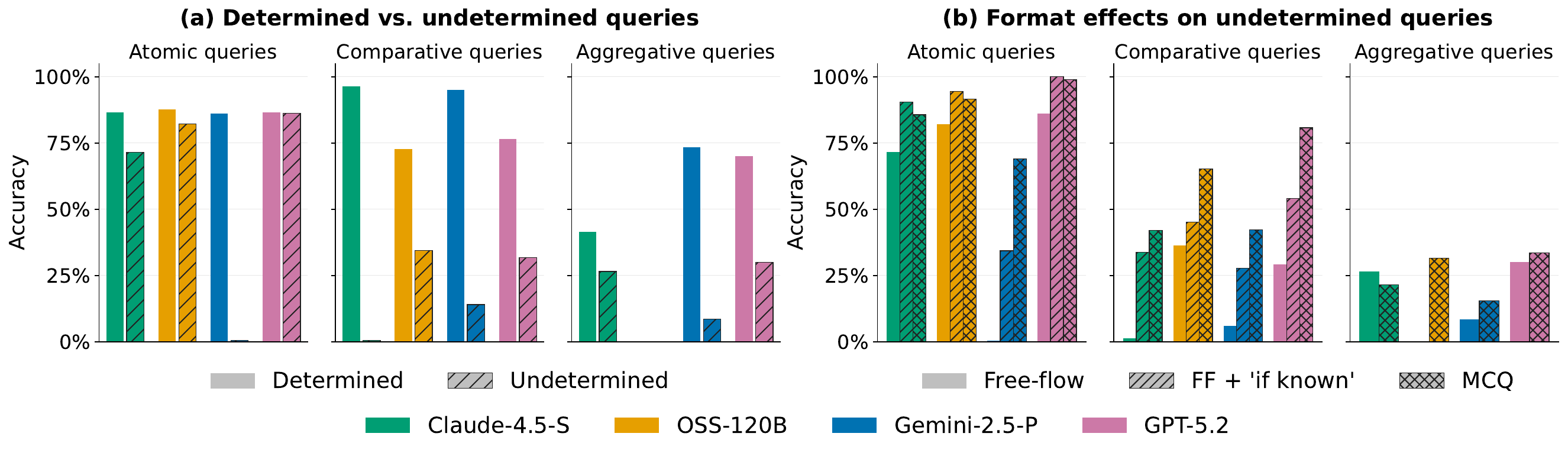}
    \caption{Comparing LLMs' performance (a) on determined vs. undetermined preference queries, and (b) with different prompting formats on undetermined preference queries, split by task type.}
    \label{fig:pref_bars_combined}
\end{figure}

\textbf{Determined vs. Undetermined Queries.} 
We compare performance on determined and undetermined queries across the question types in \Cref{fig:pref_questions_teaser}, considering Medium and Large 
preferences. To reflect a more realistic setting, queries are issued in a \textit{free-flow} format that does not provide explicit options to choose from. Responses are scored against the ground truth on determined queries, and on whether the model correctly indicates that the question is unanswerable, on undetermined ones (see \Cref{app:prompts} for the exact prompt templates).
\Cref{fig:pref_bars_combined} (a) shows that accuracy on undetermined queries is substantially lower than on the matched determined queries, collapsing to near-zero in several settings. The gap is driven by assumptions the models silently impose on the input, which suppress the under-specification rather than flag it. The gap is smallest on atomic queries, where the model only has to notice that an item is missing from the list, and largest on comparative queries, where nothing is missing and it has to work out that no answer follows from the preferences it was given. 

\begin{figure}
\centering
\includegraphics[width=0.7\linewidth]{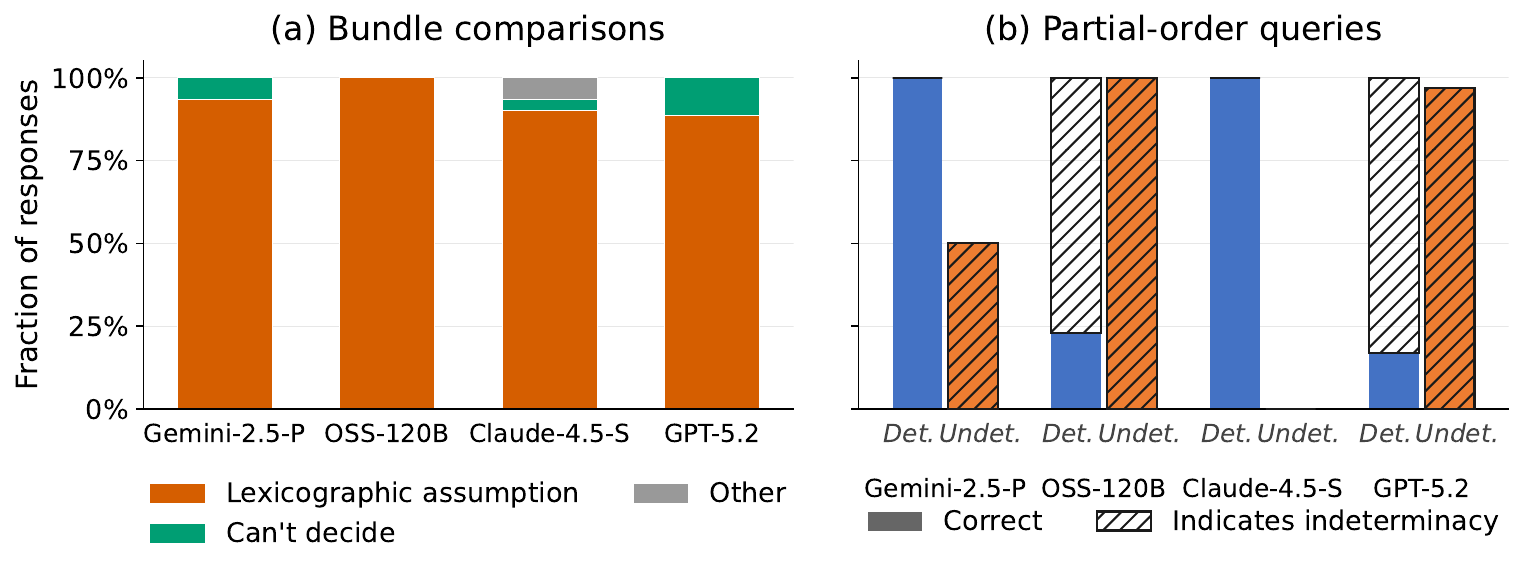}
\caption{Behavior of LLMs on RS-incomparable bundle comparisons (a), aggregated across different preference types, and partial-order pairwise queries (b). 
}
\label{fig:pref_assumptions}
\end{figure}

\textbf{LLMs' Systematic Biases.} We illustrate two such patterns, on bundle comparisons and on partial-order pairwise queries. Two bundles are \textit{responsive set} (RS) incomparable when neither can be paired with the other via an injection that maps each item to a weakly preferred counterpart. Intuitively, neither bundle dominates the other item-by-item. 
Resolving such a comparison requires an extra-axiomatic assumption, such as the lexicographic rule, which compares bundles by their most-preferred item, breaking ties by the next-most-preferred. As \Cref{fig:pref_assumptions} (a) shows, all four models default to lexicographic ordering on almost every RS-incomparable pair, across multiple preference types. We show in \Cref{app:f2_detail} that only in a small numbers of responses, this ordering is explicitly flagged as an "assumption", while others default to the rule silently. This assumption also persists even when the lexicographically weaker bundle is much larger (20 items vs. 1).

Partial-order queries elicit a different pattern. To answer whether $a$ is preferred to $b$, a model must check whether the stated comparisons induce a chain between the two items; if no chain exists in either direction, the pair is incomparable. \Cref{fig:pref_assumptions} (b) shows that models split into two failure modes. Claude-4.5-S and Gemini-2.5-Pro commit on undetermined pairs by either rejecting the input as incorrectly specified or using ad-hoc heuristics such as out-degree counts in the DAG induced by the stated comparisons (See \Cref{app:f3_detail} for details), at the cost of accuracy when no chain exists. OSS-120B and GPT-5.2 do the opposite: they perform an incomplete search and prematurely conclude that two items are incomparable, scoring well on undetermined queries but poorly on determined ones where a chain does exist.

\textbf{Framing Effects.}
The free-flow prompt used above does not signal that a query may be undetermined. To test whether the indeterminacy gap reflects a reasoning failure or a framing artefact, we evaluate two prompt variations: \textit{free-flow + ``if known''}, where the prompt ends with the qualifier ``if known'', and \textit{MCQ}, where the model chooses among a set of options that includes ``there is not enough information to decide'' (see \Cref{app:prompts} for the exact prompts).
\Cref{fig:pref_bars_combined} (b) shows that providing a way to report indeterminacy improves how often models do so depending on the task, with the largest gains under MCQ. However, we show in \Cref{app:f3_detail} through the example of partial-order queries, the MCQ format is potentially a double-edged sword. It improves Gemini-2.5-Pro's detection of indeterminacy while biasing Claude-4.5-S into selecting that option even when the query is determined.

\textbf{Scaling.}
Beyond the indeterminacy gap, accuracy degrades as the input grows. \Cref{fig:pref_size_scaling_3panel} (\Cref{app:pref_results_appendix}) shows a clear decline as the number of alternatives or agents moves from 100 to 200. The decline is more pronounced on determined queries: models make essentially no errors on atomic and comparative queries up to size 100, but begin to fail at size 200.

\section{Structural Indeterminacy: Algorithmic Reasoning over Preferences}\label{sec:algo_reasoning}

Algorithmic reasoning over preferences extends beyond preference inference, requiring models to (i) interpret structured preference profiles, (ii) execute multi-step procedures, and (iii) satisfy the constraints imposed by a target solution concept. In this setting, a fundamental challenge arises from \textbf{structural indeterminacy}: even under fully specified preferences, a solution concept may be infeasible for a given instance. This phenomenon is central to computational social choice, where the interaction between preference domains (e.g., strict versus weak orders) and solution concepts (e.g., core stability or matching stability notions) governs the existence of admissible outcomes.

In this section, we systematically evaluate the ability of large language models to reason in such settings. Specifically, we investigate: (i) whether LLMs can carry out algorithmic reasoning reliably as market size and problem complexity scale across different solution concepts; (ii) whether they can correctly identify infeasible instances, where no solution exists; and (iii) whether their behavior changes when provided with an \textit{indeterminacy option}, allowing them to indicate that no valid solution can be found.
We sample instances based on Impartial Culture \citep{black1958theory,eugeciouglu2013impartial} (i.e. uniformly at random), and perform a \textit{rejection sampling} strategy to include an equal number of feasible and infeasible instances (for solution concepts that admit both types). See \Cref{sec:algo_benchmark} for further details on how instances are created.

\subsection{Scaling, Infeasibility, and Indeterminate Option}\label{subsec:algo_reasoning_final}

\begin{figure}[t]
    \centering
    \begin{subfigure}[t]{0.41\textwidth}
        \centering
        \includegraphics[width=\linewidth]{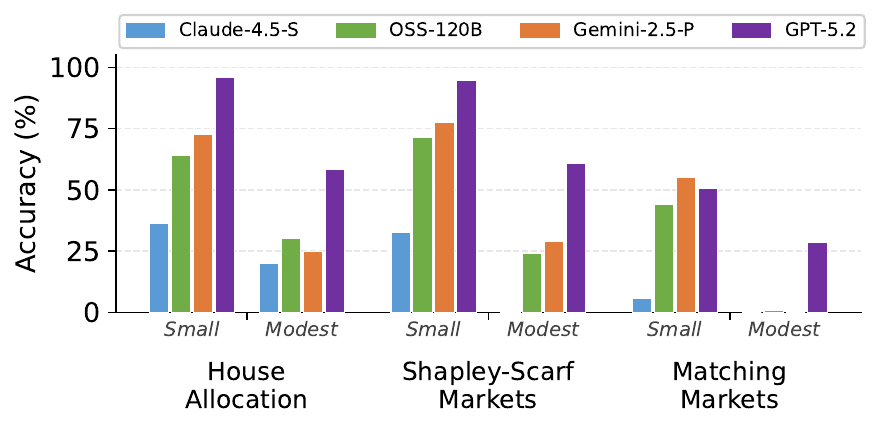}
        \caption{Decrease in accuracy of computing solutions due to an increase in input size, across different domains. 
        }
    \label{fig:size_collapse}
    \end{subfigure}
    \hfill
    \begin{subfigure}[t]{0.55\textwidth}
        \centering
        \includegraphics[width=\linewidth]{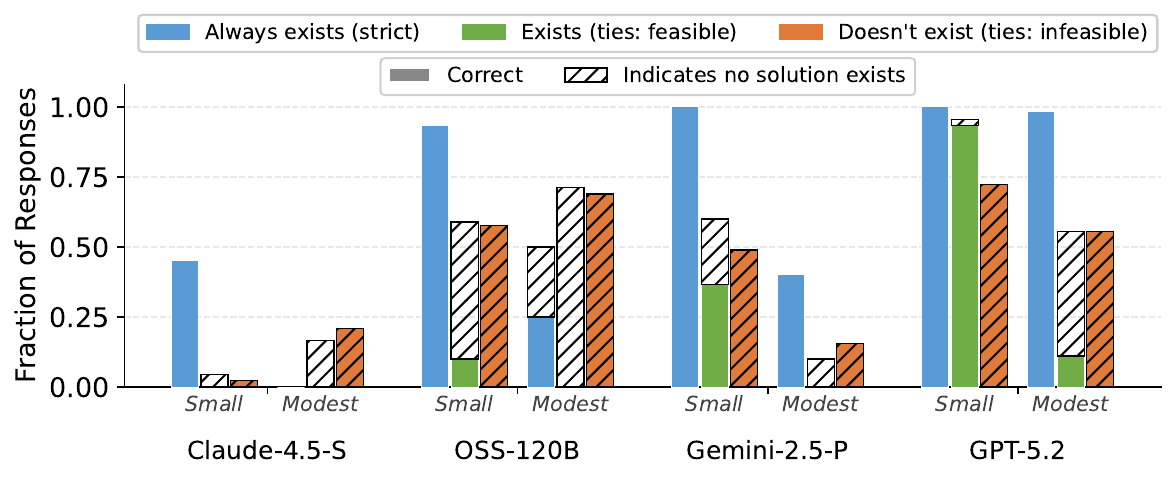}
        \caption{Performance on the generation task by model, problem size, and existence category, pooled across Shapley-Scarf and Matching Markets. 
        }
        \label{fig:feasible_infeasible_main}
    \end{subfigure}
\end{figure}

\textbf{Scaling.}
We begin with a baseline setting in which preferences are given as complete strict linear orders, ensuring that standard solution concepts are guaranteed to exist. We consider a range of canonical objectives, including Pareto optimality, core outcomes, and stability, as well as social welfare criteria such as egalitarian, utilitarian, and rank-maximal solutions.\footnote{See \Cref{app:soln_concepts}, for detailed formalism of each solution concept and \Cref{app:algo_results_appendix} (\Cref{fig:all_both_strict_complete_4}) for performance on each.}. As shown in \Cref{fig:size_collapse}, model performance degrades rapidly as market size increases, even in moderately sized instances. This decline is consistent across solution concepts, indicating limited scalability of LLMs in multi-step algorithmic reasoning tasks. While performance is comparatively stronger for concepts with well-structured canonical algorithms, the overall trend aligns with prior findings on the limitations of LLM reasoning in complex settings \citep{shojaee2025illusion, hosseini2025matching}.
In \Cref{app:algo_results_appendix}, we demonstrate that models default to canonical algorithms for the domain when the solution concept is unspecified (\Cref{app:gen_a_sol}), and display similar behavior when tasked with computing weaker solution concepts (\Cref{subsec:robustness}).

\textbf{Deciding Infeasibility.}
We consider two market settings, namely Shapley-Scarf and two-sided matching markets, in which the presence of ties in preferences gives rise to richer, but more demanding, solution concepts. 
In particular, stronger refinements such as the strict core (in contrast to the weak core) in house allocation, and super stability and strong stability (in contrast to weak stability) in matching markets, impose stringent constraints that may render instances infeasible. 
This lack of guaranteed existence is not incidental, but arises from the interaction between ties in preferences and the robustness requirements encoded by these solution concepts. 
Consequently, these markets provide a natural testbed for evaluating whether models can both construct valid solutions when they exist and correctly identify infeasibility when no such solutions are admissible.

\Cref{fig:feasible_infeasible_main} provides an overview of model performance across the two domains. As shown, all models exhibit declining performance both in correctly identifying infeasible instances, where no solution exists, and in generating valid outputs when the corresponding solution concepts are feasible.
In small markets ($n=10$), GPT-5.2 performs well in identifying infeasible instances, likely due to its ability to rely on heuristics and near-exhaustive search (see \Cref{app:reasoning-strategies} for details). However, even at modest scales (e.g., $n=30$), its performance deteriorates to the level of other models, as such approaches are no longer computationally feasible. 

\textbf{Indeterminacy Option.}
Given that certain instances may not admit a solution under specific concepts (e.g., super stability or the strict core), we further examine whether LLMs can correctly reason about such infeasibility when explicitly provided with an abstention option (i.e., declaring non-existence or returning a null solution). 

\begin{figure}
    \centering
    \includegraphics[width=\linewidth]{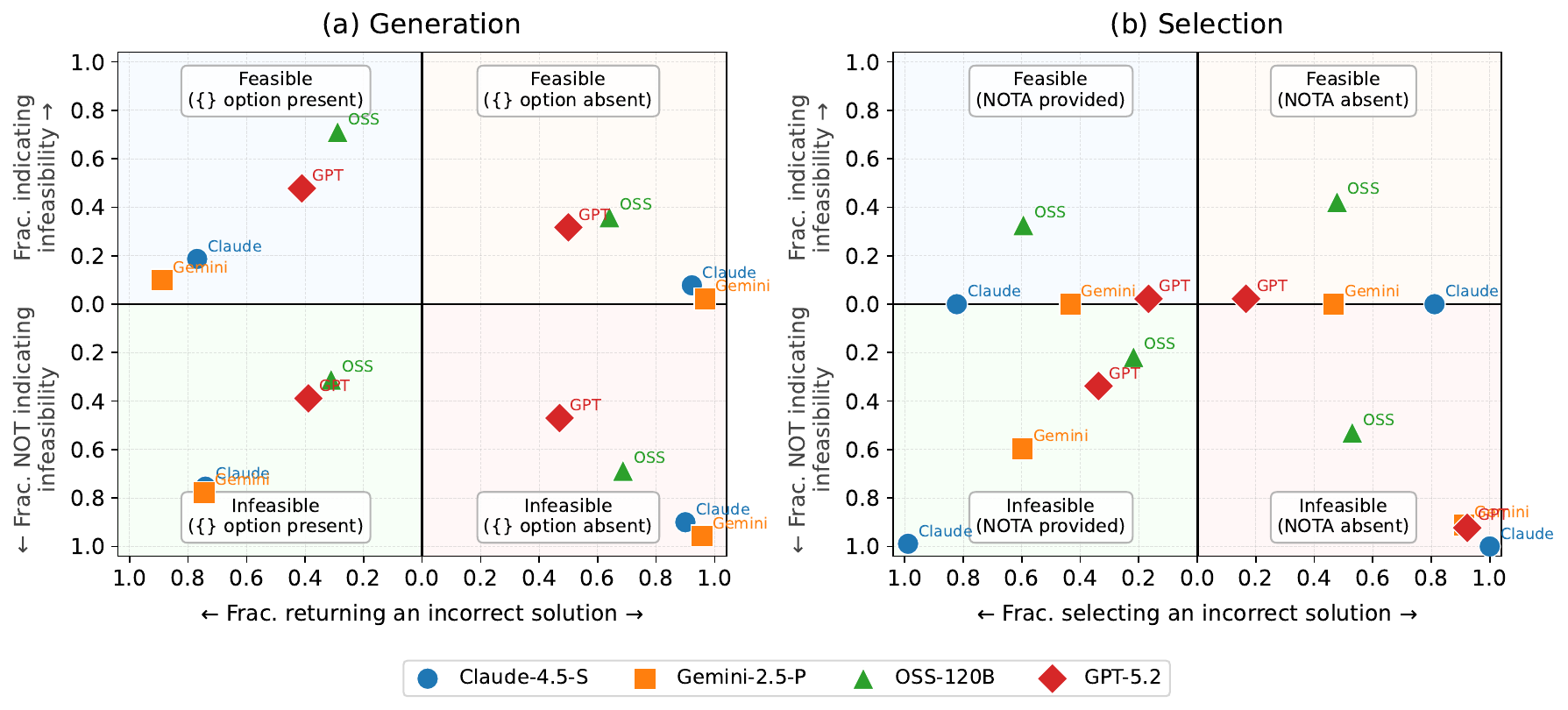}
    \caption{
    Errors made by models
    on generation (a) and selection (b) tasks as a function of problem feasibility and availability of an indeterminacy option, averaged over different solution concepts (Modest instance size). 
    An ideal model would cluster near the origin in all four quadrants. 
    }
    \label{fig:nota_vs_no_nota}
\end{figure}

\Cref{fig:nota_vs_no_nota} (a) summarizes model accuracy across three solution concepts in both markets, disaggregated by feasible and infeasible instances.
While the indeterminacy option improves accuracy in detecting infeasible instances, it introduces a systematic bias toward over-declaring non-existence, even when solutions exist. Consequently, performance on feasible instances deteriorates, as models increasingly abstain rather than produce valid solutions. 
\Cref{app:reasoning-strategies} provides an in-depth analysis of the heuristic strategies underlying these behaviors.

\subsection{Verification: Selection from a Menu of Options}

Thus far, we observe that language models struggle to correctly reason about solution concepts, both in generating feasible outcomes and in identifying infeasibility, even when explicitly provided with an abstention option.
This raises the question of whether these failures stem from limitations in algorithmic reasoning (i.e., executing the underlying steps) or from more fundamental deficiencies in feasibility reasoning. To disentangle these effects, we follow prior work (e.g., \citep{hosseini2025distributive}) in distinguishing between generation and selection tasks. Selection tasks are strictly easier, as they require only \textit{verification} of candidate solutions rather than the synthesis of a solution via multi-step procedures.

We consider the same market settings with weak preferences. For each instance, the model is presented with a set of pre-computed candidate solutions (five in general, and four for smaller instances) and is required to select a valid one. Instances are generated via rejection sampling (\Cref{sec:algo_benchmark}), and candidate solutions are constructed so that each satisfies a distinct solution concept, with no overlap across options. In certain variants, a ``none of the above'' (NOTA) option is included to permit abstention; in others, it is omitted to enforce selection.

\Cref{fig:nota_vs_no_nota} (b) shows that models frequently select incorrect candidates even when a valid solution is present.\footnote{Note that, in the selection case, nearly all incorrect responses select the ``weaker version'' of the intended notion, i.e. weak core instead of strict, weak stability instead of strong, and strong stability instead of super (see \Cref{subsec:harder_notions} for more details).} For infeasible instances, the removal of the NOTA option significantly increases the likelihood with which they select an incorrect option instead of indicating that no option applies. The effect is one-sided, i.e. when NOTA is available, no model selects it on feasible instances, and models under-select it on infeasible ones. 
Moreover, while GPT-5.2 substantially outperforms other models when a valid solution is present, its performance degrades markedly on infeasible instances, particularly as problem size increases or when the NOTA option is absent.

\textbf{Intention-Action Misalignment.} Given these failures in selection tasks, a key question is what objective LLMs are implicitly optimizing when selecting among candidate solutions.
To investigate this, we construct instances across different preference types---strict complete, complete with ties, and incomplete with ties---in each domain, such that the candidate options each satisfy distinct (and, in many cases, non-overlapping) solution concepts. Our objective is to characterize the implicit criteria guiding model behavior and to assess whether these criteria are consistently realized in the selected outcomes.

\Cref{fig:selection_stacked} (\Cref{app:intention_action}) shows the fraction of responses in which models select solutions satisfying various properties. To ensure comprehensive coverage, the candidate set includes outcomes aligned with standard welfarist criteria, such as utilitarian and egalitarian objectives.
We further infer model intentions using LLM-based judges, enabling a comparison between the property a model appears to target and the property actually satisfied by its selected solution.\footnote{We analyze reasoning traces to identify the ``intended'' concepts using LLM-as-a-judge. See \Cref{app:llm-judge} for details of the setup and validation of the judges.} This analysis reveals a systematic intention-action misalignment: models frequently imply adherence to a particular solution concept, yet select outcomes that violate the very criteria they appear to optimize.
This intention-action gap is especially pronounced in settings with ties or incomplete preferences, where structural constraints may render certain solution concepts infeasible.

\section{Performance Improvements with Assisted Reasoning}\label{sec:improvements}

We consider assisted reasoning under two settings: (i) refinement via feedback and (ii) reasoning with code execution, to examine whether the models recover their performance gap on indeterminacy. Both are presented below. In \Cref{app:prompt_mitigation}, we demonstrate how prompt-level mitigation such as providing few-shot examples or explicit instructions to not make assumption (both general and problem specific), fail to reliably calibrate the models.

\textbf{Refinements with Feedback.}\label{subsec:refinements}
\begin{figure}
    \centering
    \includegraphics[width=\linewidth]{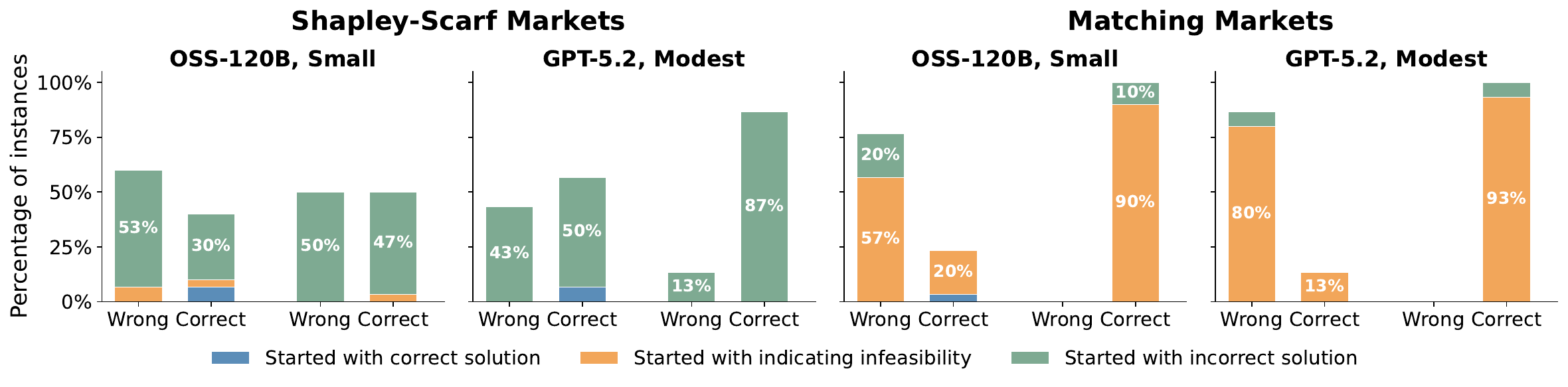}
    \caption{
    The effect of allowing models to refine their solutions using feedback.
    Each bar shows the fraction of instances that were ultimately answered correctly or incorrectly after up to three attempts with verification feedback. 
    }
    \label{fig:refinement_analysis}
\end{figure}
\Cref{subsec:algo_reasoning_final} showed that models struggle to compute solutions on potentially infeasible instances in a one-shot prompting setup. Although these are reasoning models with internal chain-of-thought, it is unclear whether their failures stem from being unable to recognize that a candidate solution is incorrect, or from being able to recognize the error but unable to repair it (because the complexity of the task inhibits them from verifying and then trying again). To distinguish the two, we evaluate OSS-120B and GPT-5.2 in a multi-round setup, where after each response we return structured feedback identifying the violation (a blocking pair in matching markets, a blocking coalition in Shapley-Scarf markets) and allow the model up to two retries (see \Cref{app:refinement} for more details). For each model we use the smallest instance size at which it begins to fail in the one-shot setting: Small for OSS-120B and Modest for GPT-5.2.

\Cref{fig:refinement_analysis} shows that feedback improves performance substantially on Shapley-Scarf markets but barely at all on matching markets. The asymmetry follows from what the models produce in the first round. On Shapley-Scarf instances they typically generate a valid (though incorrect) allocation, which the feedback then correctly redirects in many cases. On matching instances, both models are biased toward declaring infeasibility from the first round, and once a model has returned no solution there is nothing for the feedback mechanism to act on. Refinement is therefore effective only when the model produces a candidate solution to begin with; when its dominant failure mode is over-abstention, additional rounds do not change the outcome.

\textbf{Reasoning with Code Execution.}\label{subsec:code-reasoning}
In the previous sections we observed how LLMs fail to recognize when questions are undetermined, and are miscalibrated in identifying when a given solution concept is achievable for a given input. To understand whether these limitations are present only when LLMs reason in-context, we evaluate their performance when allowed to write code to solve the same problems, given their impressive ability to write code \citep{Jiminez2024SWEBench,jain2024livecodebench}.

For preference reasoning, solving determined questions with code execution completely eliminates errors related to scale. However, on undetermined queries such as (RS-incomparable) bundle comparisons and the ranking task, the answers are \textit{always} incorrect. The assumptions models make now become part of the implemented algorithmic steps. The contrast with the structural setting is informative. There, models often write an explicit feasibility check and return no solution when the search finds none (\Cref{app:reasoning-strategies}), but on undetermined preference queries they write no such check.

For algorithmic reasoning, allowing LLMs to write code to solve both feasible and infeasible problems significantly improves their performance, although not by implementing the ``correct'' approach (see \Cref{tab:code-accuracy} in \Cref{app:code_approaches} for exact accuracy levels). In fact, their approach changes with the input size. For Small instances, the models resort to brute-force enumeration of all solutions. The strategy changes for Modest and for Medium instances, where they could use slightly more efficient heuristics to solve the problem (See \Cref{app:code_approaches} for detailed descriptions of the approaches used).

However, as becomes clear with Medium instances, these approaches are not scalable. For Shapley-Scarf Markets, the models perform well on feasible instances, but often time-out on infeasible ones\footnote{We set a limit of 20 minutes after which (if the model doesn't return a response) the response is treated as timed-out.}. Similarly, the strategy used for Matching Markets misidentifies feasible instances as infeasible, in the few responses that do not time-out. Hence, while code improves performance on smaller markets, it does not enable models to recognize and use approaches relevant for deployment-level sizes.

\textbf{Dissociating the two axes.} The two forms of indeterminacy respond to different interventions, which is what separates them from a single shared calibration failure. Code execution improves structural reasoning but not epistemic queries, where the model compiles its assumption into the program it writes. Prompt-level intervention does the reverse. It raises indeterminacy detection on preference queries (\Cref{fig:pref_bars_combined}(b)) but does not improve feasibility reasoning due to structural (in)determinacy. 
This suggest that the escape hatch instead moves the error into over-abstention (\Cref{fig:nota_vs_no_nota}(a)).

\section{Concluding Remarks}

We argue that indeterminacy is a fundamental dimension of reasoning in preference-based AI systems. As language models increasingly mediate alignment, coordination, and collective decisions, current closed-world training paradigms systematically bias models toward hallucinating determinacy, feasibility, and preference information where none is entailed---a broader manifestation of \textit{verisimilitude} in AI, where outputs appear plausible despite lacking logical grounding. Our results suggest that robust AI will require open-world reasoning frameworks (e.g. multi-valued semantics), richer notions of non-entailment and impossibility, and benchmarks that evaluate not only correctness, but the ability to recognize when no justified answer exists.

\section*{Acknowledgments}
This research was supported in part by NSF Awards IIS-2144413 and IIS-2107173. We also thank the anonymous reviewers for their careful reading and constructive feedback, which improved the paper.

\bibliography{biblio}
\bibliographystyle{plainnat}

\clearpage

\appendix

\begin{center}\large \centering
    \textbf{Technical Appendices and Supplementary Material}
\end{center}

\section{Generative AI Use Statement}\label{app:genai}
AI tools were used in three ways in the course of this work. First, they assisted with implementation, including writing and debugging the benchmark generation, inference, and analysis code. Second, they were used to polish the writing of the main paper, without generating any of its claims or results. Third, they helped synthesize findings about the reasoning traces collected in the LLM-judge experiments of \Cref{app:reasoning-strategies} and \Cref{app:llm-judge}, where the judge outputs are themselves model-generated and are validated against additional judges and a human annotator in \Cref{app:judge_validation}. All experimental results, numbers, and claims in the paper were produced and checked by the authors.

\section{Limitations and future directions}\label{app:limitations}

Our evaluation is bounded in three ways that are worth stating. First, we test four models, spanning two closed families and one open-weights family. They are the current frontier, but the panel is small, so we report per-model results throughout rather than a single pooled number. Second, we sample instances under Impartial Culture, which lets us build matched determined and undetermined instances on the same profile and control feasibility exactly. The sampler decides which instances turn out feasible, not the reasoning a task demands, so we do not expect structured distributions such as Mallows or single-peaked preferences to change the picture, though we have not tested this. Third, two of our analyses infer what a model intended from its reasoning trace, and so depend on a judge model. \Cref{app:judge_validation} reports that judge's agreement with two additional judges and with a human annotator, and flags the one cell where they disagree. Every other result is scored by deterministic checkers, which we validate against a brute-force solver in the same appendix. 

These bounds aside, our evaluation establishes what current reasoning models do when confronted with indeterminate preference queries, and leaves open why they do it. The systematic assumptions we identify (e.g. lexicographic tie-breaking on bundles, premature termination on partial-order chains) are robust empirical patterns, but their underlying mechanism is opaque to the input-output methodology used here. Recent advances in mechanistic interpretability~\citep{lindsey2025biology, ameisen2025circuit, templeton2024scaling} suggest that these failures could be traced to specific circuits or features within the model, illuminating whether models silently override their own indeterminacy detection or fail to recognize it altogether. 

The role of prompt framing is likewise only partly mapped. We vary it along three coarse axes (free-flow, free-flow + ``if known'', MCQ) but do not characterize the threshold at which a prompt becomes informative enough to elicit reliable abstention. Finer-grained probing along the lines of \citet{sclar2023quantifying} and \citet{kirichenko2025abstentionbench} could map this transition more precisely. 

A final question is which aspects of training data and post-training procedures determine a model's calibration to indeterminacy. Recent advances in scalable data attribution~\citep{choe2026what} and frameworks that bridge interpretability and data influence~\citep{pan2026mda}, combined with analyses of calibration shifts during alignment~\citep{kadavath2022language, tian2023just}, could distinguish whether targeted fine-tuning would address the underlying calibration issue or merely mask the surface symptoms.

\section{Broader Impacts}\label{app:broader_impacts}

The findings in this paper bear directly on two deployment contexts where LLMs are increasingly used to reason over preferences. First, agentic systems acting on behalf of users — making purchases, scheduling commitments, mediating between conflicting goals — must distinguish between situations in which the user's stated preferences determine an action and situations in which they do not. Our results show that current frontier models systematically fail at this distinction, which means an agentic system built on these models may, by default, take actions that the user has not actually authorized, while presenting them as reflecting the user's preferences. Second, platforms that aggregate preferences for collective decision-making — participatory budgeting tools, deliberation platforms, large-scale opinion-mapping systems — depend on the assumption that the aggregation faithfully reflects the inputs. When the input is structurally indeterminate (incomplete preferences, infeasibility under ties, etc.), a model that confidently produces an aggregate is generating an artificial consensus that does not exist in the data. Our work surfaces these failure modes in a controlled evaluation, with the goal of making them measurable and addressable before such systems are deployed in higher-stakes settings.

\section{Extended Related Work}\label{app:related_work}

\textbf{LLM Reasoning Benchmarks.}
Existing benchmarks evaluate LLMs on mathematical reasoning~\citep{CobbeGSM8K2021, HendrycksMATH2021, wang2024mmlu}, scientific reasoning~\citep{phan2025humanity, ReinGPQA2023}, and code generation~\citep{jain2024livecodebench, Jiminez2024SWEBench}. These assess whether a model reaches the correct answer, but provide limited insights into the ability of LLMs to execute algorithms correctly. Two recent papers show that LLMs fail to execute well-known algorithms as instance size grows: \citet{hosseini2025matching} trace this collapse in the commonly used Deferred Acceptance algorithm to compounding step-level errors, and \citet{shojaee2025illusion} find the same pattern on combinatorial puzzles (e.g. the missionaries and cannibals problem, tower of Hanoi, etc.). Both papers characterize algorithmic failure without distinguishing among algorithms — that is, they do not ask whether some algorithms are harder than others or why. We build on this by asking a finer question: do LLMs fail uniformly across algorithms, or is performance related to properties of the algorithm itself? Our findings suggest the latter, with training-data familiarity playing a larger role than computational complexity.

\textbf{Rationality and Strategic Reasoning.}
A complementary literature asks whether LLMs behave as rational agents in strategic settings. Recent reasoning models come closer to equilibrium play than earlier ones~\citep{Jia2024Strategic} and identify strategic patterns faster than humans~\citep{Wang2026Strategic}, yet LLMs still exhibit anchoring effects and model-specific biases that prevent convergence to optimal solutions~\citep{Rios2025Tacit, Qian2026Strategic}, and their stated probabilities over outcomes frequently violate basic Bayesian consistency conditions~\citep{Yamin2026Rational, Huynh2025Game}. These papers evaluate LLMs as strategic actors optimizing over outcomes. Our focus is orthogonal: we evaluate LLMs as procedure executors applying a specified algorithm to a given input. This distinction is consequential because optimal strategy computation in many settings itself requires running sophisticated algorithms (for instance, computing the core of a cooperative game or finding a stable matching) so failures in procedure execution place a ceiling on strategic rationality as well~\citep{nisan2007algorithmic}. \citet{Hua2024Game} show that LLMs can achieve Pareto-optimal and envy-free outcomes in negotiation settings; our results suggest this may reflect approximate pattern-matching rather than principled computation, since the same notions are computed unreliably when formally defined and applied to structured inputs.

\textbf{LLMs for Social Choice.} 
Computational social choice studies how individual preferences can be aggregated into collective decisions, with active subliteratures on house allocation \citep{abdulkadirouglu1999house,long2021balanced,Hosseini2024House, Hosseini2026House,Schlotter2025Core, Aziz2024House,Waldinger2021House}, resource allocation \citep{Amanatidis2022Resource, Aziz2022Goods,Liu2024Mixed, Igarashi2024Repeated}, and matching markets \citep{echenique2023online, Celebi2022Matching,rong2024core, Aziz2022Matching,Knipe2025Improvable}. As LLMs are increasingly proposed as components of decision-making pipelines in these domains, recent work has begun to ask both whether LLMs can compute the outcomes that social-choice theory prescribes, and whether their default behaviour aligns with how humans think about fairness.

On the fairness alignment side, \citet{hosseini2025distributive} and \citet{Cookson2026Fairness} both find that when LLMs are asked to allocate resources fairly, their choices differ from human judgments in the same way: they tend to maximise overall welfare rather than spread benefits equally, and this pattern is robust to changes in how the prompt is phrased. \citet{hosseini2025distributive} additionally find that what the LLM says it will do does not always match the allocation it actually produces. We recover both findings on a broader range of solution concepts and preference structures (Section~\ref{app:intention_action}), suggesting that they extend beyond resource allocation to allocation and matching tasks more generally. On the solving side, \citet{hosseini2025matching} evaluate LLMs as one-shot solvers on stable-matching instances, while \citet{fish2025econevals} embed an LLM in an iterative loop, giving it feedback in the form of blocking pairs and measuring how its solution improves over rounds. Both papers analyse the kinds of failures the model makes; we contribute to this line of work by studying how LLMs behave when a question is underspecified or when no valid solution exists. Other recent work has applied LLMs to voting~\citep{Yang2024Voting}, where the choice an LLM makes among candidates depends on the order in which the options are listed and on which voting rule is used, and to participatory budgeting~\citep{Thach2026PB}, where LLMs are used to design heuristics for selecting which projects to fund.

\textbf{LLMs for Preference Elicitation.}
LLMs are increasingly deployed as proxies for human preferences in elicitation pipelines, using techniques ranging from Bayesian query selection~\citep{Handa2024Preference, Austin2024Preference} to LLM-simulated preference profiles~\citep{Huang2025Accelerated, Montazeralghaem2025Preference} and direct comparative elicitation~\citep{Li2025Preferences}. These systems implicitly assume that LLMs can reason reliably about structured preferences. Our benchmark directly tests this assumption and finds it to be optimistic, particularly under incomplete information and multi-agent aggregation.

\textbf{Abstention and Uncertainty.}
\citet{kirichenko2025abstentionbench} show that abstention, i.e. the ability to refrain from answering when unsure,
is unsolved across frontier models and that reasoning fine-tuning can degrade it. Systematic miscalibration between models' intrinsic uncertainty and how they express it is well documented~\citep{yona-etal-2024-large, kapoor2024large}, and models can hallucinate with high confidence even on questions they have the capacity to answer correctly~\citep{Simhi2025Hallucinate,abbasi-yadkori2024to,bhattacharyya2026confidence}. A related failure appears in moral decision-making, where models do not reproduce the indecision humans express on hard choices, even though that indecision signals a need for further deliberation~\citep{dickerson2026Kidney}. Our results contribute a more structured instance of this failure: when preference queries are formally underdetermined, models do not abstain but instead make implicit, predictable assumptions, such as lexicographic tie-breaking or alphabetical completion of partial lists. Crucially, providing an explicit uncertainty option eliminates this behavior in some settings but not others, suggesting the failure is not purely one of output formatting but of recognizing underdetermination itself.

\section{Extended Preliminaries}\label{app:prelims}

\begin{table}[ht]
    \centering \footnotesize
    \caption{Preference expressivity, axioms, and properties. In each case, we consider complete orders and incomplete orders where the preference is truncated (missing the tail). 
    }
    \label{tab:preference-structures}
    \begin{tabular}{lp{5cm}p{5cm}}
        \toprule
        \textbf{Preference Expressivity} & \textbf{Description} & \textbf{Determining Property} \\
        \midrule
        Complete Strict Linear Order & Asymmetric, transitive, and total. & Always determined. \\
        \addlinespace
        Complete Order with Ties & Allows ties ($a \sim b$); total. & Determined (indifference is explicit). \\
        \addlinespace
        Incomplete Order with Ties & Allows ties and missing pairs. & Undetermined if incomparable ($a \parallel b$). \\
        \addlinespace
        Partial Order / Pairwise Sets & Transitive and asymmetric; missing pairs. & Determined only if in the transitive closure. \\
    \bottomrule
    \end{tabular}
\end{table}

\subsection{Solution Concepts}\label{app:soln_concepts}
We define the solution concepts evaluated in this work, grouped by the domain and preference structure in which they apply.

\textbf{Pareto optimality.} An outcome $\mu$ is \emph{Pareto-optimal} (PO) if there exists no feasible outcome $\mu'$ such that $\mu'(i) \succeq_i \mu(i)$ for all $i \in N$ and $\mu'(j) \succ_j \mu(j)$ for some $j \in N$. Under incomplete preferences, we additionally consider \emph{maximum-cardinality Pareto-optimal} (MCPO) allocations: among all allocations that match the maximum number of agents to ranked alternatives, those that are Pareto-optimal.

 \textbf{Core (Shapley--Scarf housing market).}
The core and TTC are notions of the endowment economy, so we state them for the Shapley--Scarf market rather than for house allocation without endowments. Let $S \subseteq N$ be a coalition. Under \textbf{strict preferences}, $S$ \emph{blocks} an allocation $\mu$ if there exists a reassignment $\mu_S$ of $\{e(i) : i \in S\}$ among members of $S$ such that $\mu_S(i) \succ_i \mu(i)$ for all $i \in S$. An allocation is in the \emph{core} if no coalition blocks it. The core always exists under strict preferences and coincides with the unique output of Top Trading Cycles (TTC)~\citep{shapley1974cores,roth1977weak}.
 
Under \textbf{preferences with ties}, the definition of blocking depends on how ties are treated, giving rise to two notions:
\begin{itemize}
    \item \textbf{Weak core}: $S$ blocks $\mu$ if there exists a reassignment $\mu_S$ such that $\mu_S(i) \succ_i \mu(i)$ for all $i \in S$ (every member strictly improves). An allocation is in the weak core if no such coalition exists. The weak core always exists and coincides with the TTC output under a consistent tie-breaking rule.
    \item \textbf{Strict core}: $S$ blocks $\mu$ if there exists a reassignment $\mu_S$ such that $\mu_S(i) \succeq_i \mu(i)$ for all $i \in S$ and $\mu_S(j) \succ_j \mu(j)$ for some $j \in S$ (every member weakly improves, at least one strictly improves). An allocation is in the strict core if no such coalition exists. The strict core may be empty, but can be computed in polynomial time if it exists~\citep{QuintWakoHA}.
\end{itemize}
 
Note that under strict preferences the two notions coincide (weak preference improvements always resolve to strict ones when no indifference exists), so we simply refer to the \emph{core} in that setting. It is worth making the uniqueness explicit: under strict preferences the strict core is a singleton, namely the TTC allocation~\citep{roth1977weak}, so the distinction between the weak and the strict core only has content once ties are present.
 
\textbf{Welfare objectives.} For an allocation or matching $\mu$:
\begin{align*}
    \mathrm{UW}(\mu) &= \sum_{i \in N} -1*\mathrm{rank}_i(\mu(i)), \qquad \text{(Utilitarian Welfare, maximized)}\\
    \mathrm{EW}(\mu) &= \max_{i \in N}  -1*\mathrm{rank}_i(\mu(i)). \qquad \text{(Egalitarian Welfare, maximized)}
\end{align*}
A \emph{utilitarian welfare-maximizing} (UW) outcome maximizes $\mathrm{UW}(\mu)$ \citep{AZIZ2023Welfare}; an \emph{egalitarian welfare-maximizing} (EW) outcome maximizes $\mathrm{EW}(\mu)$ \citep{Sen73Welfare}.
 
\textbf{Rank-maximality.} For an outcome $\mu$, define its \emph{rank vector} $r(\mu) = (r_1, r_2, \ldots, r_m)$, where $r_k$ is the number of agents assigned an alternative of rank $k$. An outcome is \emph{rank-maximal} (RM) if its rank vector is lexicographically maximal over all feasible outcomes \citep{Irving2006RankMax}. 

\begin{example}[Rank vector]
With three agents assigned to rank-1 alternatives, none to rank-2, and one to rank-3, the rank vector is $(3,0,1)$. Rank-maximality prefers $(3,0,1)$ to $(2,2,0)$: as many rank-1 assignments as possible first, then as many rank-2, and so on.
\end{example}
 
\textbf{Stability (two-sided matching).}
Let $(m, w)$ be an unmatched pair. Under \textbf{strict preferences}, $(m, w)$ \emph{blocks} a matching $\mu$ if $w \succ_m \mu(m)$ and $m \succ_w \mu(w)$, i.e.\ both strictly prefer each other to their current partners. A matching is \emph{stable} if no blocking pair exists. Stable matchings always exist and can be computed by Deferred Acceptance (DA)~\citep{gale1962college}.
 
Under \textbf{preferences with ties}, the meaning of ``blocking'' is no longer unique, yielding three stability concepts of increasing strength:
\begin{itemize}
    \item \textbf{Weak stability}: $(m, w)$ blocks $\mu$ if both strictly prefer each other to their assigned partners ($w \succ_m \mu(m)$ and $m \succ_w \mu(w)$). This coincides with the strict-preference definition; weakly stable matchings always exist.
    \item \textbf{Strong stability}: $(m, w)$ blocks $\mu$ if both weakly prefer each other to their partners and at least one strictly prefers ($w \succeq_m \mu(m)$, $m \succeq_w \mu(w)$, and at least one inequality is strict). Strongly stable matchings may not exist, but can be found in polynomial time when they do~\citep{IRVING1994Indifference}.
    \item \textbf{Super stability}: $(m, w)$ blocks $\mu$ if both weakly prefer each other to their partners ($w \succeq_m \mu(m)$ and $m \succeq_w \mu(w)$). Super stable matchings may not exist, but can be found in polynomial time when they do~\citep{IRVING1994Indifference}.
\end{itemize}
 
The three notions are nested: every super stable matching is strongly stable, and every strongly stable matching is weakly stable. The converse does not hold in general. Note that, the definitions of a stable matching when preferences are strict, and that of a super stable matching when preferences have ties, are equivalent (similar to the relationship between strict core and core in the house allocation setting).

\textbf{Canonical algorithms.}
Each domain admits a canonical algorithm, i.e. one that is not only computationally efficient but is uniquely characterized by a combination of desirable axiomatic properties, making it the natural first choice for practitioners and the standard reference in the literature.
 
\begin{itemize}
    \item \textbf{Top Trading Cycles (TTC)} for the Shapley--Scarf housing market computes the unique allocation in the core \citep{shapley1974cores} and is the only mechanism that is individually rational, Pareto-optimal, and strategy-proof \citep{ma1994strategy,roth1982economics}.
    \item \textbf{Serial Dictatorship (SD)} for object allocation is the unique mechanism that is Pareto-optimal, strategy-proof, and non-bossy \citep{svensson1999strategy}.
    \item \textbf{Deferred Acceptance (DA)} for stable matching computes a stable matching that is weakly preferred by the proposing side to all other stable matchings, and is the unique stable and strategy-proof mechanism (for that side) \citep{gale1962college,roth1982economics}.
\end{itemize}

\section{Dataset Creation Details}

\subsection{Preference Reasoning Tasks}
\label{sec:pref_benchmark}

\subsubsection{Instance Generation}
\label{sec:pref_instance_gen}
 
\textbf{Strict and complete preferences.}
Preferences are sampled under the impartial culture model: each agent's ranking is drawn uniformly at random from all permutations of $A$. Profile sizes range from 30 to 200 items; results are reported at sizes 50, 100, and 200.
 
\textbf{Strict and incomplete preferences.}
Starting from a strict and complete preference list, each agent's ranking is truncated to a length drawn uniformly from $[\lfloor 0.5n \rfloor, n]$, where $n$ is the number of items. Items beyond the truncation point are unranked.
 
\textbf{Complete preferences with ties.}
Starting from a strict and complete preference list, ties are introduced by randomly selecting contiguous ranges of items and merging them into indifference classes. Each generated profile contains two tie groups, with at least one item ranked outside any tie. The starting and ending position of each tie are selected randomly from possible values. Once these position are selected, we insert brackets into these positions and indicate that the items within the brackets are tied.
 
\textbf{Partial order preferences.}
Partial-order preferences are generated as Erd\H{o}s--R\'{e}nyi DAGs~\citep{erdds1959random} (a common approach in prior work \citep{brightwell1993models,Jang2016Pairwise}) oriented consistently with a latent total order. We use up to 100 nodes and edge counts of 50, 80, or 150 (denser graphs yielding fewer incomparable pairs).
 
\subsubsection{Benchmark Design}
\label{sec:pref_benchmark_design}
 
The preference reasoning benchmark consists of task families, each targeting a different aspect of preference comprehension.

\textbf{Atomic queries.}
Retrieval regarding single item within the preference profile:
\begin{itemize}
    \item \textbf{Position query}: at what position does a given item appear in the agent's preference?
    \item \textbf{Item query}: which item does the agent rank at a given position $k$?
\end{itemize}

\textbf{Comparative queries.}
Relative preference of multiple items by an agent:
\begin{itemize}
    \item \textbf{Item comparison}: between 2 given items, which is preferred by the agent?
    \item \textbf{Item ranking}: given a set of four items, rank them according to the agent's preference.
\end{itemize}

Under strict and complete preferences, each question has a unique determined answer. Under strict and incomplete preferences, some questions become undetermined (e.g.\ the queried item may be absent, or $k$ may exceed the length of the list), and the correct response is to indicate that the answer cannot be determined. Under complete preferences with ties, we additionally include a \textbf{top-$k$ query}: does a given item belong to the agent's top-$k$ alternatives? When $k$ falls within a tie group, this question is undetermined.
 
For all question types, models are prompted to return their answer in a specified format. In the free-flow format, we do not provide explicit options. For undetermined questions, two free-flow variants are used: one that includes ``if known'', and one that does not. We additionally create MCQ versions for all problems and explore the impact of prompt format on model responses.
 
\textbf{Bundle comparisons.}
A special case of comparative task. Given a single agent's preference ranking and two bundles $B, B' \subseteq A$, the model is asked which bundle the agent prefers. Pairs are drawn from three preference types (strict and complete, strict and incomplete, complete with ties), and are constructed to include both RS-comparable cases (where a definite answer exists) and RS-incomparable cases (where the correct response is that the comparison is undetermined). 
 
\textbf{Aggregative queries.}
Given a preference profile over multiple agents, the model is asked: ``how many agents prefer item $a$ over item $b$?'' This requires scanning all agents' lists and tallying correctly. We evaluate accuracy (fraction of exact correct counts) and mean absolute error on incorrect responses.
 
\textbf{Partial order queries.}
Given a preference specified as a set of pairwise comparisons, the model is asked whether item $a$ is preferred to item $b$. We test both comparable pairs (connected by a chain of stated comparisons) and incomparable pairs (no such chain exists).
 
We additionally evaluate a second representation (\textbf{Format~2}) in which partial order information is conveyed as query-answer pairs: each data point specifies a subset of items and the most preferred item within it. Although logically equivalent to pairwise comparisons, this representation requires an additional inference step to recover the underlying structure. 
 
\subsubsection{Prompt Counts}
\label{sec:pref_prompt_counts}
 
\begin{table}
\centering
\caption{Preference reasoning benchmark composition. Profile size refers to the number of items in the preference list. In formats, ``ff'' refers to free-flow prompt, ``ff\_if\_known'' refers to free-flow prompt, added with "if known" phrase, and MCQ refers to multiple-choice question.}
\label{tab:pref_benchmark}
 \resizebox{\textwidth}{!}{
\begin{tabular}{lllllc}
\hline
\textbf{Task family} & \textbf{Pref. Type} & \textbf{Question types} & \textbf{Profile sizes} & \textbf{Formats} & \textbf{Prompts} \\
\hline
\multirow{2}{*}{Atomic} & \makecell[l]{Strict +\\complete} & \makecell[l]{Item query,\\position query,~} & 50, 100, 200 & ff, MCQ & 360 \\
 & \makecell[l]{Strict +\\incomplete} & \makecell[l]{Item query\\(undetermined),\\position query\\(undetermined),} & 50, 100, 200 & \makecell[l]{ff, \\ff\_if\_known,\\MCQ} & 540 \\
\hline
\multirow{2}{*}{Comparative} & \makecell[l]{Strict +\\complete} & \makecell[l]{item comparison,\\item ranking} & 50, 100, 200 & ff, MCQ & 360 \\
 & \makecell[l]{Strict +\\incomplete} & \makecell[l]{item comparison\\(undetermined),\\item ranking} & 50, 100, 200 & \makecell[l]{ff, \\ff\_if\_known,\\MCQ} & 720 \\
\hline
\multirow{3}{*}{\makecell[l]{Bundle\\comparison}} & \makecell[l]{Strict +\\complete} & \makecell[l]{RS-comparable,\\RS-incomparable,\\larger bundle,\\unequal bundle sizes} & 50, 100 & \makecell[l]{ff, \\ff\_if\_known,\\MCQ} & 300 \\
 & \makecell[l]{Strict +\\incomplete} & \makecell[l]{RS-comparable,\\RS-incomparable} & 50 & ff\_if\_known & 60 \\
 & \makecell[l]{Ties +\\complete} & \makecell[l]{RS-comparable,\\RS-incomparable} & 50 & ff\_if\_known & 60 \\
\hline
\makecell[l]{Top-k\\query} & \makecell[l]{Ties +\\complete} & \makecell[l]{Determined,\\Undetermined,\\break-down queries} & 50 & \makecell[l]{ff, \\ff\_if\_known} & 120 \\
\hline
\multirow{3}{*}{Aggregative} & \makecell[l]{Strict +\\complete} & Agent count query & 50, 100 & ff, MCQ & 120 \\
 & \makecell[l]{Strict +\\incomplete} & Agent count query & 50, 100 & ff, MCQ & 120 \\
 & \makecell[l]{Ties +\\complete} & Agent count query & 50, 100 & ff & 60 \\
\hline
\multirow{2}{*}{\makecell[l]{Partial\\order}} & \makecell[l]{Pairwise\\(Format~1)} & \makecell[l]{Comparable pair,\\incomparable pair} & \makecell[l]{100 nodes\\+ \{50, 80, 150\}\\edges} & \makecell[l]{ff, \\ff\_if\_known,\\MCQ} & 420 \\
 & \makecell[l]{Query-answer\\(Format~2)} & \makecell[l]{Comparable pair,\\incomparable pair} & \makecell[l]{30 nodes\\+ \{30, 50\} edges} & \makecell[l]{ff,\\ ff\_if\_known,\\MCQ} & 120 \\
\hline
\multicolumn{5}{l}{\textbf{Total}} & \textbf{3360} \\
\hline
\end{tabular}}
\end{table}

The complete breakdown of the types of questions asked for each task, preference-type, and size, are provided in \Cref{tab:pref_benchmark}. Total count of 3360 prompts includes different prompting format. The unique number of prompts, not multiplying by format, is 1740.

\subsection{Algorithmic Reasoning Tasks}
\label{sec:algo_benchmark}
 
\subsubsection{Instance Generation}
\label{sec:instance_gen}
 
\textbf{Preference sampling.}
Strict and complete preferences are sampled under the impartial culture model, with each agent's ranking drawn uniformly at random from all permutations of $A$. Complete preferences with ties are derived by post-processing: adjacent items in a strict ranking are merged into a tie-group independently with probability~$p$ ($p{=}0.3$ for house allocation; $p{=}0.2$ for stable matching). Strict and incomplete preferences are obtained by truncating each agent's ranking to a length drawn uniformly from $[\lfloor 0.5n \rfloor,\, n]$. An agent is never assigned an alternative it has not ranked, and is left unmatched instead, following the standard convention that unranked alternatives are unacceptable. This is what keeps each solution concept well defined, so that an infeasible instance is infeasible because of the instance rather than because of how the truncation is read.
 
\textbf{Rejection sampling with distinctness constraints.}
Instances are selected via rejection sampling from the stream of randomly generated profiles. Every accepted instance must satisfy a \emph{distinctness constraint}: the optimal outcome for each solution concept evaluated in that setting must be distinct from every other, and no outcome satisfying one concept may simultaneously satisfy another. This ensures that correct model responses can be attributed unambiguously to the target notion, ruling out cases where a model succeeds by accidentally satisfying an easier notion.
 
\subsubsection{Benchmark Design}
\label{sec:benchmark_composition}
 
For each combination of domain, preference type, and size $n \in \{10, 30\}$, we collect 30 instances satisfying the distinctness constraint. Each instance gives rise to two types of tasks: \textbf{generation}, where the model computes a solution from scratch given the preference profile, and \textbf{selection}, where it chooses among a set of presented candidates. Generation prompts include a definition of the target notion and, by default, an instruction that the model may return $\{\}$ (i.e. an empty solution) if it determines no valid solution exists. Selection prompts present one candidate per solution concept evaluated in that setting, together with one or two \emph{contaminated} candidates obtained by randomly swapping one or two assignments in the canonical algorithm's output, with options in uniformly random order. All models are evaluated on the same instances in a zero-shot, single-turn setting at their default temperature.
 
The benchmark is structured around two components reflecting a natural partition of the solution concepts.
  
\textbf{Strict preferences.}
Under strict preferences, all solution concepts of interest are guaranteed to exist for every instance. For each applicable solution concept, one generation prompt is produced per instance; a further \emph{unspecified} generation prompt is included in which no notion or definition is provided and the model is simply asked to compute ``a matching'' or ``an allocation.'' This condition tests whether models default to the canonical algorithm in the absence of an explicit instruction. One selection task is also produced per instance, asking the model to select the option it ``prefers the most'' from the candidate set without naming a target notion; this tests whether models' default preferences align with canonically characterised solutions.
 
We evaluate house allocation, object allocation, and stable matching under strict and complete preferences as the primary settings. We additionally evaluate house allocation under strict and incomplete preferences to examine whether incompleteness affects performance; since this involves a single domain with a narrower set of notions, we treat it as a secondary setting.
 
\textbf{Preferences with ties.}
We evaluate house allocation and stable matching under complete preferences with ties.\footnote{Given the similarity between house allocation and object allocation, we do not separately evaluate object allocation with ties.} 
For certain notions such as weak core, weak stability, and the welfare and rank-based notions, the algorithms required are shared with the strict setting (e.g. TTC for weak core, DA for weak stability, Hungarian algorithm for utilitarian welfare, etc.). This enables a direct comparison of performance across preference types. 
Table~\ref{tab:benchmark_composition} summarises all settings and prompt counts for these standard tasks.
 
Ties additionally involve three solution concepts not guaranteed to exist for every instance: the strict core (house allocation), and strong and super stability (stable matching). These are excluded from the standard task structure above and are instead evaluated under a dedicated experimental design. For each notion, we collect 30 \emph{feasible} instances (where a valid solution exists, drawn from the standard pool) and 30 \emph{infeasible} instances (where no solution satisfying the notion exists).\footnote{Infeasible instances for super stability are constructed to admit a strongly stable solution, to test whether models conflate the two concepts.} Each instance receives four prompt variants forming a $2 \times 2$ design:
 
\begin{itemize}
    \item \textbf{Generation with and without escape hatch.} By default the model is instructed it may return $\{\}$ if no solution exists; in the ablation this instruction is removed. The latter tests whether models generate overconfident solutions when no valid one exists, or express infeasibility through other means (e.g.\ returning \texttt{None}, populating the output dictionary with \texttt{"None"} values, or issuing a natural-language disclaimer).
    \item \textbf{Selection with and without NOTA.} We vary whether a ``None of the Above'' (NOTA) option is included in the candidate set. The NOTA condition tests whether models correctly identify infeasibility when given an explicit escape; the no-NOTA condition tests whether they select a spurious solution under pressure.
\end{itemize}
 
This yields $2~\text{(feasibility)} \times 2~\text{(task type)} \times 2~\text{(escape hatch / NOTA)} \times 3~\text{(notions)} \times 30~\text{(instances)} = 720$ additional prompts per size.
 
\textbf{Total.}
The benchmark comprises \textbf{1,830 prompts per size} $n$, consisting of 1,110 from the standard settings in Table~\ref{tab:benchmark_composition} and 720 from the harder-notions experiment, for a total of \textbf{3,660 prompts} across both sizes $n \in \{10, 30\}$. Each prompt is answered independently by all four models, yielding \textbf{14,640 model responses} in total.

\begin{table}[t]
\centering
\caption{Benchmark composition for standard tasks (solution concepts guaranteed to exist). Each setting uses 30 instances. Generation includes one prompt per listed notion (each with the $\{\}$ instruction by default) plus one \emph{unspecified} prompt; selection asks the model to choose its most preferred candidate without specifying a target notion. $n_\text{gen}$ and $n_\text{sel}$ denote prompts per instance; total $= 30 \times (n_\text{gen} + n_\text{sel})$. Harder notions under preferences with ties (strict core; strong and super stability) are excluded from this table and evaluated separately. $\dagger$~Treated as a secondary setting. $\ddagger$~PO for house allocation with ties is verified via strong Pareto-optimality under weak preferences.}
\label{tab:benchmark_composition}
\small
\setlength{\tabcolsep}{5pt}
\begin{tabular}{llp{5cm}ccc}
\toprule
\textbf{Domain} & \textbf{Pref.\ type} & \textbf{Notions (generation)} & $n_\text{gen}$ & $n_\text{sel}$ & \textbf{Prompts} \\
\midrule
\multirow{3}{*}{House allocation}
  & Strict + complete             & Core, PO, UW, EW, RM                            & 6 & 1 & 210 \\
  & Strict + incomplete$^\dagger$ & Core, RM, MCPO                                  & 4 & 1 & 150 \\
  & Ties + complete               & Weak core, PO$^\ddagger$, UW, EW, RM            & 6 & 1 & 210 \\[3pt]
Object allocation
  & Strict + complete             & PO, UW, EW, RM                                  & 5 & 1 & 180 \\[3pt]
\multirow{2}{*}{Stable matching}
  & Strict + complete             & Stable, UW, EW, RM                              & 5 & 1 & 180 \\
  & Ties + complete               & Weakly stable, UW, EW, RM                       & 5 & 1 & 180 \\
\midrule
\multicolumn{5}{l}{Subtotal per size $n$} & \textbf{1,110} \\
\bottomrule
\end{tabular}
\end{table}

\section{Model and Inference Details}
\label{app:model_details}

We evaluate four frontier models, listed in
Table~\ref{tab:model_specs} with their exact identifiers, hosting
providers, and per-call inference settings. All four are accessed
through their providers' hosted APIs at decoding time, with no
fine-tuning. Temperature is fixed at $1.0$ across all models; we
do not override top-$p$, top-$k$, or frequency / presence
penalties. The full inference scripts (auth-key handling,
retry / resume logic, code-assisted tool wiring) are released
alongside this paper.

\textbf{Temperature.} The two frontier reasoning models, GPT-5.2 and Gemini-2.5-Pro, do not expose a temperature control and run at a fixed provider default, so a temperature-0 condition is not available for them. Claude-4.5-Sonnet and OSS-120B do expose it, and for these two we re-ran a 40-instance subset of the hard-notion generation task at $n=10$, drawn equally from feasible and infeasible instances of each notion, at temperature 0.

The results are essentially unchanged. Overall accuracy on this subset is 0.00 for Claude-4.5-Sonnet at temperature 1.0 and 0.03 at temperature 0, and 0.03 at both temperatures for OSS-120B, and the rate at which each model correctly identifies an infeasible instance does not move between the two temperatures. The undetermined preference queries behave the same way: on 60 RS-incomparable bundle comparisons, both models abstain on none of them at either temperature, so the lexicographic default is deterministic as well. Temperature is not what drives these failures, and they are not sampling noise.

\begin{table}[h]
\centering
\small
\caption{Inference settings for the four evaluated models.
``Max tokens'' lists the per-call output budget on the generation
task and (where it differs) the selection task. ``Reasoning
effort'' is the provider-side knob exposed by the OpenAI
Responses API; the other three providers do not expose an
analogue.}
\label{tab:model_specs}
\setlength{\tabcolsep}{4pt}
\resizebox{\textwidth}{!}{
\begin{tabular}{@{}l p{2.7cm} p{3.6cm} p{2cm} l p{2.2cm}@{}}
\toprule
\textbf{Label} & \textbf{Provider / API} & \textbf{Model identifier}
& \textbf{Max tokens} & \textbf{Timeout} & \textbf{Reasoning effort} \\
\midrule
GPT-5.2
  & OpenAI \newline (Responses API)
  & \texttt{gpt-5.2}
  & $30$k (gen) \newline $40$k (sel)
  & $20$ min
  & medium \newline (concise summary) \\[0.4em]
Claude Sonnet 4.5
  & Anthropic \newline (Messages API)
  & \texttt{claude-sonnet-4-5-} \newline \texttt{20250929}
  & $30$k (gen) \newline $15$k (sel)
  & $20$ min
  & --- \\[0.4em]
Gemini 2.5 Pro
  & Google \newline (\texttt{google.genai})
  & \texttt{gemini-2.5-pro}
  & provider \newline default
  & provider \newline default
  & --- \\[0.4em]
GPT-OSS 120B
  & Groq \newline (chat-completions)
  & \texttt{openai/gpt-oss-120b}
  & $20$k (gen) \newline $15$k (sel)
  & provider \newline default
  & --- \\
\bottomrule
\end{tabular}}
\end{table}

\section{Additional Results: Preference Reasoning}\label{app:pref_results_appendix}

This appendix provides the per-model, per-condition results underlying the findings about preference reasoning. In each subsequent section, we present the accuracy of models on each task, and break down the assumptions they make leading to determined responses in undetermined problems.

\subsection{Additional Figures}

\begin{figure}
    \centering
    \includegraphics[width=0.8\linewidth]{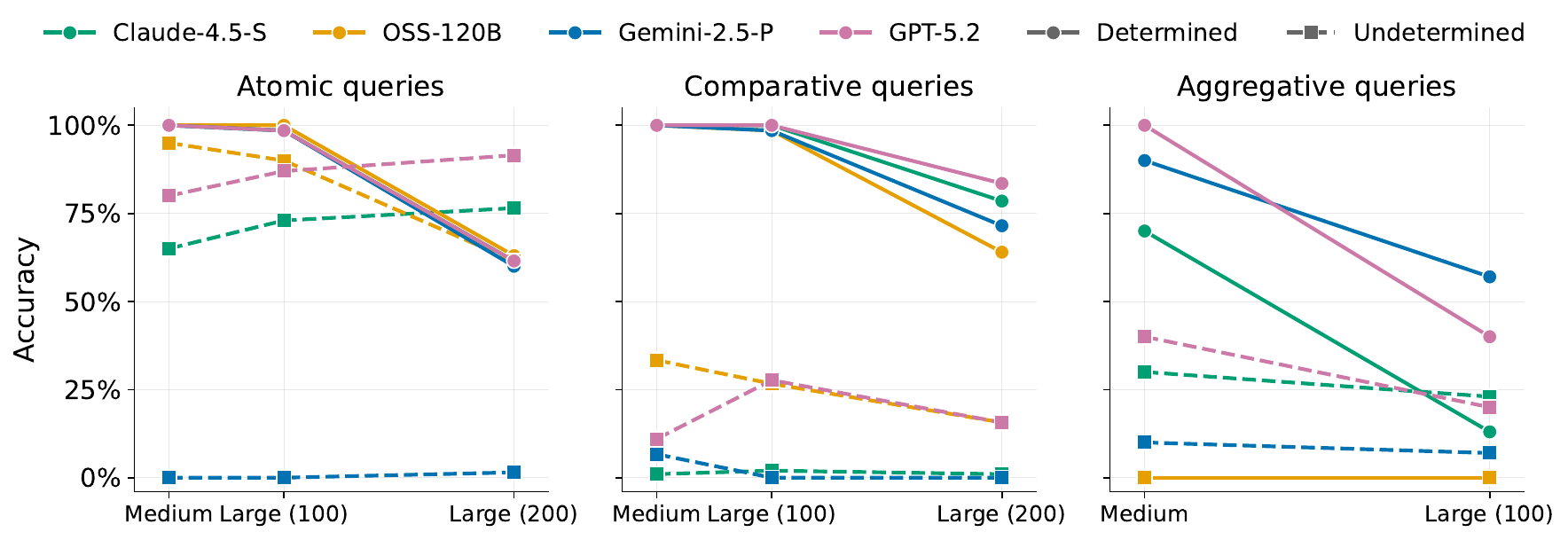}
    \caption{The models' accuracy degrades with increasing problem size in both determined and undetermined queries.}
    \label{fig:pref_size_scaling_3panel}
\end{figure}

In Figure \ref{fig:pref_size_scaling_3panel}, we show that as the preference profile size increases, the accuracy of models decrease significantly on determined queries. For atomic and comparative queries, models remain almost perfect accuracy up to size 100, but begins to fail at size 200. For aggregative query, the threshold is much lower as model errors increase significantly at size 100, presumably related to the increase of both number of agents and length of preference list leading to quadratic increase in the profile. The effect of sizes on undetermined problems is less pronounced and model variations exist, reflecting their diverse behavior pattern, either affirming incomparability with greater confidence or becoming more prone towards assumptions as input complexity increases. A breakdown of model performance on different tasks is provided in subsequent sections.

\begin{figure}
    \centering
    \includegraphics[width=\linewidth]{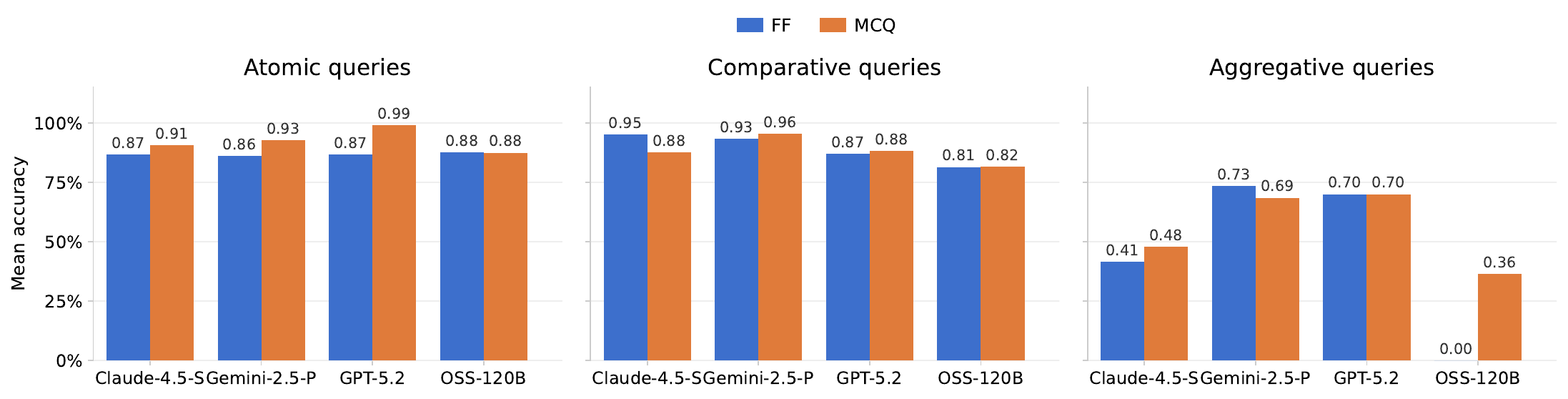}
    \caption{Comparing LLMs' performance with different prompting formats on determined preference queries, split by task type. The most notable effect of format is \OSS ~on aggregative tasks. While completely failing to perform accurate aggregation in free-flow format, it selects the correct count in MCQ for a certain portion, yet still significantly trailing other 3 models}
    \label{fig:query_fmt_comp_det}
\end{figure}

\subsection{Detailed results of reasoning over total-order preference profiles}
\label{app:f1_detail}

\textbf{Atomic and comparative queries}
Table~\ref{tab:atomic_comp_det_ff} reports atomic and comparative queries on strict and complete preferences, which have a determined answer.

Accuracy on item/position queries are high up to n=100, but significantly decrease as preference profile hit n=200. Rate of decline varies across 4 models. 

Accuracy on pairwise comparison and ranking also decreases. Pairwise comparison is less affected than atomic query (presumably due to relative rather than precise location required), while the accuracy is lower in the ranking task, due to the accumulation of pairwise comparisons needed to provide the ranking of 4 items.

\begin{table}[h]
\centering
\caption{Accuracy on determined atomic and comparative queries, under free-form prompt.}
\label{tab:atomic_comp_det_ff}
\small
\setlength{\tabcolsep}{5pt}
\begin{tabular}{l ccc ccc ccc ccc}
\toprule
& \multicolumn{3}{c}{\textbf{Item query}} & \multicolumn{3}{c}{\textbf{Position query}} & \multicolumn{3}{c}{\textbf{Comparison}} &\multicolumn{3}{c}{\textbf{Ranking}} \\
\cmidrule(lr){2-4}\cmidrule(lr){5-7}\cmidrule(lr){8-10}\cmidrule(lr){11-13}
\textbf{Model} & 50 & 100 & 200 & 50 & 100 & 200 & 50 & 100 & 200 & 50 & 100 & 200 \\
\midrule
Gemini-2.5-P    & 1.00 & 0.967 & 0.50  & 1.00 & 1.00  & 0.70  & 1.00 & 1.00 & 0.80 & 1.00 & 0.967 & 0.633  \\
OSS-120B        & 1.00 & 1.00  & 0.68 & 1.00 & 1.00  & 0.59 & 1.00 & 1.00 & 0.83 & 1.00 & 0.967 & 0.45 \\
Claude-4.5-S    & 1.00 & 0.967 & 0.50 & 1.00 & 1.00  & 0.73 & 1.00 & 1.00 & 0.90 & 1.00 & 1.00  & 0.80 \\
GPT-5.2         & 1.00 & 0.967 & 0.60 & 1.00 & 1.00  & 0.633 & 1.00 & 1.00 & 1.00 & 1.00 & 1.00  & 0.667 \\
\bottomrule
\end{tabular}
\end{table}

Table~\ref{tab:single-agent-incomplete} reports accuracy on undetermined atomic and comparative queries based on strict and incomplete preferences.

On atomic queries, Gemini most often fails to acknowledge indeterminacy. It invokes assumptions that lead to a definite output, including assumed typographical error in query, and manual completion of the incomplete preference list, by filling the missing items in lexical order behind the listed ones.

On comparative queries, when both items to be compared are not listed, models make error by choose one definite answer, using assumption such as lexical ordering. This accumulates in the ranking task, where all models most often return a full ranking of 4 items, even when only a partial ranking can
be determined, as 2 of the items are not listed in the preference profile.

\begin{table}[h]
\centering
\caption{Accuracy on undetermined atomic and comparison tasks, with "ff + if known" prompts. Correct response is: for item query, ``unknown'' (queried item absent from profile); for position query, ``unknown'' (queried position out of bounds); for comparison, ``unknown'' (both items not listed); for ranking, return only the listed items. A score of~0 indicates the model never gives the correct response.}
\label{tab:single-agent-incomplete}
\small
\setlength{\tabcolsep}{6pt}
\begin{tabular}{l ccc ccc ccc ccc}
\toprule
& \multicolumn{3}{c}{\textbf{Item query}} & \multicolumn{3}{c}{\textbf{Position query}} & \multicolumn{3}{c}{\textbf{Comparison}} & \multicolumn{3}{c}{\textbf{Ranking}} \\
\cmidrule(lr){2-4}\cmidrule(lr){5-7}\cmidrule(lr){8-10}\cmidrule(lr){11-13}
\textbf{Model} & 50 & 100 & 200 & 50 & 100 & 200 & 50 & 100 & 200 & 50 & 100 & 200 \\
\midrule
Gemini-2.5-P    & 0.50 & 0.30 & 0.00 & 0.63 & 0.40 & 0.23 & 0.63 & 0.50 & 0.37 & 0.17 & 0.20 & 0.00 \\
OSS-120B        & 1.00 & 0.93 & 0.87 & 1.00 & 1.00 & 1.00 & 1.00 & 0.57 & 0.22 & 0.00 & 0.00 & 0.00 \\
Claude-4.5-S    & 0.87 & 0.80 & 0.93 & 1.00 & 1.00 & 0.83 & 0.70 & 0.90 & 0.47 & 0.00 & 0.00 & 0.00 \\
GPT-5.2         & 1.00 & 1.00 & 1.00 & 1.00 & 1.00 & 1.00 & 0.43 & 1.00 & 1.00 & 0.00 & 0.00 & 0.00 \\
\bottomrule
\end{tabular}
\end{table}

\textbf{Aggregative queries}
Table~\ref{tab:aggregation} reports aggregation accuracy across the three preference types. The parenthetical numbers under incomplete profiles give the fraction of \emph{incorrect} responses that match the count where agents  ranking the preferred item in question but not the other one are included (an implicit and unwarranted disambiguation). OSS-120B fails entirely across all conditions; the other three models are reasonable at $n=50$ on strict-and-complete profiles but degrade sharply at $n=100$ and on tie-containing profiles. The mean absolute errors of the incorrect responses are shown in Table~\ref{tab:aggregation:error}. Errors for incomplete preferences are not calculated, since the errors are not only caused by counting, but also induced by assumption on preference over item not being listed.

\begin{table}[h]
\centering
\caption{Accuracy on the aggregative task. For incomplete preferences, values in parentheses give the fraction of \emph{incorrect} responses matching the assumption made above.}
\label{tab:aggregation}
\small
\setlength{\tabcolsep}{5pt}
\begin{tabular}{l cc cc cc}
\toprule
& \multicolumn{2}{c}{\textbf{Strict + complete}} & \multicolumn{2}{c}{\textbf{Strict + incomplete}} & \multicolumn{2}{c}{\textbf{Complete + ties}} \\
\cmidrule(lr){2-3}\cmidrule(lr){4-5}\cmidrule(lr){6-7}
\textbf{Model} & Size 50 & Size 100 & Size 50 & Size 100 & Size 50 & Size 100 \\
\midrule
Gemini-2.5-P    & 0.90 & 0.57 & 0.10 (0.89) & 0.17 (0.52) & 0.70 & 0.50 \\
OSS-120B        & 0.00 & 0.00 & 0.00 (---)  & 0.00 (---)  & 0.00 & 0.00 \\
Claude-4.5-S    & 0.70 & 0.13 & 0.30 (0.43) & 0.23 (0.17) & 0.30 & 0.10 \\
GPT-5.2         & 1.00 & 0.40 & 0.40 (0.83) & 0.40 (0.33) & 0.80 & 0.33 \\
\bottomrule
\end{tabular}
\end{table}

\begin{table}[h]
\centering
\caption{Mean absolute error of incorrect responses for the aggregation task.}
\label{tab:aggregation:error}
\small
\setlength{\tabcolsep}{5pt}
\begin{tabular}{l cc cc}
\toprule
& \multicolumn{2}{c}{\textbf{Strict + complete}} & \multicolumn{2}{c}{\textbf{Complete + ties}} \\
\cmidrule(lr){2-3}\cmidrule(lr){4-5}
\textbf{Model} & Size 50 & Size 100 & Size 50 & Size 100 \\
\midrule
Gemini-2.5-P    & 1.00 & 4.54 & 1.08 & 3.93 \\
OSS-120B        & 4.60 & 27.7 & 9.60 & 32.5 \\
Claude-4.5-S    & 1.80 & 2.62 & 2.65 & 3.22 \\
GPT-5.2         & 0.00 & 2.83 & 1.13 & 2.69 \\
\bottomrule
\end{tabular}
\end{table}

\textbf{OSS-120B on aggregation: a note on catastrophic failure.}
OSS-120B returns 0.00 accuracy on aggregation across all preference types and both sizes. Inspection of responses shows that the model returns counts that differ from the ground truth by large margins, rather than failing in a structured way. We treat this as a per-model anomaly and exclude OSS-120B from finding-level claims about aggregation that depend on the structured-error pattern.

\subsection{Special undetermined queries: detailed bundle comparison and top-$k$ results}
\label{app:f2_detail}

\textbf{Bundle comparison, all preference types and sizes.}
Table~\ref{tab:bundle} gives the full per-model breakdown of the lexicographic-ordering finding. The values in parentheses give the fraction of RS-incomparable responses where the model applied lexicographic ordering by highest-ranked element. Across all twelve (model $\times$ preference type) cells, the lexicographic share is 80\% or higher; the only deviation from pure lexicographic behaviour is Claude-4.5-S, which occasionally uses the sum of item ranks instead, which can yield the opposite conclusion.

 Our results score an undetermined response correct only when it abstains. A model that commits to an answer while naming the assumption behind it is arguably doing something different from one that commits silently, so it is worth asking how much of the gap survives if the former is credited too. We sorted every RS-incomparable response into four kinds: an explicit abstention, a completion whose lexicographic rule is stated as an assumption, a silent lexicographic completion, and a non-lexicographic answer.

Table~\ref{tab:bundle_rescore} reports the strict scoring (abstention only) against the lenient scoring (abstention or a flagged assumption). The lenient scoring raises accuracy by at most 0.17, and every model stays far below its accuracy of roughly 1.00 on the matched determined comparisons. Flagged assumptions are rare, at most 0.17 of responses, while silent lexicographic completion accounts for 0.70 to 1.00. It is this silent completion, rather than flagging that goes uncredited, that the epistemic results measure.

Additionally, we test the bundle comparison tasks on two more schemes. Firstly, we increase the size of bundles to be compared. We randomly generate one bundle, and substitute several items within it with both more preferred and less preferred items, this creates a pair of equal-sized bundles that are RS-incomparable. Secondly, we create bundle pairs with unequal sizes. We select a single item that ranked relatively highly to form a one-item bundle, and then select multiple items less preferred to it to form the second bundle, with varying size up to half of the preference list (over 20 items). In both cases, models keep the lexicographic bias when asked to select the preferred bundle. Results are in Table ~\ref{tab:bundle_additional}.

\begin{table}[h]
\centering
\caption{Bundle comparison accuracy ($n=50$). For RS-comparable pairs (a definite answer exists), accuracy is reported. For RS-incomparable pairs (no determinate answer exists), the main value is the fraction of responses that correctly select ``can't decide,'' with the fraction applying lexicographic ordering by highest-ranked element in parentheses.}
\label{tab:bundle}
\small
\setlength{\tabcolsep}{5pt}
\begin{tabular}{l cc cc cc}
\toprule
& \multicolumn{2}{c}{\textbf{Strict + complete}} & \multicolumn{2}{c}{\textbf{Strict + incomplete}} & \multicolumn{2}{c}{\textbf{Complete with ties}} \\
\cmidrule(lr){2-3}\cmidrule(lr){4-5}\cmidrule(lr){6-7}
\textbf{Model} & RS-comp. & RS-incomp. & RS-comp. & RS-incomp. & RS-comp. & RS-incomp. \\
\midrule
Gemini-2.5-P    & 1.00 & 0.00 (1.00) & 1.00 & 0.00 (1.00) & 1.00 & 0.20 (0.80) \\
OSS-120B        & 1.00 & 0.00 (1.00) & 0.85 & 0.00 (1.00) & 1.00 & 0.00 (1.00) \\
Claude-4.5-S    & 1.00 & 0.00 (0.90) & 1.00 & 0.00 (0.90) & 1.00 & 0.10 (0.90) \\
GPT-5.2         & 1.00 & 0.17 (0.83) & 1.00 & 0.07 (0.93) & 1.00 & 0.10 (0.90) \\
\bottomrule
\end{tabular}
\end{table}

\begin{table}[h]
\centering
\caption{Bundle comparison scoring for RS-incomparable pairs. In the strict scoring, only ``can't decide'' is accepted. In the lenient scoring, returning a lexicographically preferred bundle while explicitly stating the rule as an ``assumption'' is allowed}
\label{tab:bundle_rescore}
\small
\setlength{\tabcolsep}{5pt}
\begin{tabular}{l cccc}
\toprule
\textbf{scoring} & \textbf{Gemini-2.5-P} & \textbf{OSS-120B} & \textbf{Claude-4.5-S} & \textbf{GPT-5.2} \\
\midrule
Abstention only (strict)                 & 0.10 & 0.00 & 0.03 & 0.17 \\
Plus flagged assumption (lenient)        & 0.23 & 0.00 & 0.20 & 0.20 \\
\bottomrule
\end{tabular}
\end{table}

\begin{table}[h]
\centering
\caption{Additional results on RS-incomparable bundles. The main value is the fraction of responses that correctly indicate the comparison is undetermined (``can't decide''), with the fraction using lexicographic ordering in parentheses.}
\label{tab:bundle_additional}
\small
\setlength{\tabcolsep}{5pt}
\begin{tabular}{l cc}
\toprule
model & increased bundle size & one vs many \\
\midrule
Gemini-2.5-P    & 0.00 (1.00) & 0.00 (1.00) \\
OSS-120B        & 0.00 (1.00) & 0.00 (1.00) \\
Claude-4.5-S    & 0.00 (0.80) & 0.00 (1.00) \\
GPT-5.2         & 0.00 (1.00) & 0.00 (1.00) \\
\bottomrule
\end{tabular}
\end{table}

\textbf{Top-$k$ queries (complete preference with ties): full prompt-variant results.}
A unique problem for preference with ties involves indeterminacy due to tie-breaking. We ask whether an item belongs to the top-$k$ choices of the agent. The answer is determined, when exactly $k$ choices can be uniquely selected from the preference list, or when the queried item is not within a tie. Indeterminacy arises when the ”$k$th” position lies within a tie, and the queried item is part of the tie, since there is no definitive way of identifying the ``top-$k$'' items to judge whether the agent belongs to them. Table~\ref{tab:top-k} shows the way a top-$k$ membership question is asked changes both the answer the model gives and the listing behaviour underneath it. Prompt~1 asks this directly with ``yes / no / uncertain'' as the available answers. Prompt~2 instead asks the model to list the agent's top-$k$ items without mentioning $A$; correctness is then about whether the response breaks the tie to return only $k$ items, or returns the entire tie group. Prompt~3 combines the two: first asks for the top-$k$ list, then asks whether $A$ belongs, with only ``yes / no'' available.

\begin{table}[h]
\centering
\caption{Top-$k$ selection task under preferences with ties. $k$ is chosen so that the $k$th position falls within a tie group, and item~$A$ is a member of that tie. Prompt~1 directly asks whether $A$ is in the top-$k$; Prompt~2 asks for the top-$k$ list only; Prompt~3 asks for the list and then asks whether $A$ belongs.}
\label{tab:top-k}
\small
\setlength{\tabcolsep}{5pt}
\begin{tabular}{l cc cc cc}
\toprule
& \multicolumn{2}{c}{\textbf{Prompt 1}} & \multicolumn{2}{c}{\textbf{Prompt 2}} & \multicolumn{2}{c}{\textbf{Prompt 3}} \\
\cmidrule(lr){2-3}\cmidrule(lr){4-5}\cmidrule(lr){6-7}
\textbf{Model} & uncertain & yes/no & keeps tie & breaks tie & yes & no \\
\midrule
Gemini-2.5-P    & 0.00 & 0.90/0.10 & 0.00 & 1.00 & 1.00 & 0.00 \\
OSS-120B        & 0.30 & 0.70/0.00 & 0.10 & 0.90 & 0.90 & 0.10 \\
Claude-4.5-S    & 0.40 & 0.60/0.00 & 0.50 & 0.50 & 1.00 & 0.00 \\
GPT-5.2         & 0.63 & 0.37/0.00 & 0.60 & 0.40 & 0.90 & 0.10 \\
\bottomrule
\end{tabular}
\end{table}

Two contrasts in the table are worth surfacing. First, the Prompt~1 ``uncertain'' column is the only place the data set offers a calibrated abstention rate when the response options include ``uncertain'': Gemini never picks it (consistent with its over-commitment in other tasks), whereas GPT-5.2 picks it 63\% of the time. Removing the uncertain option in Prompt~3 does not redistribute these uncertainties evenly between yes and no; almost all are absorbed into yes. Second, the Prompt~2 ``keeps tie'' column shows a wide spread across models (0\% for Gemini, 60\% for GPT) when no specific item is mentioned, which then collapses to 90--100\% across the board in Prompt~3 once a specific item is named. Both contrasts point to the same underlying observation: surface phrasing reshapes the model's response on a fixed underlying ambiguity. Whether this should be read as a calibration failure (the model has no stable answer to give) or as a context-sensitivity feature (the model adapts to the apparent intent of the prompt) depends on the deployment context; we report the data without taking a position.

\subsection{Partial-order pairwise comparison: detailed results}
\label{app:f3_detail}

\textbf{Partial-order queries on comparable pairs.}
Table~\ref{tab:partial_order_pairwise_comparable} reports the per-model breakdown of accuracy on comparable partial-order pairs along both complexity dimensions: edge count varied at fixed chain length~9, and chain length varied at fixed 150 edges. Gemini-2.5-P and Claude-4.5-S handle long chains in dense graphs reliably; OSS-120B and GPT-5.2 degrade in both dimensions, sometimes failing to find chains they had earlier identified at lower density.

\begin{table}[h]
\centering
\caption{Accuracy on comparable item pairs in partial orders ($n=100$ items). Left block: chain length fixed at~9, edge count varied. Right block: edge count fixed at 150, chain length varied.}
\label{tab:partial_order_pairwise_comparable}
\small
\setlength{\tabcolsep}{4pt}
\begin{tabular}{l ccc ccc}
\toprule
 & \multicolumn{3}{c}{\textbf{Chain length = 9}} & \multicolumn{3}{c}{\textbf{Edges = 150}} \\
\cmidrule(lr){2-4}\cmidrule(lr){5-7}
\textbf{Model} & 50 edges & 80 edges & 150 edges & chain of 7 & chain of 5 & chain of 3 \\
\midrule
Gemini-2.5-P    & 1.00 & 1.00 & 1.00 & 1.00 & 1.00 & 1.00 \\
OSS-120B        & 1.00 & 1.00 & 0.23 & 0.50 & 0.73 & 1.00 \\
Claude-4.5-S    & 1.00 & 1.00 & 1.00 & 1.00 & 1.00 & 1.00 \\
GPT-5.2         & 1.00 & 0.93 & 0.17 & 0.47 & 0.80 & 1.00 \\
\bottomrule
\end{tabular}
\end{table}

\textbf{Incomparable pairs: free-flow vs.\ MCQ across graph densities.}
Table~\ref{tab:partial_order_pairwise_incomparable} gives the per-density performance of undetermined partial order queries. The dominant pattern: GPT-5.2 and OSS-120B acknowledge incomparability reliably even under free-flow; Gemini-2.5-P declines as density grows; Claude-4.5-S almost never acknowledges incomparability under free-flow, collapsing to 0\% at 150 edges. Switching to MCQ recovers near-perfect accuracy across all four models.

\begin{table}[h]
\centering
\caption{Accuracy on incomparable item pairs in partial orders ($n=100$ items), across graph densities and prompt formats.}
\label{tab:partial_order_pairwise_incomparable}
\small
\setlength{\tabcolsep}{4pt}
\begin{tabular}{l ccc ccc}
\toprule
& \multicolumn{3}{c}{\textbf{Free-flow}} & \multicolumn{3}{c}{\textbf{MCQ}} \\
\cmidrule(lr){2-4}\cmidrule(lr){5-7}
\textbf{Model} & 50 edges & 80 edges & 150 edges & 50 edges & 80 edges & 150 edges \\
\midrule
Gemini-2.5-P    & 0.80 & 0.63 & 0.50 & 1.00 & 1.00 & 1.00 \\
OSS-120B        & 1.00 & 1.00 & 1.00 & 1.00 & 1.00 & 1.00 \\
Claude-4.5-S    & 0.13 & 0.07 & 0.00 & 1.00 & 1.00 & 1.00 \\
GPT-5.2         & 1.00 & 0.97 & 0.97 & 1.00 & 1.00 & 1.00 \\
\bottomrule
\end{tabular}
\end{table}

When Claude-4.5-S and Gemini-2.5-P fail to acknowledge incomparability under free-flow, inspection of the reasoning traces reveals consistent patterns: assuming the user has made a typographical error in specifying the input and constructing an artificial chain to make the items comparable; selecting the item that appears anywhere in the input when the other does not; and inferring preference from the relative number of items each one dominates or is dominated by. These are structured assumptions of the same kind documented for bundle comparisons and aggregation. A breakdown of the assumptions is listed in Table~\ref{tab:partial_order_assumptions}.

\begin{table}[h]
\centering
\caption{Distribution of assumptions that models make to give a definite answer to the unanswerable partial order comparison task under free-flow prompt.}
\label{tab:partial_order_assumptions}
\small
\setlength{\tabcolsep}{4pt}
\begin{tabular}{l cc}
\toprule
\textbf{Model} & Gemini-2.5-P  & Claude-4.5-S \\
\midrule
Determining via relative preference   & 0.30 & 0.87 \\
Assuming typo in query or conditions & 0.30 & 0.00 \\
Acknowledging lack of chain yet still committing a choice    & 0.40 & 0.13 \\
\bottomrule
\end{tabular}
\end{table}

\textbf{Comparable pairs under MCQ: the cost of an explicit ``unknown'' option.}
Table~\ref{tab:partial_order_pairwise_comparable_mcq} reports the effect of MCQ prompt on determined partial order query, evaluated at chain length~9 with 150 edges over 100 nodes. On these inputs a chain in the input determines the answer, so the correct response is to identify the preferred item rather than abstain. Adding the MCQ ``not enough information to decide'' option leaves Gemini unaffected (1.00 in both formats) but degrades the other three models. Claude shows the largest drop, from 100\% under free-flow to 33\% under MCQ; OSS and GPT decline further from already-weak free-flow baselines. Together with Table~\ref{tab:partial_order_pairwise_incomparable}, the result is that the MCQ option moves models in opposite directions on the two question types: where Claude was over-committing (incomparable pairs), it is pulled back toward abstention; where it was answering correctly (comparable pairs), it is pulled away from a definite answer. 

\textbf{Discussion on the effect of prompt formats} The determined partial order query is the only preference reasoning task where we observe the model falsely indicating indeterminacy for a determined task, potentially due to the complexity of partial order reasoning. In reasoning with total order profiles, models almost always commit to a determined response, either correct or incorrect, even tending to commit when the problem is logically undetermined. As we implement alternative prompt formats such as MCQ to them, the performance does not change significantly for most determined tasks (Figure~\ref{fig:query_fmt_comp_det}),  while free-flow with "if-known" prompt and MCQ lead the models to acknowledge indeterminacy more often overall in the undetermined tasks, as we demonstrated in the main text (Figure~\ref{fig:pref_bars_combined}).

\begin{table}[h]
\centering
\caption{Accuracy on comparable item pairs in partial-order preferences ($n=100$ items, 150 edges, target chain length 9), under free-flow and MCQ formats.}
\label{tab:partial_order_pairwise_comparable_mcq}
\small
\setlength{\tabcolsep}{4pt}
\begin{tabular}{l cc}
\toprule
\textbf{Model} & \textbf{Free-flow} & \textbf{MCQ} \\
\midrule
Gemini-2.5-P    & 1.00 & 1.00 \\
OSS-120B        & 0.23 & 0.10 \\
Claude-4.5-S    & 1.00 & 0.33 \\
GPT-5.2         & 0.17 & 0.10 \\
\bottomrule
\end{tabular}
\end{table}

\textbf{Alternative partial-order representation (Format 2).}
Table~\ref{tab:partial_order_query_answer} reports results on a query-answer representation of partial orders, where each input data point specifies a subset of items and the most preferred item within that subset (rather than a pairwise comparison). This representation is informationally equivalent to a partial order but requires an inference step to recover the underlying pairwise structure. The same qualitative pattern as Format 1 holds: free-flow performance varies sharply across models, with Claude-4.5-S collapsing on incomparable pairs; MCQ recovers near-perfect accuracy across all models. The replication across representations supports the MCQ-recovers-abstention pattern not being an artifact of the input format.

\begin{table}[h]
\centering
\caption{Accuracy on incomparable item pairs in partial-order preferences using the query-answer representation ($n=30$ items), across graph densities and formats.}
\label{tab:partial_order_query_answer}
\small
\setlength{\tabcolsep}{4pt}
\begin{tabular}{l cc cc}
\toprule
& \multicolumn{2}{c}{\textbf{Free-flow}} & \multicolumn{2}{c}{\textbf{MCQ}} \\
\cmidrule(lr){2-3}\cmidrule(lr){4-5}
\textbf{Model} & 30 edges & 50 edges & 30 edges & 50 edges \\
\midrule
Gemini-2.5-P    & 0.50 & 0.20 & 1.00 & 1.00 \\
OSS-120B        & 1.00 & 1.00 & 1.00 & 1.00 \\
Claude-4.5-S    & 0.20 & 0.00 & 1.00 & 1.00 \\
GPT-5.2         & 1.00 & 1.00 & 1.00 & 1.00 \\
\bottomrule
\end{tabular}
\end{table}

\subsection{Alternative preference presentation: natural language rendering}
\label{app:nat_lang_pref}

To verify whether the model biases on undetermined queries are an artifact of the symbolic lists we use to represent preference profiles by default, we re-ran the same queries under three levels of natural-language rendering of the profile. To make the task as easy as possible, each prompt was also cut down to the queried agent's list alone, which is the only preference information the question needs.
\begin{itemize}
    \item Level 1 is the original symbolic list, for example ["C9", "C33", ...] given for agent A38.
    \item Level 2 states the same ranking in words, for example "A38 most prefers C9, then C33, and so on".
    \item Level 3 renders the items as real-world objects and narrates a person choosing a meal, for example "They like pizza the most, then sushi, then ...", with the question rewritten as "Which bundle does the person prefer between \{sushi, tiramisu\} and \{burgers, jambalaya\}?".
\end{itemize}

 Alongside the bundle task we include an undetermined atomic query, asking about preference of an item absent from the preference list.

We generated 60 undetermined bundle comparisons with the construction rule of \Cref{sec:pref_benchmark}, together with 60 matched determined comparisons in which one bundle clearly dominates, and ran both on GPT-5.2 and OSS-120B.

As shown in Table~\ref{tab:nat_lang_pref}, correct abstention on the bundle task is at or near 0.00 at all three levels for both models, while the matched determined comparisons are answered correctly in every case. The near-zero abstention is therefore a genuine failure to recognize incomparability, not low ability on the task, and natural-language rendering does not recover the correct ``can't decide''.

What the models do recognize depends on the source of the indeterminacy. They almost never flag that two bundles of ranked items are formally incomparable, but they flag an unknown answer far more often when the queried item was never ranked. The natural-language levels do not change either behaviour, so the lexicographic default is not an artifact of how the list is written.

\begin{table}[h]
\centering
\caption{Accuracy on undetermined problems, with different levels of natural language representation of preferences. Level 1: original symbolic list  ({"A38": ["C9", ...]}). Level 2: same items, natural sentences. Level 3: items replaced by real dishes, narrated as a person choosing a meal. Each cell is the fraction of 60 responses that correctly indicate indeterminacy. The last row is the matched determined control, where one bundle dominates and a definite answer exists.}
\label{tab:nat_lang_pref}
\small
\setlength{\tabcolsep}{4pt}
\begin{tabular}{ll ccc}
\toprule
\textbf{Task} & \textbf{Model} & \textbf{Level 1} & \textbf{Level 2} & \textbf{Level 3} \\
\midrule
RS-incomparable bundle        & GPT-5.2  & 0.00 & 0.00 & 0.00 \\
RS-incomparable bundle        & OSS-120B & 0.00 & 0.00 & 0.02 \\
unranked item                 & GPT-5.2  & 0.48 & 0.50 & 0.42 \\
unranked item                 & OSS-120B & 0.50 & 0.53 & 0.33 \\
determined bundle (control) & both & 1.00 & 1.00 & 1.00 \\
\bottomrule
\end{tabular}
\end{table}

\section{Additional Results: Algorithmic Reasoning over Preferences}\label{app:algo_results_appendix}

\subsection{Performance Differs Across Solution Concepts}

\begin{figure}[t]
    \centering
    \includegraphics[width=0.9\linewidth]{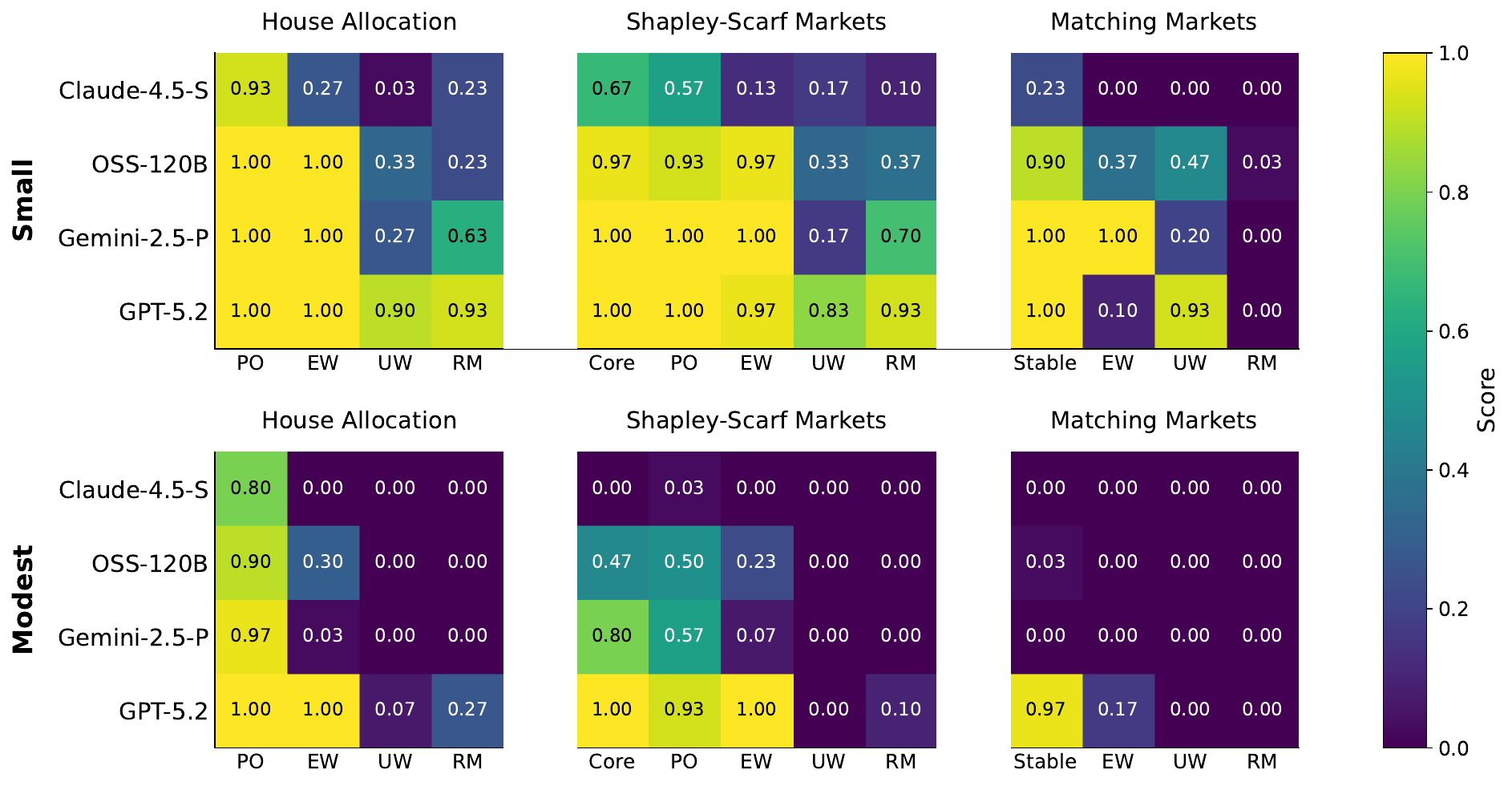}
    \caption{Performance of LLMs in terms of achieving specific notions across all three domains under strict and complete preferences, for instance sizes Small and Modest. Each cell is the fraction of instances where the model's output satisfies the target notion.}
    \label{fig:all_both_strict_complete_4}
\end{figure}

As shown in \Cref{fig:all_both_strict_complete_4}, all models achieve higher and more consistent accuracy on notions computed by canonical algorithms (core/TTC, PO/SD, stability/DA) than on non-canonical ones (UW, EW, RM). At $n=30$, most models continue to perform well on canonical notions, while accuracy on UW and RM falls sharply, approaching zero for all models except GPT-5.2.
EW occupies an intermediate position, with GPT-5.2 and Gemini-2.5-P maintaining reasonable accuracy on house allocation but all models failing in stable matching.
The model hierarchy, unlike in \Cref{sec:pref_results}, is consistent: GPT-5.2 performs best overall, Gemini-2.5-P and OSS-120B show mixed relative performance across domains and notions, and Claude-4.5-S performs worst. 

\subsection{Robustness to Preference Structure}
\label{subsec:robustness}

\textbf{Performance is robust to incompleteness.}

\begin{figure}
    \centering
    \includegraphics[width=0.8\linewidth]{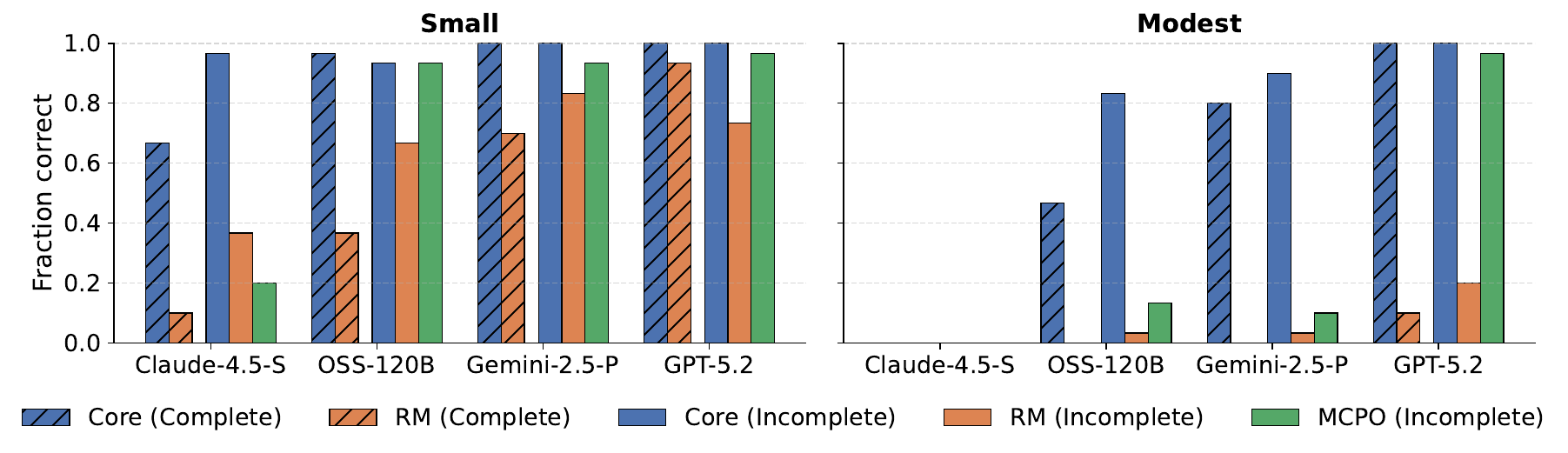}
    \caption{Performance on house allocation under strict and complete vs.\ strict and incomplete preferences.}
    \label{fig:strict_vs_incomplete}
\end{figure}

For notions evaluated under both strict and incomplete preferences, performance does not decrease when preference lists are truncated (\Cref{fig:strict_vs_incomplete}). In some cases it is marginally better, likely because shorter lists reduce the number of reasoning steps required. Models also handle MCPO allocations (a notion specific to the incomplete setting) reasonably well, though with a sharper size-induced decline than for the core.\footnote{When a max-cardinality matching is infeasible, models consistently assign unranked items to agents rather than leaving them unmatched, revealing an implicit assumption that receiving any item is preferable to receiving none.}

\textbf{Performance is robust to ties, when the same algorithm works.}

\begin{figure}
    \centering
    \includegraphics[width=0.8\linewidth]{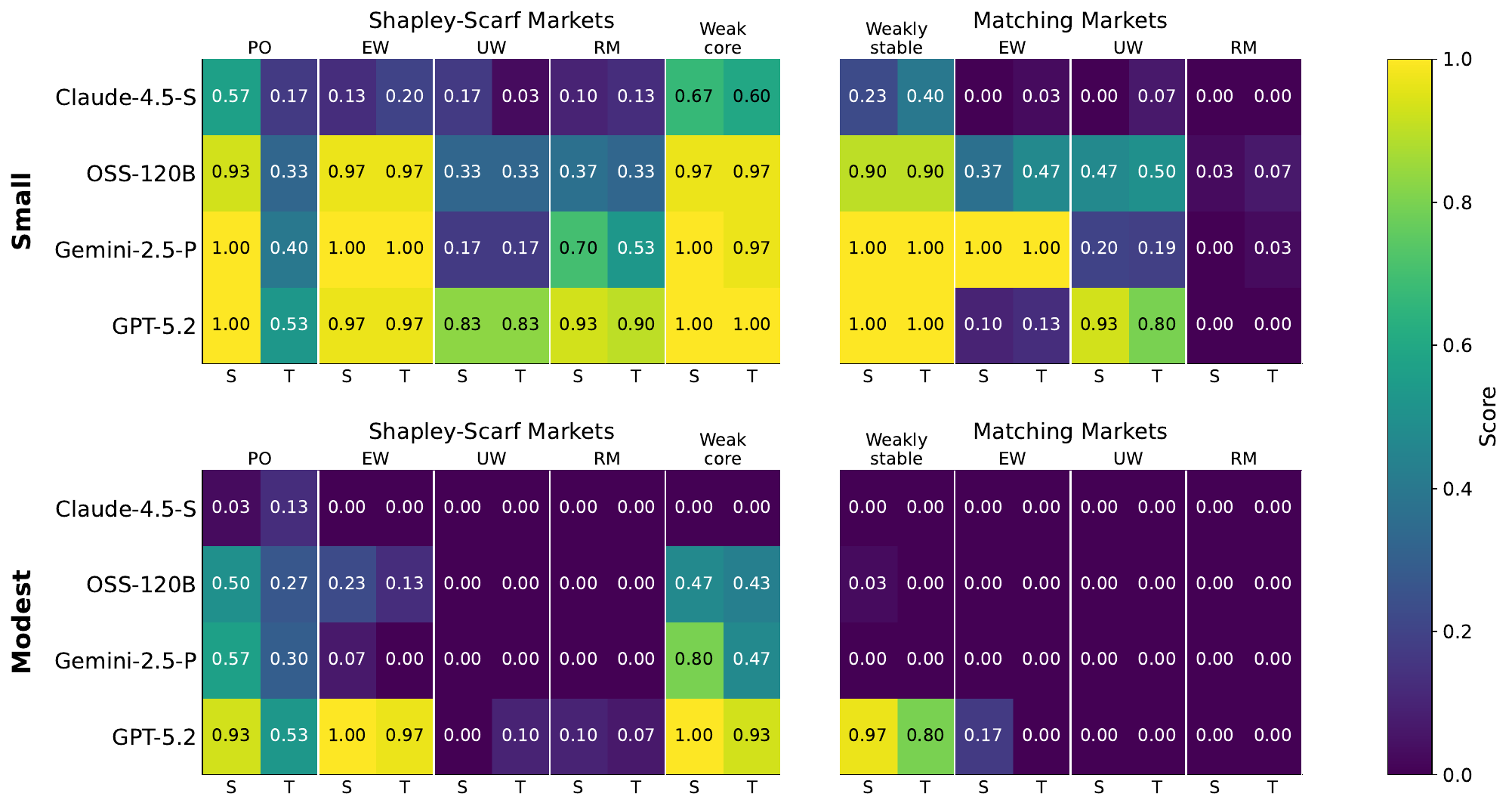}
    \caption{Difference in LLMs' performance in computing various types of solutions with strict preferences (S) and preferences having ties (T). }\label{fig:abs_strict_vs_ties_thin_separators}
\end{figure}

The introduction of ties likewise does not significantly affect performance on notions whose underlying algorithm is unchanged relative to the strict setting, such as the weak core, UW, EW, and RM. The one case where performance does drop is Pareto-optimality in house allocation: TTC guarantees a PO outcome under strict preferences but not under ties, so models that correctly apply TTC continue to succeed on the core yet fail to achieve PO in the same instances.

\subsection{Generation without a Specific Objective}\label{app:gen_a_sol}

\begin{figure}
    \centering
    \includegraphics[width=\linewidth]{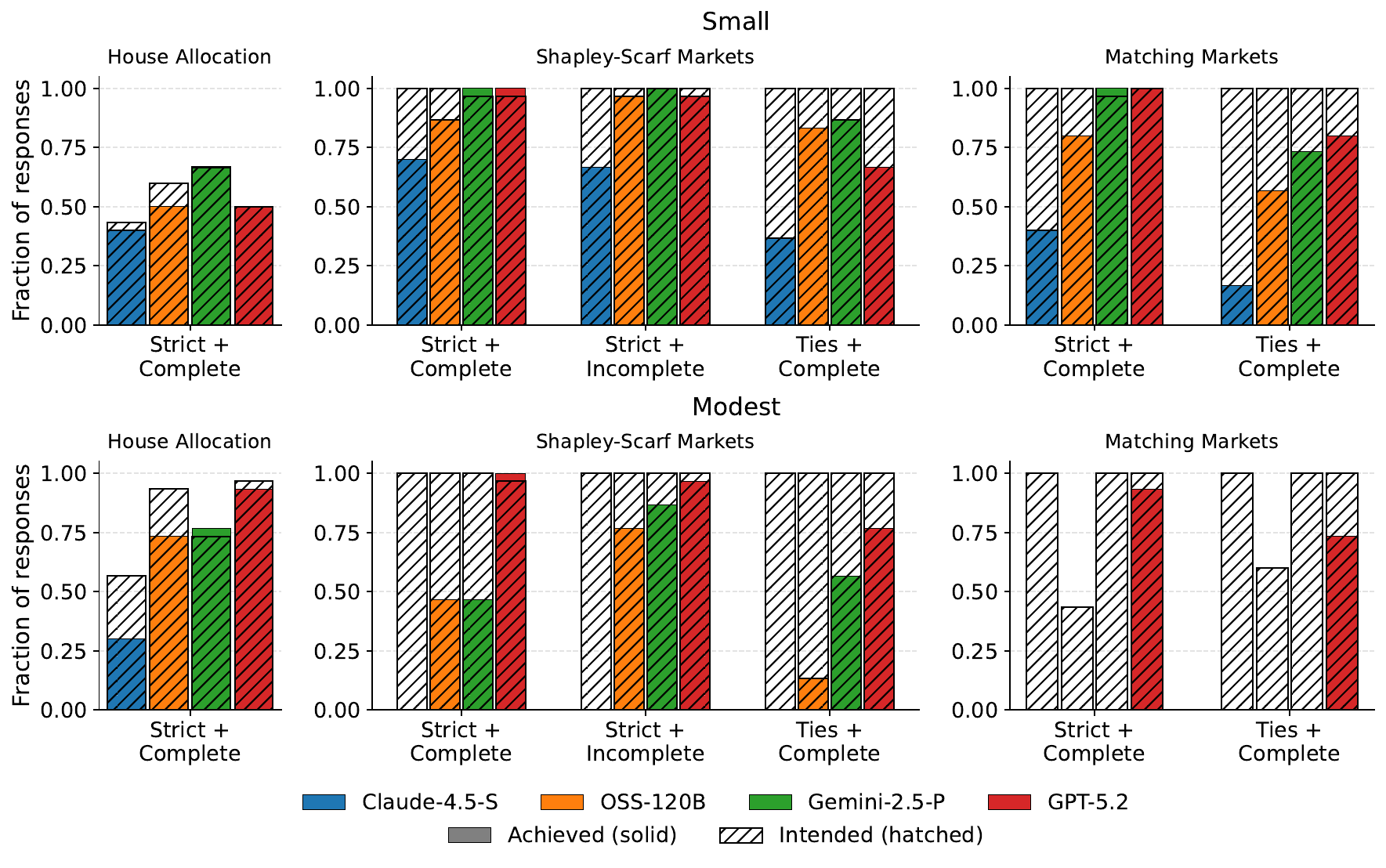}
    \caption{Fraction of responses where LLMs \textit{intend} to use the canonical algorithm (hatched) and where they \textit{achieve} the corresponding solution (solid), across domains and preference types, for $n=10$ and $n=30$.}
    \label{fig:unspecified}
\end{figure}

When asked to compute a solution without specifying a target notion, all models default to the canonical algorithm for the relevant domain across all preference types (\Cref{fig:unspecified}). The gap between hatched and solid bars reveals that models often correctly identify the canonical algorithm yet fail to execute it, particularly at $n=30$.\footnote{We identify the algorithm the model attempts to use, using the method described in \Cref{app:llm-judge}.} An exception arises in object allocation at $n=10$, where some models attempt to compute a solution satisfying stronger notions such as UW rather than simply applying SD; this tendency largely disappears at $n=30$, consistent with models falling back to the simpler SD procedure as more demanding computations exceed their reasoning budget.

\subsection{Failures in Infeasibility Detection}
\label{subsec:harder_notions}

\textbf{Models rely on heuristic search, which fails at scale.}

\begin{table}[ht]
  \centering
  \small
  \caption{Hard notions: accuracy by model, notion, instance size, and treatment.
    Each cell shows the fraction of correct responses.
    For \textbf{E} rows, the value in brackets is the fraction where the model
    returned an invalid solution---i.e., a response that is neither a timeout
    nor a valid one-to-one allocation.
    For \textbf{NE} rows with the \texttt{\{\}} instruction, a response is correct
    only if the model returned \texttt{\{\}}.
    For \textbf{NE} rows without the instruction, a response is correct if it is
    neither a timeout nor a valid one-to-one allocation.
    A superscript $^*$ indicates that at least one response timed out and was
    counted as incorrect; timeout counts are given in the note below.}
  \label{tab:hard_notions_existence}
  \resizebox{\textwidth}{!}{
  \begin{tabular}{@{}llc|cc|cc|cc|cc@{}}
    \toprule
    & & & \multicolumn{2}{c|}{Claude-4.5-S}
          & \multicolumn{2}{c|}{OSS-120B}
          & \multicolumn{2}{c|}{Gemini-2.5-P}
          & \multicolumn{2}{c}{GPT-5.2} \\
    Notion & $n$ & & with \{\} & without \{\} & with \{\} & without \{\} & with \{\} & without \{\} & with \{\} & without \{\} \\
    \midrule
    \multirow{4}{*}{\makecell{Strict\\Core}} & \multirow{2}{*}{10} & \textbf{E} & 0\% & 10\% & 3\% & 3\% & 17\% & 17\% & 100\% & 77\% \\
     &  & \textbf{NE} & 0\% & 0\% & 0\% & 0\% & 3\% & 0\% & 80\% & 20\% \\
    \cmidrule(l){3-11}
     & \multirow{2}{*}{30} & \textbf{E} & 0\% [3\%] & 0\% & 0\% [17\%] & 0\% & 0\% [3\%] & 3\% & 10\% & 7\% \\
     &  & \textbf{NE} & 0\% & 0\% & 13\% & 0\% & 0\% & 0\% & 0\% & 0\% \\
    \midrule
    \multirow{4}{*}{\makecell{Strongly\\Stable}} & \multirow{2}{*}{10} & \textbf{E} & 0\% & 7\% & 23\% [67\%] & 37\% [20\%] & 70\% [10\%] & 90\% & 97\% [3\%] & 90\%$^*$ \\
     &  & \textbf{NE} & 0\% & 0\% & 87\% & 63\% & 70\% & 0\% & 63\%$^*$ & 73\%$^*$ \\
    \cmidrule(l){3-11}
     & \multirow{2}{*}{30} & \textbf{E} & 0\% & 0\% & 0\% [97\%] & 0\% [27\%] & 0\% [7\%] & 0\% [7\%] & 23\% [40\%]$^*$ & 40\%$^*$ \\
     &  & \textbf{NE} & 0\% & 0\% & 97\% & 23\% & 7\% & 7\% & 57\%$^*$ & 30\%$^*$ \\
    \midrule
    \multirow{4}{*}{\makecell{Super\\Stable}} & \multirow{2}{*}{10} & \textbf{E} & 0\% [13\%] & 0\% & 3\% [80\%] & 10\% [67\%] & 23\% [60\%] & 33\% & 87\% [3\%] & 80\%$^*$ \\
     &  & \textbf{NE} & 7\% & 0\% & 87\% & 87\% & 73\% & 0\% & 65\%$^*$ & 50\%$^*$ \\
    \cmidrule(l){3-11}
     & \multirow{2}{*}{30} & \textbf{E} & 0\% [43\%] & 0\% & 0\% [100\%] & 0\% [100\%] & 0\% [20\%] & 0\% & 0\% [93\%]$^*$ & 3\%$^*$ [70\%] \\
     &  & \textbf{NE} & 53\% & 0\% & 97\% & 100\% & 40\% & 0\% & 97\% & 83\%$^*$ \\
    \bottomrule
  \end{tabular}}
  \par\smallskip
  {\footnotesize\noindent
  \textbf{Timeouts (out of 30).}
  \textit{GPT-5.2 with \{\}:}
  Strongly Stable $n{=}10$, \textbf{NE}: 9;\;
  Strongly Stable $n{=}30$, \textbf{E}: 3, \textbf{NE}: 4;\;
  Super Stable $n{=}10$, \textbf{E}: 3, \textbf{NE}: 9;\;
  Super Stable $n{=}30$, \textbf{E}: 1.
  \textit{GPT-5.2 without \{\}:}
  Strongly Stable $n{=}10$, \textbf{E}: 2, \textbf{NE}: 7;\;
  Strongly Stable $n{=}30$, \textbf{E}: 5, \textbf{NE}: 15;\;
  Super Stable $n{=}10$, \textbf{E}: 6, \textbf{NE}: 15;\;
  Super Stable $n{=}30$, \textbf{E}: 4, \textbf{NE}: 2.
  All timeouts are counted as incorrect.}
\end{table}

Models largely fail to compute strict core, strong stability, and super stability even at $n=10$ (\Cref{tab:hard_notions_existence}), with performance declining further at $n=30$. Where models do succeed at $n=10$, particularly GPT-5.2 and Gemini-2.5-P, inspection of reasoning traces reveals that they rely on heuristic search rather than correct algorithms: they enumerate candidate solutions, verify blocking conditions case by case, or use variants of the correct algorithm that still require heuristic completion (see Appendix~\ref{app:reasoning-strategies}). This strategy can succeed for small instances but fails systematically at larger ones, where the search space grows beyond the model's effective reasoning capacity.

\textbf{Models cannot reliably detect infeasibility.}

The pattern on infeasible instances reveals two distinct failure modes. The first is a \emph{default-to-empty bias}: models that achieve high NE accuracy often do so by returning $\{\}$ indiscriminately, as evidenced by the simultaneously high invalid-solution rates (bracket values) on feasible instances. The apparent NE accuracy for these models reflects a bias toward claiming infeasibility rather than genuine detection. This bias is also sensitive to instance difficulty: some models that return $\{\}$ rarely at $n=10$ return it far more often at $n=30$, when the problem is large enough that the model appears to abandon computation.

The second failure mode is \emph{prompt-contingent infeasibility detection}: removing the $\{\}$ escape hatch causes NE accuracy to collapse for most models, confirming that their detection depends on the prompt providing an explicit mechanism to express it rather than on genuine reasoning. GPT-5.2 is more robust to this change, though even its NE accuracy degrades at $n=30$.

The tendency to claim infeasibility also varies by notion in a consistent ordering across all models: nearly no model returns $\{\}$ for the strict core, while high rates are common for strong and super stability. At $n=10$, GPT-5.2 largely escapes this pattern because it can genuinely solve these problems; at $n=30$, where it can no longer do so, its NE rates follow the same difficulty-ordered pattern as the other models. This suggests that the propensity to declare infeasibility is driven by perceived problem difficulty rather than correct reasoning about whether a solution exists.

Overall, infeasibility detection that disappears without an explicit escape hatch, NE accuracy dominated by a default-to-empty bias, and a notion ordering tracking difficulty rather than logic all indicate that models lack reliable knowledge of these harder notions and adapt their outputs to whatever the prompt makes available.

\textbf{Models cannot reliably identify infeasibility when selecting.}

\begin{figure}[t]
    \centering
    \includegraphics[width=0.9\linewidth]{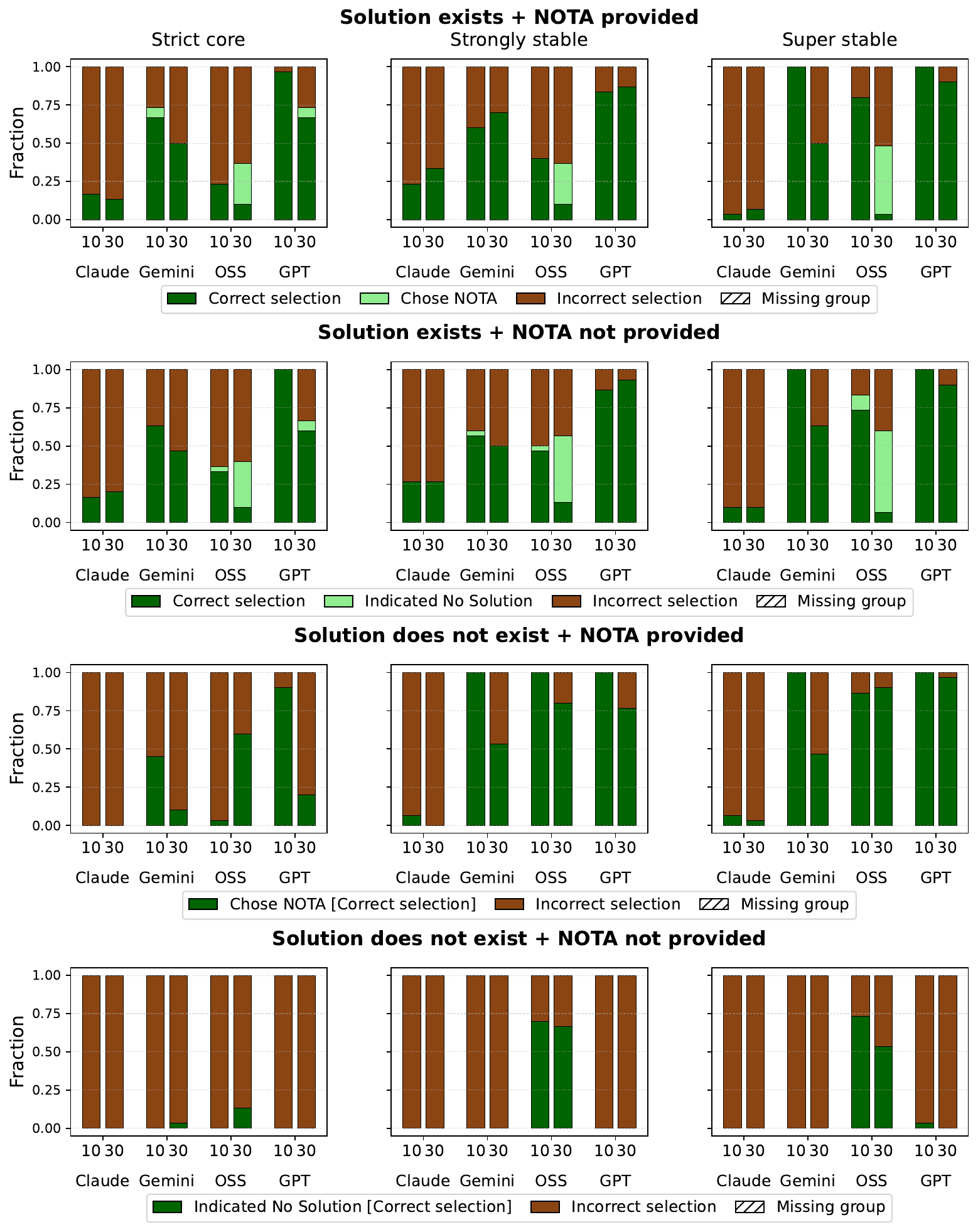}
    \caption{Selection outcomes for harder notions (strict core, strong and super stability), separated by whether a solution exists (E/NE) and whether a ``None of the Above'' (NOTA) option was included, at $n=10$ and $n=30$.}
    \label{fig:selection_grid}
\end{figure}

We additionally ask models to identify a specific notion among a set of candidates, a task that should be easier than generation since it only requires solution \emph{verification} rather than executing an algorithm to arrive at a solution. As shown in \Cref{fig:selection_grid}, even this task proves difficult: models frequently select incorrect candidates when a correct option exists. When NOTA is available, models use it in poorly calibrated ways. Notably, when no correct option is present, models tend to select whichever candidate satisfies a weaker version of the target notion, such as weak core instead of strict core or a weakly stable matching instead of a strongly stable one. This connects directly to the pattern observed above: the bias toward weaker notions in preference selection translates into a specific type of error when the target notion does not exist. While GPT-5.2 significantly outperforms other LLMs when the intended solution exists, its performance drops significantly when the solution does not exist - especially on larger instances, or when the NOTA option is not provided.

\subsection{Selecting from Options}\label{app:intention_action}

\begin{figure}[t]
    \centering
    \includegraphics[width=\linewidth]{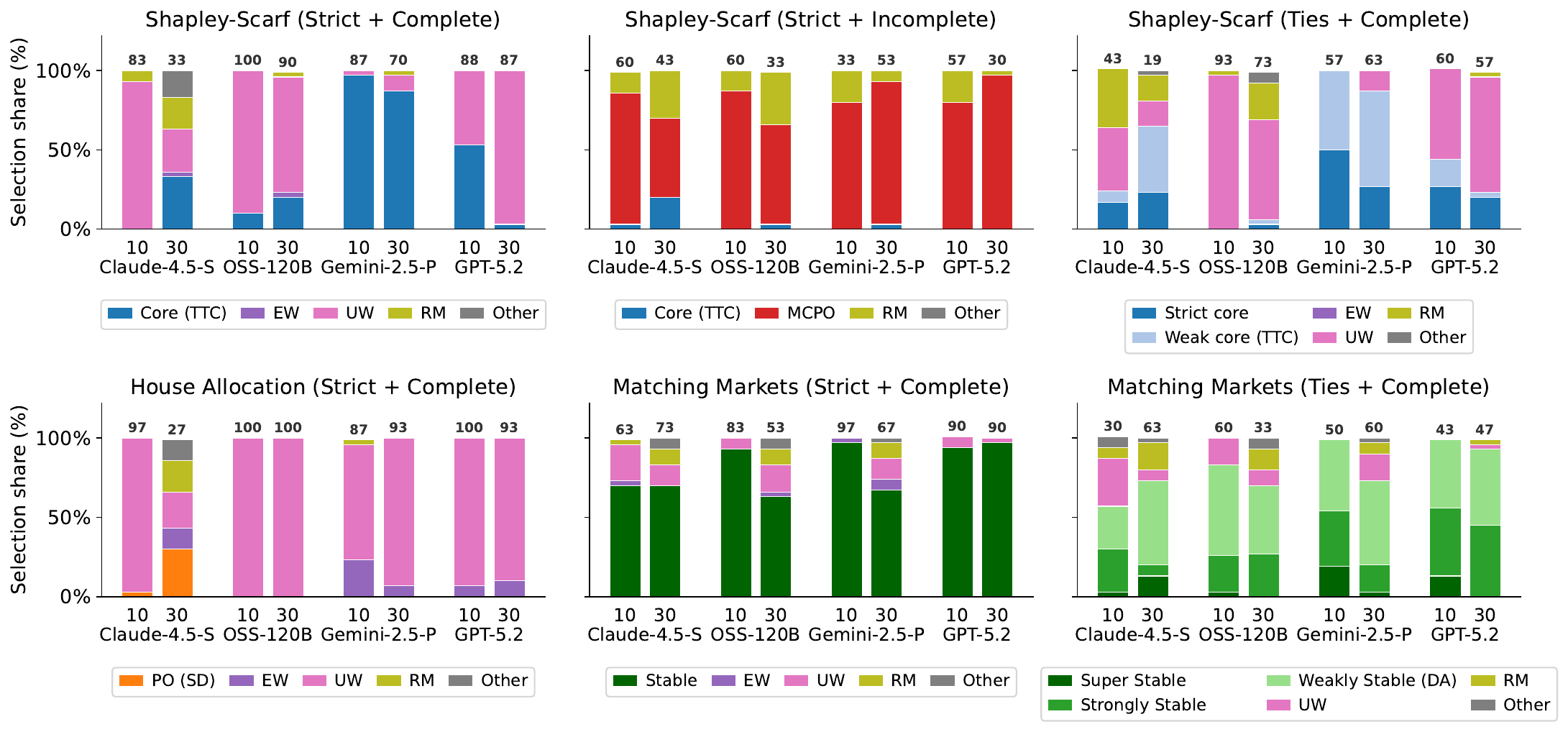}
    \caption{Share of selections for each solution concept across models, domains, and preference types, at $n=10$ and $n=30$. The number on top of each bar represents the percentage of responses where the \textit{intended} notion matches the \textit{actual} notion, i.e. that satisfied by the selected option.}
    \label{fig:selection_stacked}
\end{figure}

\paragraph{Models reveal systematic preference biases when selecting from candidates.}
When presented with a candidate set and asked for their most preferred solution, models reveal preferences that diverge from their generation behaviour (\Cref{fig:selection_stacked}). In allocation domains, models select utilitarian welfare-maximising outcomes far more often than canonical solutions, contrary to their default behaviour in generation. This recovers and extends the utilitarian bias while generating allocations of resources or tasks observed in previous work \citep{hosseini2025distributive,shi2025social,Cookson2026Fairness}, showing that it also surfaces in selection and with a wider set of solution concepts. In stable matching, models favor weakly stable solutions even when stronger options are present. 

\paragraph{Intention-action Gap.} \Cref{fig:selection_stacked} shows that models often end up selecting a solution that satisfies a property different from the one intended.\footnote{The notion a model ``intended'' to satisfy is identified using an LLM judge (Gemini-2.5-Flash); the judge prompt is provided in Appendix~\ref{app:llm-judge}.} This divide is higher under incomplete and tied preferences than under strict and complete preferences. In the incomplete setting, models tend to prioritize max-cardinality outcomes regardless of the notion they express an intent to target.\footnote{This is the one cell where the judges disagree, so we report it as judge-sensitive rather than pooling it (\Cref{app:judge_validation}).} Under preferences with ties, models frequently fail to distinguish between weaker and stronger versions of the same notion, for instance selecting a weakly stable matching when a strongly stable one is among the options.\footnote{From an analysis of the reasoning traces, it is clear that models use the absence of (weak) blocking pairs as the primary criterion to evaluate matchings, and compare all matchings that are at least weakly stable in terms of other metrics such as welfare or Pareto-improvements.}

\providecommand{\takeawaybox}{}%
\renewcommand{\takeawaybox}[1]{%
  \begin{tcolorbox}[
      colback=gray!7,
      colframe=gray!50,
      boxrule=0.4pt,
      arc=2pt,
      left=6pt,right=6pt,top=4pt,bottom=4pt,
      breakable]
    \small #1
  \end{tcolorbox}
}

\providecommand{\modelquote}[1]{%
  \begin{tcolorbox}[
      colback=blue!2,
      colframe=blue!25,
      boxrule=0.3pt,
      arc=1pt,
      left=6pt,right=6pt,top=3pt,bottom=3pt,
      breakable,
      fontupper=\small\itshape]
    #1
  \end{tcolorbox}
}

\section{Statistical analysis of quantitative results}
\label{app:statistics}

We compute 95\% confidence intervals for each individual experiment, and paired tests for comparisons between problems, models and formats. Intervals are Wilson intervals for a binomial proportion. Paired comparisons on the same instances use McNemar's test, and comparisons across instance sizes use Fisher's exact test. Every analysis in this appendix runs on the responses already collected, so no additional model queries were needed. Because each prompt is scored from a single draw, these intervals are the statement of uncertainty over instances; \Cref{app:model_details} separately shows the results are unchanged at temperature 0 for the two models that expose it.

\subsection{Indeterminacy gap of preference reasoning tasks}
For bundle comparison and ranking tasks, the determined-vs-undetermined gap (\Cref{fig:pref_bars_combined}(a)) is robust, with determined queries having accuracy near 1 and undetermined queries at near 0. The confidence intervals do not overlap for every task and model. These are the cases where recognizing indeterminacy requires reasoning. On bundle comparison (\Cref{tab:bundle}), determined accuracy is 1.00 in eleven of the twelve (model $\times$ preference type) cells and 0.85 in the twelfth, while undetermined accuracy never exceeds 0.20.

The gap does not have this shape everywhere, and where it does not is informative. On atomic and aggregative queries the intervals overlap and the gap is model-dependent, since an unranked item is missing by inspection and recognizing it needs no reasoning step. On partial-order queries the pattern is mixed and in two cases reversed. At 150 edges (\Cref{tab:partial_order_pairwise_comparable,tab:partial_order_pairwise_incomparable}), Claude-4.5-S scores 1.00 on determined pairs against 0.00 on undetermined ones and Gemini-2.5-P 1.00 against 0.50, whereas GPT-5.2 scores 0.17 against 0.97 and OSS-120B 0.23 against 1.00, because producing a determinate answer on a determined input is itself hard at this graph density (\Cref{app:f3_detail}). We therefore state the indeterminacy gap for the comparative tasks where it is clean, and report the other families per cell.

\subsection{Model differences on hard notions of algorithmic reasoning}
For the hard notion problems in \Cref{tab:hard_notions_existence}, the model differences are significant, with all six pairwise comparisons having $p \leq 0.004$ on the McNemar's test. GPT-5.2 is the only model with non-trivial accuracy on the matched feasible instances. Table~\ref{tab:stat_mcnemar} reports the per-pair values.

\begin{table}[h]
\centering
\caption{Hard-notion generation: pairwise McNemar tests over 360 paired instances. Per-model accuracies are given in \Cref{tab:hard_notions_existence}.}
\label{tab:stat_mcnemar}
\small
\begin{tabular}{lc}
\toprule
\textbf{Pair} & \textbf{$p$} \\
\midrule
GPT-5.2 vs.\ Gemini-2.5-P      & $<0.001$ \\
GPT-5.2 vs.\ Claude-4.5-S      & $<0.001$ \\
GPT-5.2 vs.\ OSS-120B          & $<0.001$ \\
Gemini-2.5-P vs.\ Claude-4.5-S & $<0.001$ \\
Gemini-2.5-P vs.\ OSS-120B     & $<0.001$ \\
Claude-4.5-S vs.\ OSS-120B     & $0.004$ \\
\bottomrule
\end{tabular}
\end{table}

\subsection{Scaling of input size on generation tasks}
For the comparison of accuracy on small and modest market sizes in \Cref{fig:size_collapse}(a), a majority of comparisons are significant under Fisher's exact test, in 32 of 53 cells. The significance holds for models that achieve at least some level of performance on the small market, with room to decline on the modest market size.

\subsection{Format effect on undetermined preference queries}
For the comparison of different formats on undetermined preference queries in \Cref{fig:pref_bars_combined}, the comparison is significant on approximately half of the cases, and the effect is strongly task-dependent (Table~\ref{tab:stat_format}), which is why we report it per cell rather than as a single average.

\begin{table}[h]
\centering
\caption{Format effect (free-flow vs.\ MCQ) on undetermined preference queries: cells with non-overlapping 95\% intervals.}
\label{tab:stat_format}
\small
\begin{tabular}{lc}
\toprule
\textbf{Model} & \textbf{Significant cells} \\
\midrule
Gemini-2.5-P & 11 of 20 \\
GPT-5.2      & 10 of 20 \\
Claude-4.5-S & 10 of 19 \\
OSS-120B     & 8 of 20 \\
\bottomrule
\end{tabular}
\end{table}

\subsection{Influence of "NOTA" option on hard selection problems}
The effect of the provided option in \Cref{fig:selection_grid} is significant and one-sided. No model selects NOTA when a solution exists, and on infeasible instances models under-select it (Table~\ref{tab:stat_nota}). Offering the option therefore does not induce spurious abstention here; the error is a failure to abstain when abstention is correct.

\begin{table}[h]
\centering
\caption{Selection with a ``none of the above'' option: fraction of instances on which the model selects NOTA.}
\label{tab:stat_nota}
\small
\begin{tabular}{lcc}
\toprule
\textbf{Model} & \textbf{Feasible (NOTA is wrong)} & \textbf{Infeasible (NOTA is correct)} \\
\midrule
GPT-5.2      & 0.00 & 0.61 \\
Gemini-2.5-P & 0.00 & 0.59 \\
OSS-120B     & 0.00 & 0.56 \\
Claude-4.5-S & 0.00 & 0.03 \\
\bottomrule
\end{tabular}
\end{table}

\section{Prompt Templates}
\label{app:prompts}

\subsection{Preference reasoning tasks.}

All preference-reasoning prompts share a common template:

\begin{quote}
\small\ttfamily\noindent
You are an intelligent agent who is an expert in algorithms. Consider the following instance of the object allocation problem, where \textit{[n]} alternatives have to be allocated to \textit{[n]} agents. Given below are the preferences agents have over the alternatives.\\[4pt]
\textless preferences\textgreater\\
\textit{[preferences\_json]}\\
\textless/preferences\textgreater\\[4pt]
\textit{[task\_string]}\\[4pt]
\textit{[format\_string]}
\end{quote}

Task strings and format strings for each query type are listed in Table~\ref{tab:preference_prompt}. In addition to the base free-flow prompt, for undetermined queries, additional wording is appended to permit an ``unknown'' or ``can't decide'' response. We refer to this as free-flow + ``if known''. Additionally, for each problem we design MCQ (multiple-choice problem) prompts for each task, with the option being potential answers, plus an option of "There is insufficient information to decide."

\begin{table}[h]
\centering
\caption{Preference reasoning prompts: task and formatting instructions. The base free-flow prompt is shown, while the wordings in \textit{italics} are added for free-flow + "if known" format.}
\label{tab:preference_prompt}
\small
\begin{tabular}{p{6.8cm} p{6.8cm}}
\toprule
\textbf{Task string} & \textbf{Format string} \\
\midrule
What position is [choice] at in [agent]'s preference list? & Return\textless answer\textgreater X\textless/answer\textgreater, where X is the number indicating the position, \textit{if known}. \\[3pt]
\midrule
What is [agent]'s [k]th most preferred choice? & Return\textless answer\textgreater X\textless/answer\textgreater, where X is the name of the choice, \textit{if known}. \\[3pt]
\midrule
Which alternative does [agent] prefer between [choice 1] and [choice 2]? & Return\textless answer\textgreater X\textless/answer\textgreater, where X is the name of the choice, \textit{if known}. \\[3pt]
\midrule
Rank [4 choices] according to [agent]'s preference. & Return the solution in the following format: \textless answer\textgreater ranking\textless/answer\textgreater, with \textit{only ranked} choice names in descending order connected by '\textgreater' sign. \\[3pt]
\midrule
How many agents prefer [choice 1] over [choice 2]? & Return \textless answer\textgreater X\textless/answer\textgreater, where X is the number of agents. \\[3pt]
\midrule
Which bundle does [agent] prefer between \{[choice 1], [choice 2]\} and \{[choice 3], [choice 4]\}? & Return the solution in the following format: \textless answer\textgreater\{A, B\}\textless/answer\textgreater, where \{A, B\} is the preferred bundle. \textit{If there is no preference, return \textless answer\textgreater Can't decide\textless/answer\textgreater.} \\[3pt]
\midrule
If [agent] selects its top-[k] alternatives, does [choice] belong to this list? & Return \textless answer\textgreater solution\textless/answer\textgreater with solution being yes, no \textit{or uncertain}. \\[3pt]
\midrule
What are the top-[k] choices of [agent]? & Return \textless answer\textgreater solution\textless/answer\textgreater with choice names in a list format. \\[3pt]
\midrule
Select the top-[k] choices of [agent]. Does [choice] belong to this list? & Return the solution in the following format: \textless answer\textgreater solution\textless/answer\textgreater. \\[3pt]
\midrule
(Partial order) Between [choice 1] and [choice 2], which is preferred, \textit{if known}? & Return the answer in a single word in the following format: \textless answer\textgreater word\textless/answer\textgreater. \\[3pt]
\bottomrule
\end{tabular}
\end{table}

\subsection{Algorithmic reasoning tasks.}
 
\paragraph{Generation tasks.}
\begin{quote}
\small\ttfamily\noindent
You are an intelligent agent who is an expert in algorithms. Consider the following instance of the \textit{[domain]} problem, where \textit{[n]} alternatives have to be allocated to \textit{[n]} agents. Given below are the preferences agents have over the alternatives.\\[4pt]
\textless preferences\textgreater\\
\textit{[preferences\_json]}\\
\textless/preferences\textgreater\\[4pt]
\textless endowment\textgreater\quad\textit{(Shapley-Scarf market only)}\\
\textit{[endowment\_json]}\\
\textless/endowment\textgreater\\[4pt]
Your task is to compute a \textit{[notion]} allocation for the given preferences.\\[4pt]
\textit{[notion\_definition]}\\[4pt]
Return the solution in the following format:\\
\textless answer\textgreater\{\textquotedbl A1\textquotedbl: \textquotedbl\textit{assigned alternative}\textquotedbl, \ldots\}\textless/answer\textgreater\\[4pt]
If there is no \textit{[notion]} solution, return \textless answer\textgreater\{\}\textless/answer\textgreater.
\end{quote}
 
\paragraph{Selection tasks.}
\begin{quote}
\small\ttfamily\noindent
You are an intelligent agent who is an expert in algorithms. Consider the following instance of the \textit{[domain]} problem, where \textit{[n]} alternatives have to be allocated to \textit{[n]} agents. Given below are the preferences agents have over the alternatives.\\[4pt]
\textless preferences\textgreater\\
\textit{[preferences\_json]}\\
\textless/preferences\textgreater\\[4pt]
\textless endowment\textgreater\quad\textit{(Shapley-Scarf market only)}\\
\textit{[endowment\_json]}\\
\textless/endowment\textgreater\\[4pt]
Your task is to select the \textit{[notion]} allocation among the given options:\\[4pt]
A: \textit{[allocation\_json]}\\
B: \textit{[allocation\_json]}\\
\ldots\\
\textit{[K]}: None of the above\quad\textit{(NOTA condition only)}\\[4pt]
\textit{[notion\_definition]}\\[4pt]
Return only the letter of your chosen option inside \textless answer\textgreater\ \textless/answer\textgreater\ tags,\\
e.g., \textless answer\textgreater A\textless/answer\textgreater.
\end{quote}

\paragraph{Preference encoding.} Preferences are passed to the model as JSON, with one entry per agent listing alternatives in decreasing order of preference. In Shapley-Scarf market tasks, the preferences are one-sided (agents over objects) and the endowment block lists each agent's initial holding.

\begin{quote}
\small\ttfamily\noindent
\textless preferences\textgreater\\
\{\\
\quad \textquotedbl A1\textquotedbl: [\textquotedbl O3\textquotedbl, \textquotedbl O1\textquotedbl, \textquotedbl O2\textquotedbl],\\
\quad \textquotedbl A2\textquotedbl: [\textquotedbl O1\textquotedbl, \textquotedbl O2\textquotedbl, \textquotedbl O3\textquotedbl],\\
\quad \textquotedbl A3\textquotedbl: [\textquotedbl O2\textquotedbl, \textquotedbl O3\textquotedbl, \textquotedbl O1\textquotedbl]\\
\}\\
\textless/preferences\textgreater\\[4pt]
\textless endowment\textgreater\\
\{\textquotedbl A1\textquotedbl: \textquotedbl O1\textquotedbl, \textquotedbl A2\textquotedbl: \textquotedbl O2\textquotedbl, \textquotedbl A3\textquotedbl: \textquotedbl O3\textquotedbl\}\\
\textless/endowment\textgreater
\end{quote}

In matching market tasks, preferences are two-sided: each side has its own preference dictionary over the other side, and no endowment block is included.

\begin{quote}
\small\ttfamily\noindent
\textless preferences\textgreater\\
\{\\
\quad \textquotedbl A\textquotedbl: \{\\
\quad\quad \textquotedbl A1\textquotedbl: [\textquotedbl B2\textquotedbl, \textquotedbl B1\textquotedbl, \textquotedbl B3\textquotedbl],\\
\quad\quad \textquotedbl A2\textquotedbl: [\textquotedbl B1\textquotedbl, \textquotedbl B3\textquotedbl, \textquotedbl B2\textquotedbl],\\
\quad\quad \textquotedbl A3\textquotedbl: [\textquotedbl B3\textquotedbl, \textquotedbl B2\textquotedbl, \textquotedbl B1\textquotedbl]\\
\quad \},\\
\quad \textquotedbl B\textquotedbl: \{\\
\quad\quad \textquotedbl B1\textquotedbl: [\textquotedbl A2\textquotedbl, \textquotedbl A1\textquotedbl, \textquotedbl A3\textquotedbl],\\
\quad\quad \textquotedbl B2\textquotedbl: [\textquotedbl A1\textquotedbl, \textquotedbl A3\textquotedbl, \textquotedbl A2\textquotedbl],\\
\quad\quad \textquotedbl B3\textquotedbl: [\textquotedbl A3\textquotedbl, \textquotedbl A2\textquotedbl, \textquotedbl A1\textquotedbl]\\
\quad \}\\
\}\\
\textless/preferences\textgreater
\end{quote}

When preferences include ties, items at the same indifference level are grouped in nested lists, e.g.\ \texttt{[\textquotedbl O1\textquotedbl, [\textquotedbl O2\textquotedbl, \textquotedbl O3\textquotedbl], \textquotedbl O4\textquotedbl]} represents $O_1 \succ \{O_2 \sim O_3\} \succ O_4$. Incomplete preferences omit the unranked alternatives entirely.

Here's a draft paragraph for the generation-prompt subsection:

\paragraph{Reasoning trace elicitation.} GPT-5.2 and OSS-120B do not expose their internal reasoning traces by default, unlike Gemini-2.5-Pro and Claude-4.5-Sonnet, which return a visible chain-of-thought alongside the final answer. To enable an analysis of the strategies being used by these models to solve the given problems, we append the following instruction to the prompt for these two models:
\begin{quote}
\small\ttfamily\noindent
Briefly explain your approach within \textless scratchpad\textgreater\ \textless/scratchpad\textgreater\ tags after providing the answer in the above format.
\end{quote}
This instruction is appended after the answer-format specification after the conditional \{\} instruction if included, and otherwise after the answer format instruction.

\paragraph{Notion Definitions Used in Prompts}
\label{app:definitions}
 
Tables~\ref{tab:defs_allocation} and~\ref{tab:defs_matching} list the exact definitions included verbatim in generation prompts for each solution concept. Definitions were held constant across all models and instance sizes.

\begin{table}[h]
\centering
\caption{Definitions provided to models for Shapley-Scarf market and house allocation tasks. The prompts label this notion \textsc{core} under strict preferences and \textsc{weak core} under ties. Both labels carry the same blocking condition, because the two notions coincide, so we merge the rows here. The refinement that genuinely differs is the strict core, which blocks when every member of a coalition weakly improves and at least one improves strictly.}
\label{tab:defs_allocation}
\small
\begin{tabular}{p{2.8cm} p{10cm}}
\toprule
\textbf{Notion} & \textbf{Definition} \\
\midrule
Pareto-optimal & An allocation is Pareto-optimal if there is no other feasible allocation that makes at least one agent strictly better off without making any other agent worse off. \\
\cmidrule(lr){1-2}
MCPO & A max-cardinality Pareto-optimal (MCPO) allocation matches as many agents as possible to alternatives they rank, and among all such maximum-size matchings it is Pareto-optimal (no Pareto-improving reassignment exists). \\
\cmidrule(lr){1-2}
Core (weak core) & An allocation is in the core if there is no coalition of agents that can reshuffle their initially endowed houses among themselves so that every coalition member is strictly better off than in the allocation. \\
\cmidrule(lr){1-2}
Strict core & An allocation is in the strict core if there is no blocking coalition that can reshuffle endowed houses so that every member weakly improves and at least one member strictly improves. \\
\cmidrule(lr){1-2}
UW-maximizing & A utilitarian welfare-maximizing allocation minimizes the total sum of agents' assigned ranks (equivalently, maximizes total ordinal utility). \\
\cmidrule(lr){1-2}
EW-maximizing & An egalitarian welfare-maximizing allocation minimizes the worst (largest) rank any agent receives (i.e., it optimizes the welfare of the worst-off agent). \\
\cmidrule(lr){1-2}
Rank-maximal & A rank-maximal allocation maximizes the number of agents receiving a 1st-choice alternative; subject to that, it maximizes the number receiving a 2nd-choice; and so on (lexicographic maximization of the rank-count vector). \\
\bottomrule
\end{tabular}
\end{table}
 
\begin{table}[h]
\centering
\caption{Definitions provided to models for matching market tasks.}
\label{tab:defs_matching}
\small
\begin{tabular}{p{2.8cm} p{10cm}}
\toprule
\textbf{Notion} & \textbf{Definition} \\
\midrule
Weakly stable & A matching is weakly stable if there is no blocking pair where both agents strictly prefer each other to their current partners; ties do not create blocking unless both sides strictly gain. \\
\cmidrule(lr){1-2}
Strongly stable & A matching is strongly stable if there is no blocking pair where both agents weakly prefer each other to their current partners and at least one agent strictly prefers the other. \\
\cmidrule(lr){1-2}
Super stable & A matching is super stable if there is no blocking pair where both agents weakly prefer each other to their current partners; even indifference on both sides can block. \\
\cmidrule(lr){1-2}
UW-maximizing & A utilitarian welfare-maximizing matching minimizes the sum of (ordinal) partner ranks. \\
\cmidrule(lr){1-2}
EW-maximizing & An egalitarian welfare-maximizing matching minimizes the maximum (worst) assigned partner rank. \\
\cmidrule(lr){1-2}
Rank-maximal & A rank-maximal matching lexicographically maximizes the rank-count vector: as many agents as possible get a 1st-choice partner; subject to that, as many as possible get a 2nd-choice; and so on. \\
\bottomrule
\end{tabular}
\end{table}

\section{Reasoning Strategies on Feasible and Infeasible Tasks (GPT-5.2)}
\label{app:reasoning-strategies}

We restrict the qualitative analysis in this appendix to GPT-5.2.
Among the four frontier models we evaluated (Claude Sonnet~4.5,
Gemini~2.5 Pro, GPT-OSS~120B, and GPT-5.2), GPT-5.2 is the only
model to achieve non-trivial correctness on the three solution
concepts that admit both feasible and infeasible instances: the
strict core of a Shapley--Scarf housing market, strongly stable
matching, and super stable matching. On the feasible instances of
Small size, GPT-5.2's correctness rates are $100\%$, $97\%$ and
$87\%$ respectively, against runner-up rates of $17\%$, $70\%$ and
$23\%$ for Gemini~2.5 Pro and at most $23\%$ for the other two
models. Other models do return correct non-existence claims on the
infeasible instances, but they do so mainly by abstaining broadly,
since their accuracy on the matched feasible instances is close to
zero. GPT-5.2 is the only model that gets both sides right. We therefore focus on
what GPT-5.2 actually does and how those reasoning strategies
behave as the problem size grows from $n=10$ (Small) to $n=30$
(Modest).

The analysis below is built on GPT-5.2's own self-reports of its
reasoning. Every prompt in our benchmark closes with the
instruction \emph{``Briefly explain your approach within
\texttt{<scratchpad> </scratchpad>} tags after providing the answer
in the above format''}, and the scratchpads it produces are the
input to the qualitative coding here. To process them at scale, we
combine two views: a deterministic verifier that replays canonical
algorithms (Top Trading Cycles with first-in-list tie-breaking,
Irving's strongly-stable and super-stable matching algorithms, and
a polynomial-time strict-core check via blocking-cycle search on
the weakly-preferred-endowment digraph) on each model output, and
an LLM-judge (DeepSeek~V4~Pro) that extracts six verbatim features
per scratchpad: the named algorithm, any explicit trading cycle of
length~$\geq 3$, any pre-algorithm forced-pair derivations, the
stated tie-break rule, the verification step (re-checks vs.
property-citation vs. none), and the final claim
(allocation/matching produced vs.\ no-solution claim). The exact
judge setup, including a representative prompt, is given in
Section~\ref{app:reasoning:judge}.

The data underlying every percentage in this appendix is 720
GPT-5.2 trajectories on the generation task, $120$ per
$(\textit{notion},\textit{feasibility})$ cell, evenly split between
Small and Modest sizes. Within each $(\text{cell}, \text{size})$
pair the dataset contains $30$ instances, with two trajectories per
instance drawn from two prompt variants. The first variant, used
for the cross-model evaluation elsewhere in the paper, explicitly
instructs the model that ``if no [strict core / strongly stable /
super stable] solution exists, return \texttt{<answer>\{\}</answer>}'';
we will refer to it as the \emph{empty-hint} prompt. The second
variant omits this instruction and is otherwise identical; we will
refer to it as the \emph{no-hint} prompt. The two variants produce
strongly divergent behaviour on infeasible cells, where under the
no-hint prompt GPT-5.2 \emph{never} returns an explicit no-solution
answer on any of the $180$ infeasible-cell trajectories issued under that variant, so we
split per-cell statistics by prompt variant whenever the difference
matters.

\subsection{Strict core in Shapley--Scarf housing markets}
\label{app:reasoning:strict-core}

\paragraph{Feasible instances.}
On feasible strict-core instances, GPT-5.2 invariably runs Top
Trading Cycles (TTC). What varies is the tie-break rule, and on
this dataset the rule chosen determines correctness almost
entirely. The successful pattern is what we call the
\emph{constructive chain heuristic}: at each indifference, the
model picks the tied alternative that allows a long trading cycle
to close, rather than the lexicographically first tied alternative.
The choice is occasionally verbalised, as in this Small-size trace:

\begin{quote}\itshape
Approach: 1) Compute a core candidate via Top Trading Cycles (TTC)
adapted to ties by selecting one ``top'' house per agent when
indifferent. Running TTC on the endowment graph and choosing
$A_7$'s top pointer as $C_4$ (rather than keeping $C_{10}$) yields
the allocation above. 2) Verify the strict-core condition by
searching for a blocking coalition: any blocking reshuffle
decomposes into trading cycles, so it suffices to check whether
there is a directed cycle of agents
$i_1 \to i_2 \to \cdots \to i_k \to i_1$ in which each $i_t$ weakly
prefers the endowment of $i_{t+1}$ to its assigned house and at
least one strictly prefers it. Building this digraph for the
allocation above yields no such cycle, hence no blocking coalition
exists.
\end{quote}

\noindent
We label a trajectory as constructive-chain if the model's
allocation differs from canonical first-in-list TTC \emph{and} lies
in the strict core (the verifier produces the canonical TTC output
for each instance). At Small size, $30$ of $60$ trajectories
($50\%$) are constructive-chain, and $29$ of those $30$ ($97\%$)
land in the strict core. A residual non-mechanical bucket
(``arbitrarily but consistently'' chosen ties, in the model's own
words) accounts for another $26$ trajectories of which $24$
($92\%$) are also in the core. The empty-hint clause provides a
small lift: $30/30$ correct under the empty-hint prompt versus
$23/30$ ($77\%$) under the no-hint prompt.

At Modest size the picture inverts. The constructive-chain
stratum collapses to $2$ of $60$ ($3\%$), neither correct;
$30$ of $60$ trajectories ($50\%$) match canonical first-in-list
TTC exactly, none of which is in the strict core; the residual
non-mechanical bucket falls to $5$ correct of $28$ ($18\%$).
GPT-5.2 abandons the constructive heuristic in favour of a
mechanical rule that the TTC theorem does not justify under weak
preferences, and that empirically never recovers the strict core
on these instances.

\paragraph{Infeasible instances.}
On infeasible instances, the only correct strategy is to argue
that no allocation can satisfy the strict-core condition. GPT-5.2
does this via what we call \emph{iterative mandatory-pair
enumeration}: a sequence of forced assignments, each justified by
exhibiting a coalition that would block any allocation violating
it, terminating in a contradiction. A representative trace:

\begin{quote}\itshape
Approach: show non-existence by exhibiting a blocking coalition
for every possible allocation. If an allocation $x$ gives $A_8$
anything other than $C_2$, then coalition
$\{A_8, A_5, A_9, A_{10}\}$ blocks $x$ via
$A_8 \!\leftarrow\! C_2$ (top choice, strict improvement),
$A_5 \!\leftarrow\! C_5$, $A_9 \!\leftarrow\! C_1$,
$A_{10} \!\leftarrow\! C_7$ (each weakly improves). So any
strict-core allocation must assign $C_2$ to $A_8$. If an allocation
gives $A_4$ anything other than $C_2$, coalition
$\{A_4, A_5, A_9\}$ blocks similarly, so any strict-core
allocation must also assign $C_2$ to $A_4$. But $C_2$ is a single
house, so the strict core is empty.
\end{quote}

\noindent
Whether the model produces this argument at all is controlled
almost entirely by the prompt variant. At Small size under the
empty-hint prompt, $24$ of $30$ trajectories ($80\%$) explicitly
claim no allocation exists, and all $24$ are correct; under the
no-hint prompt, $0$ of $30$. The mandatory-pair argument is in the
model's repertoire, but it surfaces only when the prompt explicitly
mentions the empty-answer option. At Modest size, the empty-hint
prompt no longer suffices: $0$ of $30$ trajectories under either
variant claim non-existence. The forced-pair chains required at
$n=30$ are longer and the blocking coalitions larger, and even
with explicit permission to return $\{\}$, the model commits to
producing an allocation rather than working through the chain.

\subsection{Strongly stable matchings}
\label{app:reasoning:strongly-stable}

\paragraph{Feasible instances.}
The strongly-stable task is set on a bipartite instance with two
sides of agents, denoted $A$ and $B$. On feasible instances,
GPT-5.2 pairs deferred acceptance with hand-coded local repair: it
first identifies one or two forced pairs from the preference
structure (for instance, an $(A_i, B_j)$ where each side is the
unique top choice of the other), runs A-proposing deferred
acceptance on a tie-refinement to populate the matching, scans the
result for strong-blocking pairs created by the indifferences that
the tie-refinement collapsed, and applies short local swaps until
none remain. A representative trace:

\begin{quote}\itshape
Approach (strong stability with ties): I first constructed a
complete matching via an A-proposing deferred-acceptance run
(breaking ties arbitrarily just to get a candidate), then explicitly
checked the strong-stability condition. The initial DA matching
was not strongly stable: $B_5$ ranks $A_4$ as its unique top
choice, while $A_4$ ranks $B_5$ in its 2nd tier (tied with $B_6$).
If $A_4$ is not matched to $B_5$ and not to its strictly-better
option $B_8$, then $(A_4, B_5)$ is a strong blocking pair. This
forces $A_4 - B_5$. After enforcing $A_4 - B_5$, I iteratively
repaired the matching to remove remaining strong blocking pairs
caused by indifferences. The final matching is strongly stable.
\end{quote}

\noindent
At Small size, $45$ of $60$ trajectories ($75\%$) produce a
strongly stable matching and $11$ more produce a super-stable
matching (which strictly implies strong stability), for a combined
$56/60$ ($93\%$). Pre-algorithm forced-pair reasoning is the most
reliable distinguishing feature of correct trajectories.

The approach itself does not change at Modest size, but its
execution quality does. Strong stability is met by $15/60$
($25\%$) and super-stability by $4/60$, for a combined $19/60$
($32\%$); $25$ trajectories ($42\%$) produce no parseable matching,
and $13$ produce a matching that is not even weakly stable. The
local repair loop is the bottleneck: as the matching grows, every
swap to fix one strong-blocking pair tends to introduce another.
Notably, the empty-hint clause \emph{hurts} on this cell at Modest
size: $7/30$ ($23\%$) correct under the empty-hint prompt versus
$12/30$ ($40\%$) under the no-hint prompt. The most plausible
reading is that the explicit $\{\}$ option, when offered on a
feasible instance the model is struggling to solve at scale,
occasionally tempts it to retreat to the empty answer rather than
persist with the local-repair loop.

\paragraph{Infeasible instances.}
On infeasible strongly-stable instances, GPT-5.2 produces the
matching-theoretic analogue of the mandatory-pair impossibility
chain, again starting from a forced pair (typically a mutual top
tie) and deriving subsequent forced assignments that culminate in
an unavoidable strong-blocking pair:

\begin{quote}\itshape
Approach (proof of non-existence by forced blocking pairs). $A_5$
has a top tie $\{B_4, B_6\}$. If $A_5$ is matched to $B_6$, then
$(A_5, B_4)$ blocks: $A_5$ weakly prefers $B_4$ to $B_6$
(indifferent), and $B_4$ ranks $A_5$ in its top tie (with $A_2$),
so $B_4$ weakly prefers $A_5$ to any partner. To avoid this, $B_4$
must be matched to $A_2$. But then $(A_2, B_2)$ blocks: $A_2$
strictly prefers $B_2$ to $B_4$, and $B_2$ ranks $A_2$ as its top
choice. Therefore $A_5$ cannot be matched to $B_6$. \ldots\ Each
branch leads to an unavoidable strong-blocking pair, so no strongly
stable matching exists.
\end{quote}

\noindent
As with the strict-core infeasible cell, the rate at which the
model produces this argument is controlled by the prompt variant.
Under the empty-hint prompt: $19/30$ ($63\%$) at Small and
$17/30$ ($57\%$) at Modest. Under the no-hint prompt: $0/30$ at
each size. Across all $36$ traces that produce an impossibility
argument, every single one is correct; the forced-pair chain is
reliable enough that whenever the model commits to it, the
conclusion is right. The size behaviour of the empty-hint
trajectories is also notable: the rate is essentially
size-invariant ($63\% \to 57\%$), in contrast to the feasible cell
of this same notion (where Modest correctness drops from $93\%$ to
$32\%$).

\subsection{Super stable matchings}
\label{app:reasoning:super-stable}

\paragraph{Feasible instances.}
On feasible super-stable instances, GPT-5.2 names the
Irving--Manlove super-stable matching algorithm (often as ``Irving
SUPER'' or ``the standard super-stable SMTI reduction'') and at
Small size applies it convincingly:

\begin{quote}\itshape
Approach: I applied the standard super-stable SMTI reduction
procedure (Irving--Manlove style): repeatedly let free $A$-agents
propose to everyone in their current first (possibly-tied) tier;
each $B$-agent keeps the best (possibly-tied) proposers seen so
far and deletes from their list everyone strictly worse than the
kept tier; deletions are mirrored on the $A$-side. If any agent's
list becomes empty, no super-stable matching exists. \ldots\ This
pruning forced several pairs ($B_2$ can only take $A_1$, $B_5$
only $A_8$, $B_7$ only $A_2$), reducing the instance to a small
feasible edge set. From the remaining edges I constructed a
perfect matching and validated super-stability directly.
\end{quote}

\noindent
At Small size, $49$ of $60$ trajectories ($82\%$) produce a
super-stable matching; the remaining $11$ produce no parseable
matching. Strikingly, no Small trajectory produces a well-formed
matching that fails super-stability; the failure mode is
exclusively non-production rather than mis-production.

At Modest size, the picture changes qualitatively. Only $1/60$
($1.7\%$) produces a super-stable matching; $52$ ($87\%$) produce
no parseable matching, and $7$ produce a well-formed matching that
fails super-stability ($4$ are not even weakly stable, $3$ are
strongly stable but not super-stable). The model continues to
invoke Irving--Manlove by name, but the edge-deletion machinery is
rarely executed in detail: the trajectory typically names the
procedure and asserts its outcome.

\paragraph{Infeasible instances.}
On infeasible super-stable instances, GPT-5.2 produces either a
forced-pair impossibility chain or an Irving-style certificate in
which it runs the edge-deletion procedure and reports that some
agent's list became empty. A typical Irving-style certificate runs:

\begin{quote}\itshape
The reductions force $A_1 - B_9$ (mutual top), $A_9$ must be with
$B_1$ (deletions eliminate all other feasible partners), and
$B_{10}$ ends up only feasibly matchable with $A_2$. But then the
pair $(A_9, B_{10})$ is unavoidable as a blocking pair: $A_9$ is
indifferent between $B_1$ and $B_{10}$, so $A_9$ weakly prefers
$B_{10}$ to $B_1$; $B_{10}$ is indifferent between $A_9$ and
$A_2$, so $B_{10}$ weakly prefers $A_9$ to $A_2$. Therefore the
instance admits no super stable matching.
\end{quote}

\noindent
As on the other infeasible cells, the rate is determined by the
prompt variant. Under the empty-hint prompt: $20/30$ ($67\%$) at
Small and $29/30$ ($97\%$) at Modest, all $49$ correct. Under the
no-hint prompt: $0/30$ at each size. This is the only cell on
which a reasoning behaviour improves with size: the larger
instance carries more structural constraints that make the
forced-pair chain easier to spot, and the empty-hint prompt gives
the model permission to follow it to the impossibility conclusion.

\paragraph{Discussion.}
A consistent picture emerges. On feasible cells, GPT-5.2 succeeds
at Small size by deploying non-mechanical heuristics that exploit
instance-specific structure: the constructive chain heuristic for
strict core, deferred acceptance with forced-pair pre-processing
for strongly stable, and Irving--Manlove edge-deletion for super
stable. As the instance scales from $n=10$ to $n=30$ each of these
heuristics degrades, and the rate of no-parseable-answer outputs
rises sharply. The model does not have a fallback strategy that
degrades gracefully: it has only the same heuristics applied with
less fidelity, plus an increasing tendency to abandon the
question.

On infeasible cells, the qualitative finding is more
prompt-dependent than capability-dependent. GPT-5.2 produces
correct impossibility arguments on a substantial fraction of
infeasible instances under the empty-hint prompt ($80\%$ on
strict-core Small, $63\%/57\%$ on strongly-stable, $67\%/97\%$ on
super-stable), but \emph{zero} non-existence claims under the
no-hint prompt across all $180$ infeasible-cell trajectories issued under that variant. The
forced-pair impossibility chain is in the model's repertoire and,
when prompted to consider non-existence, scales with size on the
matching cells (whereas the strict-core chain collapses at Modest
because the chains are longer and the coalitions larger). The
practical implication is that infeasibility-handling rates on
benchmarks like ours are largely a function of prompt phrasing
rather than of model capability, and care must be taken not to
conflate the two.

\subsection{LLM-judge setup}
\label{app:reasoning:judge}

The LLM-judge stage extracts structured features from each
scratchpad without itself doing any algorithmic verification. Its
output is then joined to the deterministic verifier's stratum
labels for the per-cell tables above. We use DeepSeek~V4~Pro as
the judge with temperature~$0$ and JSON-mode response formatting.
Each judge call sees: (i) the original problem statement that the
model was given, (ii) the model's scratchpad, (iii) the model's
parsed final answer, (iv) a fixed list of per-cell open probes
asking for verbatim quotes of specific trace features, and
(v) a canonical-strategy check that asks whether the trace
follows a notion-specific reference strategy (yes / partial / no).
The probe fields and the canonical-strategy descriptions were
defined in advance from a preliminary inspection of $\sim$50
scratchpads and were held fixed during the full $720$-trajectory
run. The judge is instructed to produce verbatim quotes wherever
feasible, to avoid inventing content that is not in the
scratchpad, and to return \texttt{null} or \texttt{[]} for fields
with no relevant content.

The full prompt template, with placeholders shown in
\texttt{\{}\textit{italic}\texttt{\}}, is:

\begin{quote}\small
\textbf{System.} You are an expert annotator analyzing reasoning
traces from a language model solving combinatorial matching and
allocation problems (housing markets / stable matching with ties).
For each trace you will answer a set of open-ended probe questions
about objective features of the trace, and then a
canonical-strategy check. Always output a single JSON object and
nothing else. Use verbatim quotes from the scratchpad where
requested. Do not invent content not present in the scratchpad.

\medskip
\textbf{User.}\\
PROBLEM CONTEXT (the prompt the model was given):\\
\texttt{"""}\,\textit{prompt\_text}\,\texttt{"""}

\smallskip
MODEL'S REASONING TRACE (scratchpad):\\
\texttt{"""}\,\textit{scratchpad}\,\texttt{"""}

\smallskip
MODEL'S FINAL ANSWER (parsed from \texttt{<answer>} tag):\\
\texttt{"""}\,\textit{answer}\,\texttt{"""}

\smallskip
PROBE QUESTIONS:\\
(1) FIELD \texttt{algorithm\_name}: Verbatim name(s) of the
algorithm the model invokes (e.g. ``Top Trading Cycles'',
``Gale--Shapley''). If none stated, write null. Maximum 8 words.\\
(2) FIELD \texttt{cycles\_shown}: List of explicit cycles or
rounds shown in the scratchpad, each as a chain string of the form
``$A_i \to C_j \to A_k \to \cdots \to A_i$''. If none, write \texttt{[]}.\\
(3) FIELD \texttt{tie\_break\_rule}: Verbatim quote of the
sentence(s) describing how ties within indifference classes are
broken. If unspecified, null.\\
(4) FIELD \texttt{verification\_text}: Verbatim quote of the
sentence(s) where the model verifies the result beyond a generic
property-citation. If only a generic property-citation is present,
write the exact string \texttt{PROPERTY\_CITATION\_ONLY}. If none,
null.\\
(5) FIELD \texttt{forced\_assignments\_pre\_algorithm}: List of
any ``agent must be assigned house'' deductions stated before the
main algorithm is run, each with a quoted reason. If none, \texttt{[]}.\\
(6) FIELD \texttt{final\_claim}: One of \texttt{allocation\_produced},
\texttt{claims\_no\_solution}, \texttt{abstains\_or\_refuses}.
Quote the closing sentence of the scratchpad.

\smallskip
CANONICAL STRATEGY CHECK. Indicate whether the trajectory follows
the canonical strategy below: ``yes'' (clearly and substantively),
``partial'' (some elements but deviates in others), or ``no''.
Quote 1--2 verbatim sentences as evidence. If partial or no,
briefly note in 1 sentence what differs.\\
Canonical strategy: \textit{strategy\_name}.\\
Description: \textit{strategy\_description}.

\smallskip
OUTPUT JSON OBJECT WITH EXACTLY THESE FIELDS:\\
\texttt{\{}\\
\hphantom{xx}\texttt{"algorithm\_name": <answer>,}\\
\hphantom{xx}\texttt{"cycles\_shown": <answer>,}\\
\hphantom{xx}\texttt{"tie\_break\_rule": <answer>,}\\
\hphantom{xx}\texttt{"verification\_text": <answer>,}\\
\hphantom{xx}\texttt{"forced\_assignments\_pre\_algorithm": <answer>,}\\
\hphantom{xx}\texttt{"final\_claim": <answer>,}\\
\hphantom{xx}\texttt{"canonical\_strategy\_match": "yes" | "partial" | "no",}\\
\hphantom{xx}\texttt{"canonical\_strategy\_evidence": "<verbatim quote(s)>",}\\
\hphantom{xx}\texttt{"canonical\_strategy\_deviation": "<one sentence or null>"}\\
\texttt{\}}
\end{quote}

The probe questions vary slightly across the six
$(\textit{notion},\textit{feasibility})$ cells (for instance,
the matching cells substitute ``forced pairs'' for ``trading
cycles'' in probe~(2)), but the system prompt, the
output-schema requirements, and the canonical-strategy check
template are shared across all cells. The complete cell-specific
probe definitions are released alongside the code.

\section{Prompt-level mitigation}
\label{app:prompt_mitigation}

If the epistemic failure were a surface prompting artifact, an instruction warning the model about indeterminacy should remove it. We test this with four levels of prompt mitigation on the bundle comparison problem, run on both RS-comparable and RS-incomparable cases. A genuine fix has to keep accuracy high on both, since a deployed user does not know in advance which kind of query they are issuing.

\begin{itemize}
    \item Level 1 bare, no hint of indeterminacy.
    \item Level 2 multiple choice with an explicit "not enough information" option.
    \item Level 3 caution. general = source-agnostic ("preferences may not entail an answer, do not assume one"). specific = names the structure being tested.
    \item Level 4 few-shot demonstration ending in "cannot be determined"
\end{itemize}

Table~\ref{tab:prompt_mitigation} shows that no prompt calibrates the model. Every intervention that raises undetermined accuracy also lowers accuracy on the determined queries: the general caution moves GPT-5.2 from $1.00 \mid 0.00$ to $0.43 \mid 0.80$ and Gemini-2.5-P from $1.00 \mid 0.00$ to $0.47 \mid 0.93$. The error is relocated, not removed. 

Of the prompts tried, no single prompt is safe across models. The specific caution lifts Claude and Gemini but collapses OSS-120B to $0.27 \mid 0.93$, and the explicit abstain option is ignored by three of four models, which keep imposing the lexicographic default.

Few-shot demonstrations raise undetermined accuracy sharply when they come from the same source of indeterminacy, but the gain does not transfer. When the demonstrations used for bundle comparison are applied to a different source (an item the agent never ranked), Table~\ref{tab:prompt_mitigation_cross} shows they help almost not at all. The model reproduces the pattern it was shown rather than acquiring a general ability to flag indeterminacy.

This is why the failure is not a surface prompting artifact. Prompting moves the model along a trade-off between over-commitment and over-abstention, and a caution aimed at one kind of indeterminacy does not carry over to another. This matters in practice, because a deployed user does not know in advance whether a query is undetermined, or which kind of indeterminacy it involves, so they cannot supply the matching caution beforehand. The cross-setting experiment tests this. When the demonstrations describe a different source than the query, performance falls back to the baseline, showing the model has not learned a general ability to recognize indeterminacy but only to imitate the case it was shown.

\begin{table}[h]
\centering
\caption{Accuracy on bundle comparison problems, with different levels of prompt mitigation. Levels: 1 bare, 2 MCQ with an explicit "not enough information" option, 3 caution (general = source-agnostic, specific = names the structure), 4 few-shot demos. Each is run on 30 instances, with accuracy reported as determined $\mid$ undetermined}
\label{tab:prompt_mitigation}
\small
\setlength{\tabcolsep}{4pt}
\begin{tabular}{l ccccc}
\toprule
\textbf{Model} & \textbf{L1 bare} & \textbf{L2 MCQ} & \textbf{L3 general} & \textbf{L3 specific} & \textbf{L4 few-shot} \\
\midrule
GPT-5.2      & $1.00 \mid 0.00$ & $0.97 \mid 0.00$ & $0.43 \mid 0.80$ & $0.73 \mid 0.27$ & $0.93 \mid 0.10$ \\
Claude-4.5-S & $1.00 \mid 0.00$ & $1.00 \mid 0.00$ & $0.97 \mid 0.57$ & $1.00 \mid 0.83$ & $1.00 \mid 0.97$ \\
Gemini-2.5-P & $1.00 \mid 0.00$ & $1.00 \mid 0.00$ & $0.47 \mid 0.93$ & $0.97 \mid 0.50$ & $1.00 \mid 0.90$ \\
OSS-120B     & $1.00 \mid 0.00$ & $0.83 \mid 0.37$ & $0.73 \mid 0.63$ & $0.27 \mid 0.93$ & $0.97 \mid 0.83$ \\
\bottomrule
\end{tabular}
\end{table}

\begin{table}[h]
\centering
\caption{Accuracy on undetermined atomic query, with different levels of prompt mitigation. Levels: L1 bare, L4 few-shot demos (in-domain = examples about the tested source of indeterminacy. cross-source = examples about the other source). Each is run on 30 instances, with accuracy reported.}
\label{tab:prompt_mitigation_cross}
\small
\setlength{\tabcolsep}{4pt}
\begin{tabular}{l ccc}
\toprule
\textbf{Model} & \textbf{L1} & \textbf{L4 in-domain} & \textbf{L4 cross-domain} \\
\midrule
GPT-5.2      & 0.00 & 0.10 & 0.13 \\
Claude-4.5-S & 0.00 & 0.97 & 0.00 \\
Gemini-2.5-P & 0.00 & 0.90 & 0.00 \\
OSS-120B     & 0.00 & 0.83 & 0.00 \\
\bottomrule
\end{tabular}
\end{table}

\section{Code-Assisted Reasoning: Approaches by Model and Size}
\label{app:code_approaches}

\begin{table}[t]
\centering
\caption{Code-based task accuracy by domain, problem size, and feasibility (F = feasible and I = infeasible).
  Each cell shows the percentage of correct responses out of 10 instances. 
  Dashes indicate conditions not evaluated. Timeouts and invalid responses are counted as incorrect.}
\label{tab:code-accuracy}
\small
\setlength{\tabcolsep}{4pt}
\resizebox{0.8\textwidth}{!}{
\begin{tabular}{@{}l cccccc cccccc@{}}
\toprule
& \multicolumn{6}{c}{\textbf{Shapley-Scarf Market}}
& \multicolumn{6}{c}{\textbf{Matching Market}} \\
\cmidrule(lr){2-7}\cmidrule(lr){8-13}
& \multicolumn{2}{c}{Small} & \multicolumn{2}{c}{Modest} & \multicolumn{2}{c}{Medium}
& \multicolumn{2}{c}{Small} & \multicolumn{2}{c}{Modest} & \multicolumn{2}{c}{Medium} \\
\cmidrule(lr){2-3}\cmidrule(lr){4-5}\cmidrule(lr){6-7}
\cmidrule(lr){8-9}\cmidrule(lr){10-11}\cmidrule(lr){12-13}
\textbf{Model}
  & F & I & F & I & F & I
  & F & I & F & I & F & I \\
\midrule
Gemini-2.5
  & 10\% & 0\% & 10\% & 0\% & -- & --
  & 20\% & 70\% & 0\% & 60\% & -- & -- \\
Claude-4.5-S
  & 100\% & 100\% & 0\% & 70\% & -- & --
  & 90\% & 100\% & 30\% & 100\% & -- & -- \\
GPT-5.2
  & 100\% & 100\% & 90\% & 70\% & 100\% & 30\%
  & 100\% & 100\% & 90\% & 90\% & 0\% & 10\%
\\
\bottomrule
\end{tabular}}
\end{table}

The accuracy achieved by LLMs on algorithmic reasoning questions, when using code to solve the problem, are provided in \Cref{tab:code-accuracy}. 

This section summarises the dominant strategy each model produces
in the code-assisted condition, broken down by notion and instance
size. Three patterns recur. First, at $n=10$, GPT-5.2 and Claude
Sonnet~4.5 follow a common verify-then-fall-back pattern: the
generated code first attempts the textbook algorithm for the
notion (Top Trading Cycles, deferred acceptance, or Irving's
super-stable algorithm), checks whether the resulting object
satisfies the notion's condition, and on a failed check falls back
to exhaustive enumeration over all candidate allocations or
matchings combined with a complete blocking-coalition or
blocking-pair check. The fallback is sound at $n=10$ because the
search space is small enough to enumerate. Second, at $n=30$ and
above the same models switch to heuristic strategies (randomised
tie-breaking with verification, backtracking with constraint
propagation, exponential-weight reductions to maximum-weight
matching), and the soundness of the fallback degrades as size
grows. Third, Gemini~2.5~Pro is the outlier: it does not use
exhaustive enumeration at any size, and its code typically applies
a textbook algorithm directly with no verification step.

\subsection{GPT-5.2}

GPT-5.2's code shows the cleanest version of the
verify-then-fall-back pattern. On strict core with ties at $n=10$,
it first runs Top Trading Cycles (TTC) with first-in-list
tie-breaking, verifies the output against a complete
blocking-coalition oracle, and falls back to exhaustive enumeration
of all $n!$ allocations on a verification failure, returning $\{\}$
when the enumeration finds no allocation in the strict core. At
$n=30$ the model switches to randomised TTC: it repeats TTC with
random tie-breaking until verification succeeds. This is sound on
feasible instances and remains correct on all $n=50$ feasible
instances we evaluated, but it cannot prove non-existence, so on
infeasible $n=30$ instances the model occasionally falls back to a
MILP encoding of the strict-core polytope, with mixed soundness. At
$n=50$ infeasible instances the randomised loop times out before
any code completes. On super-stable matching, GPT-5.2 progresses
from exhaustive enumeration at $n=10$, to backtracking with
constraint propagation at $n=30$, to a faulty implementation of
Irving's algorithm \textsc{Super} at $n=50$ that omits the step of
breaking all engagements of multiply-engaged agents, and
consequently produces false NE claims on every $n=50$ instance.

\begin{table}[h]
\centering
\small
\caption{GPT-5.2: dominant approach in the code-assisted condition,
by task and instance size. For notions with feasibility variation,
E and NE rows are shown separately where approaches differ.}
\label{tab:app_gpt}
\setlength{\tabcolsep}{4pt}
\begin{tabular}{@{}llp{1.2cm}p{9cm}@{}}
\toprule
\textbf{Notion} & \textbf{$n$} & \textbf{E/NE} & \textbf{Dominant approach} \\
\midrule
Core (HA, strict)
  & 10--50 & --
  & Top Trading Cycles (correct) \\
\midrule
\multirow{5}{*}{\makecell[l]{Strict Core\\(HA, ties)}}
  & 10 & E/NE & TTC attempted; on verification failure, falls back
                to exhaustive enumeration with complete
                blocking-coalition check (sound) \\
  & 30 & E   & Randomised TTC: repeat with random tie-breaking
                until a strict-core allocation is found and verified \\
  & 30 & NE  & Heuristic search failure; MILP in some responses;
                not always sound \\
  & 50 & E   & Randomised TTC; all instances correct \\
  & 50 & NE  & Timeout (no result produced) \\
\midrule
\multirow{2}{*}{\makecell[l]{Rank-Max.\ Alloc.\\(HA, ties)}}
  & 10 & -- & Brute-force enumeration over all allocations \\
  & 30--50 & -- & Exponential-weight reduction to max-weight
                  matching (one-shot; empirically correct) \\
\midrule
\multirow{4}{*}{\makecell[l]{Super Stable\\(SM, ties)}}
  & 10 & E/NE & Gale-Shapley variant attempted; on verification
                failure, falls back to exhaustive enumeration with
                complete super-blocking-pair check (sound) \\
  & 30 & E/NE & Backtracking with constraint propagation (CSP);
                mostly correct \\
  & 50 & E/NE & Algorithm \textsc{Super} attempted but incorrectly
                implemented: omits breaking all engagements of
                multiply-engaged agents, causing false NE claims on
                all instances \\
\midrule
\multirow{1}{*}{\makecell[l]{Rank-Max.\ Match.\\(SM, ties)}}
  & 10--30 & -- & Exponential-weight reduction (one-sided $A$-side
                  only); misunderstands two-sided definition \\
\bottomrule
\end{tabular}
\end{table}

\subsection{Claude Sonnet 4.5}

Claude Sonnet~4.5 follows the same overall verify-then-fall-back
pattern at $n=10$ as GPT-5.2 (exhaustive enumeration of allocations
or matchings, with complete blocking checks), but its heuristic
strategies at $n=30$ are more limited. On strict core with ties at
$n=30$, the generated code searches a random sample of allocations
and verifies them against blocking coalitions of size $\leq 5$
only. This yields low accuracy on feasible instances but
accidentally high accuracy on infeasible ones, since a small
blocking coalition is usually found quickly. On super stable at
$n=30$, the code runs Gale-Shapley with random tie-breaking and
post-hoc super-stability verification, which gives partial accuracy
on feasible instances and again finds an obstruction quickly when
none exists. On rank-maximal matching, Claude's code (like
GPT-5.2's) operates on a one-sided ranking only, indicating that
both models misread the two-sided definition.

\begin{table}[h]
\centering
\small
\caption{Claude-4.5-Sonnet: dominant approach in the code-assisted
condition, by task and instance size.}
\label{tab:app_claude}
\setlength{\tabcolsep}{4pt}
\begin{tabular}{@{}llp{1.2cm}p{9cm}@{}}
\toprule
\textbf{Notion} & \textbf{$n$} & \textbf{E/NE} & \textbf{Dominant approach} \\
\midrule
Core (HA, strict)
  & 10--30 & --
  & Top Trading Cycles (correct; minor cycle-detection bugs fixed
    in later iterations) \\
\midrule
\multirow{4}{*}{\makecell[l]{Strict Core\\(HA, ties)}}
  & 10 & E/NE & TTC attempted; on verification failure, falls back
                to exhaustive enumeration with full
                blocking-coalition verification (sound) \\
  & 30 & E   & Heuristic random allocation search with limited
                coalition-size checks ($\leq 5$); low accuracy \\
  & 30 & NE  & Heuristic: blocking coalition found quickly; mostly
                correct \\
\midrule
\multirow{2}{*}{\makecell[l]{Rank-Max.\ Alloc.\\(HA, ties)}}
  & 10 & -- & Mixed: some exhaustive enumeration, some greedy
              rank-by-rank \\
  & 30 & -- & Greedy rank-by-rank fixation (incorrect; $0\%$
              accuracy) \\
\midrule
\multirow{3}{*}{\makecell[l]{Super Stable\\(SM, ties)}}
  & 10 & E/NE & Gale-Shapley variant attempted; on verification
                failure, falls back to exhaustive enumeration of
                all matchings with super-blocking-pair check (sound) \\
  & 30 & E   & Random Gale-Shapley tie-breaking with post-hoc
                verification (heuristic; partial accuracy) \\
  & 30 & NE  & Heuristic failure after many random trials; often
                correct \\
\midrule
\multirow{1}{*}{\makecell[l]{Rank-Max.\ Match.\\(SM, ties)}}
  & 10--30 & -- & Iterative rank addition with one-sided matching;
                  misunderstands two-sided definition \\
\bottomrule
\end{tabular}
\end{table}

\subsection{Gemini 2.5 Pro}

Gemini~2.5~Pro does not use exhaustive enumeration at any size,
and its code typically lacks an explicit verification step. On
strict core with ties, it runs TTC with an arbitrary tie-breaking
rule and treats the output as the answer regardless of whether it
lies in the strict core, which means it never identifies
infeasible instances. On super stable, the code is an ad-hoc
modification of Gale-Shapley with no super-stability check, and at
$n=30$ Gemini has a high rate of false NE claims regardless of
ground-truth feasibility. On rank-maximal matching at $n=30$,
Gemini's code is a greedy per-rank algorithm that achieves $0\%$
accuracy. The textbook algorithm for the simplest notion in this
study (Core in strict-preference housing markets) is correctly
implemented, with two cycle-detection bugs at $n=30$.

\begin{table}[h]
\centering
\small
\caption{Gemini-2.5: dominant approach in the code-assisted
condition, by task and instance size.}
\label{tab:app_gemini}
\setlength{\tabcolsep}{4pt}
\begin{tabular}{@{}llp{1.2cm}p{9cm}@{}}
\toprule
\textbf{Notion} & \textbf{$n$} & \textbf{E/NE} & \textbf{Dominant approach} \\
\midrule
Core (HA, strict)
  & 10--30 & --
  & Top Trading Cycles (correct; two failures at $n=30$ due to
    cycle-detection bugs) \\
\midrule
\multirow{1}{*}{\makecell[l]{Strict Core\\(HA, ties)}}
  & 10--30 & -- & TTC with arbitrary tie-breaking and no
                  verification step (incorrect; never identifies
                  infeasible instances) \\
\midrule
\multirow{1}{*}{\makecell[l]{Rank-Max.\ Alloc.\\(HA, ties)}}
  & 10--30 & -- & Greedy sequential maximum matching per rank
                  (incorrect heuristic; near-zero accuracy at
                  $n=30$) \\
\midrule
\multirow{3}{*}{\makecell[l]{Super Stable\\(SM, ties)}}
  & 10 & E   & Custom iterative elimination or modified
                Gale-Shapley variants \\
  & 10 & NE  & Heuristic failure or custom elimination; partially
                correct \\
  & 30 & E/NE & Ad-hoc Gale-Shapley modifications; high rate of
                false NE claims regardless of ground truth \\
\midrule
\multirow{2}{*}{\makecell[l]{Rank-Max.\ Match.\\(SM, ties)}}
  & 10 & -- & Exponential-weight Hungarian (one-sided;
              misunderstands two-sided definition) \\
  & 30 & -- & Greedy iterative per-rank matching (incorrect
              heuristic; $0\%$ accuracy) \\
\bottomrule
\end{tabular}
\end{table}

\section{Refinement Experiment Details}
\label{app:refinement}

\paragraph{Setup.}
The refinement experiment tests whether LLMs can correct their answers when given verification-based feedback. We focus on two solution notions that require generating valid allocations or matchings:

\begin{itemize}
    \item \textbf{Strict core} (Shapley--Scarf housing market): agents own one house each and can trade. A valid strict core allocation has no group of agents (coalition) that could redistribute their endowed houses among themselves so that every member weakly improves and at least one strictly improves.
    \item \textbf{Super stable matching} (bipartite matching market): agents on two sides are matched one-to-one. A super stable matching has no blocking pair---a pair of agents who both weakly prefer each other to their current partners.
\end{itemize}

For each notion, we include 30 feasible instances (a valid solution exists) and 30 infeasible instances (no valid solution exists), for a total of 60 instances per model per notion. We evaluate two models: OSS-120B on Small instances ($n = 10$) and GPT-5.2 on Modest instances ($n = 30$). All instances use the Generation task format, where the model must produce a JSON allocation or matching (or \texttt{\{\}} if no solution exists).

\paragraph{Interaction protocol.}
Each instance is evaluated using a multi-turn conversation with up to three attempts:

\begin{enumerate}
    \item \textbf{Attempt~1}: The model receives the original task prompt and produces an answer.
    \item \textbf{Attempt~2} (if Attempt~1 is wrong): The model receives feedback explaining why its answer is incorrect, then produces a new answer. The full conversation history is preserved.
    \item \textbf{Attempt~3} (if Attempt~2 is also wrong): The model receives another round of feedback and makes a final attempt, again with full history.
\end{enumerate}

If any attempt produces a correct answer, the instance is marked as a success and no further attempts are made. If any attempt times out or produces an error, the remaining attempts are skipped.

\paragraph{Feedback design.}
The feedback message is constructed automatically by running the same programmatic verifier used to score the initial experiments. The content of the feedback depends on what the model got wrong.

\medskip
\noindent\textbf{When the model returns \texttt{\{\}} on a feasible instance} (incorrectly claiming no solution exists):

\begin{quote}
\small\itshape
``Your answer is incorrect. A valid [notion] solution does exist for this instance, but you returned an empty allocation. You have [N] attempts remaining. Please try again carefully and return your answer enclosed in \texttt{<answer>} \texttt{</answer>} tags.''
\end{quote}

This tells the model that a solution exists without revealing what it is.

\medskip
\noindent\textbf{When the model returns a wrong allocation for strict core:}

The verifier computes all blocking coalitions and reports them. For example:

\begin{quote}
\small\itshape
``Your answer is incorrect. Your allocation has 2 blocking coalitions under the strict core. Here are the blocking coalitions: \{A2, A10, A5, A8, A3, A9, A7\}; \{A2, A10, A5, A9, A7\}. Each coalition can reshuffle their endowed items among themselves so that every member weakly improves and at least one strictly improves. You have [N] attempts remaining.''
\end{quote}

The feedback includes:
\begin{itemize}
    \item The total number of blocking coalitions.
    \item The members of each coalition (up to 3 coalitions are shown; if there are more, a note indicates the total).
    \item A plain-language explanation of what a blocking coalition means.
\end{itemize}

\medskip
\noindent\textbf{When the model returns a wrong matching for super stable:}

The verifier computes all blocking pairs and reports them. For example:

\begin{quote}
\small\itshape
``Your answer is incorrect. Your matching has 5 blocking pairs under super stability. Here are the blocking pairs: (A1, B10), (A3, B10), (A4, B10), (A4, B9), (A8, B9). In each pair, both agents weakly prefer each other to their current partners. You have [N] attempts remaining.''
\end{quote}

The feedback includes:
\begin{itemize}
    \item The total number of blocking pairs.
    \item The specific pairs (up to 5 are shown; if there are more, a note indicates the total).
    \item A plain-language explanation of the blocking condition.
\end{itemize}

\medskip
\noindent\textbf{When the model's response cannot be parsed as valid JSON:}

\begin{quote}
\small\itshape
``Your answer could not be parsed as a valid JSON allocation. Make sure you return a JSON object mapping each agent to an assigned alternative, enclosed in \texttt{<answer>} \texttt{</answer>} tags. You have [N] attempts remaining.''
\end{quote}

\paragraph{Key design choices.}

\begin{itemize}
    \item \textbf{Conversation history is preserved.} Each attempt is sent as a new message in the same multi-turn conversation. The model sees its own prior responses and the feedback it received, without needing to re-read the original problem description.

    \item \textbf{Feedback is informative but not prescriptive.} The feedback identifies specific violations (which coalitions block, which pairs block) but does not suggest how to fix the answer. The model must figure out the repair strategy on its own.

    \item \textbf{Verification is programmatic.} Correctness is determined by the same algorithmic verifiers used in the main experiments---not by string matching or LLM-based judging. For strict core, the verifier exhaustively checks all possible coalitions. For super stability, it checks all agent pairs for the weak blocking condition.

    \item \textbf{Feasibility feedback is asymmetric.} When an instance is feasible but the model returns \texttt{\{\}}, the feedback reveals that a solution exists. When an instance is infeasible and the model returns a wrong allocation, the feedback shows the blocking coalitions or pairs---which may indirectly signal that no valid solution exists (e.g., seeing hundreds of blocking coalitions suggests infeasibility). However, the feedback never explicitly states that no solution exists.
\end{itemize}

\section{LLM-Judge Analysis Details}
\label{app:llm-judge}

We use a separate LLM-as-judge pass to label the reasoning portion of each
response, i.e., which algorithm or solution concept the responding model
appears to have invoked, independently of whether the final allocation is
correct. We run this in two settings. The \textbf{generation} setting classifies
the algorithm a model used to produce a solution when no specific target notion is provided. The \textbf{selection} setting classifies the criterion a model
prioritized when picking among several candidate solutions. The two settings
share the same judge model, prompt scaffolding, and answer format; they differ
in the option set and in how ``intended'' and ``achieved'' are scored.

\textbf{Judge configuration.}
All judging is done with Gemini-2.5-Flash through the \texttt{google-genai}
Python client, using the client's default sampling parameters (no temperature,
top-$p$, or other generation arguments are set). The judge is shown two pieces
of context: the original prompt that was given to the model under study, and
that model's full response. It is asked to select a single option from a closed
list and return its choice in the format
\verb|<answer>X</answer>|. The judge is explicitly instructed \emph{not} to verify
the algorithm or criterion by computing solutions itself; its label must be
based only on explicit textual mentions or descriptions in the response. If no
listed option matches, it must return the catch-all ``no algorithm
mentioned''/``some other notion'' option. Each judge call is wrapped in a
three-attempt retry to absorb transient API failures, and the parsed letter is
post-processed by stripping the \verb|<answer>...</answer>| tags; we additionally
fall back to extracting the letter inside \verb|\boxed{...}| when the judge
returns its answer in LaTeX form.

\textbf{Coverage and aggregation.}
Both settings sweep over the cross-product of model, instance size
$\in \{10, 30\}$, domain $\in \{$House Allocation, Shapley-Scarf Markets, Matching Markets$\}$, and preference type $\in \{$strict and complete, strict and
incomplete, with ties and complete$\}$, restricted to the relevant subset of
each model's responses (the underspecified-instance subset for the generation
setting and the selection-task subset for the selection setting).

\subsection{Generation Setting}
\label{app:judge-generation}

In the generation setting, the judge classifies the algorithm the model used to
produce a single solution to an underspecified instance. The option set is
restricted to algorithmic procedures: Top Trading Cycles, the Hungarian
min-cost algorithm, Hopcroft--Karp matching, exhaustive search, Gale--Shapley
deferred acceptance, Irving's strong-stable and super-stable matching
algorithms, serial dictatorship, ``some other algorithm'', and ``no algorithm
mentioned/described''. A response is counted as having \emph{intended} the
canonical algorithm for its domain when the judge's choice matches a fixed
mapping: house allocation $\leftrightarrow$ Top Trading Cycles, object
allocation $\leftrightarrow$ serial dictatorship, and stable matching
$\leftrightarrow$ Gale--Shapley. The \emph{achieved} count is computed
independently from the model's allocation: for instances with complete
preferences, we use the binary correctness label produced by our standard
verifier; for instances with incomplete preferences, we recompute the core
allocation and compare it against the model's allocation, restricted to the
agents that the reference solution actually matches.

A representative judge prompt for this setting is shown in
\Cref{fig:judge-prompt-generation}. The placeholders
\textsc{\textlangle original prompt\textrangle} and
\textsc{\textlangle model response\textrangle} are filled in from the row being
judged.

\begin{figure}[t]
\centering
\begin{tcolorbox}[colback=gray!5, colframe=black!40, boxrule=0.4pt,
                  left=4pt, right=4pt, top=4pt, bottom=4pt]
\small
Previously I asked another LLM the following question:\\[2pt]
``\textsc{\textlangle original prompt\textrangle}''\\[4pt]
Here is its response:\\[2pt]
\verb|"""|\\
\textsc{\textlangle model response\textrangle}\\
\verb|"""|\\[4pt]
Your task is to identify the approach or algorithm that the model has used to
answer the question. Among the following options, select the most appropriate
option for the model's response:\\[2pt]
A: Top Trading Cycles (TTC)\\
B: Hungarian Min-cost Algorithm\\
C: Hopcroft--Karp Maximum Matching Algorithm\\
D: Searching through all possibilities\\
E: Deferred Acceptance (Gale--Shapley)\\
F: Strong stable matching algorithm (Irving)\\
G: Super stable matching algorithm (Irving)\\
H: Serial Dictatorship\\
I: Some other algorithm\\
J: No algorithm mentioned/described\\[4pt]
DO NOT try to verify the algorithm that the LLM has used by computing solutions
yourself. Only rely on explicit mentions or descriptions of the algorithms in
options A--I. If no algorithm is explicitly mentioned or described, select
option~J. Return your answer in the following format:
\verb|<answer>option</answer>|. For example, if the correct option were
option~X, you would have to return \verb|<answer>X</answer>|.
\end{tcolorbox}
\caption{Judge prompt used in the generation setting.}
\label{fig:judge-prompt-generation}
\end{figure}

\subsection{Selection Setting}
\label{app:judge-selection}

In the selection setting, the responding model is presented with a set of
candidate solutions to an instance, each pre-labeled with the solution concept
it satisfies, and is asked to choose one. The judge then classifies which
solution concept or criterion the model prioritized in its written
justification. The option set covers eleven concepts: (weak) core, utilitarian
welfare, egalitarian welfare, rank-maximality, strict core, (weak) stability,
strong stability, super stability, Pareto-optimality, maximum-cardinality
Pareto-optimal matching, and a catch-all ``some other notion or criterion''.
When a response references multiple concepts, the judge is instructed to
select the one the model treats as the first priority.

For each row we record three quantities. \emph{Intended} is the concept
identified by the judge from the model's reasoning. \emph{Actual} is the
concept satisfied by the candidate the model ultimately picked, recovered by
mapping each candidate's pre-assigned label onto the same eleven-option set.
\emph{Achieved} is an indicator equal to one if and only if the intended and
actual concepts agree---i.e., the model's chosen solution in fact satisfies
the concept its reasoning claims to prioritize. The mapping from candidate
labels to option letters is fixed and deterministic; in particular, candidates
labeled either ``Core'' (under strict preferences) or ``Weak core'' (under
preferences with ties) are both mapped to the (weak) core option.

A representative judge prompt for this setting is shown in
\Cref{fig:judge-prompt-selection}.

\begin{figure}[t]
\centering
\begin{tcolorbox}[colback=gray!5, colframe=black!40, boxrule=0.4pt,
                  left=4pt, right=4pt, top=4pt, bottom=4pt]
\small
Previously I asked another LLM the following question:\\[2pt]
``\textsc{\textlangle original prompt\textrangle}''\\[4pt]
Here is its response:\\[2pt]
\verb|"""|\\
\textsc{\textlangle model response\textrangle}\\
\verb|"""|\\[4pt]
Your task is to identify the solution concept or criterion that the model
prioritizes while answering the question. Among the following options, select
the most appropriate option for the model's response:\\[2pt]
A: Core: the solution is in the core (if preferences are strict), or in the
   weak core (if preferences have ties).\\
B: Utilitarian welfare: the solution minimizes the sum of ranks of matched
   agents compared to all other options.\\
C: Egalitarian welfare: the solution has the smallest ``worst'' rank among all
   options.\\
D: Rank-maximality: the solution maximizes the number of agents matched to
   their first preference; conditional on that, it maximizes the number
   matched to their second preference, and so on.\\
E: Strict core: there is no coalition of agents who can re-assign their
   endowments such that every agent in the coalition is weakly better off and
   at least one agent is strictly better off.\\
F: (Weak) stability: the solution has no weakly blocking pair, i.e., no pair of
   unmatched agents that strictly prefer each other to their matched
   partners.\\
G: Strong stability: the solution has no strongly blocking pair, i.e., no pair
   of unmatched agents in which one agent weakly prefers the other to its
   current match and the other strictly prefers the first to its current
   match.\\
H: Super stability: the solution has no super blocking pair, i.e., no pair of
   unmatched agents that weakly prefer each other to their matched partners.\\
I: Pareto-optimality: there is no other solution in which some agent strictly
   improves without making another agent worse off.\\
J: Maximum-cardinality Pareto-optimal: the solution maximizes the number of
   agents matched to a ranked alternative and is Pareto-optimal.\\
K: Some other notion or criterion.\\[4pt]
DO NOT try to verify the property that the model claims to prioritize by
computing solutions yourself. Only rely on explicit textual mentions or
descriptions of the criteria in options A--J. If multiple criteria are
mentioned, use your discretion to identify which one is given the first
priority by the model. If no criterion is explicitly mentioned or described,
select option~K. Return your answer in the following format:
\verb|<answer>option</answer>|. For example, if the correct option were
option~X, you would have to return \verb|<answer>X</answer>|.
\end{tcolorbox}
\caption{Judge prompt used in the selection setting.}
\label{fig:judge-prompt-selection}
\end{figure}

\subsection{Additional judges and human validation}
\label{app:judge_validation}
Two judge pipelines are used in this paper. The generation-side judge identifies the algorithm or concept a model targets when it computes a solution, and supports both \Cref{app:gen_a_sol} and \Cref{app:reasoning-strategies}. The selection-side judge supports the intention-action analysis behind \Cref{fig:selection_stacked}. The lexicographic-completion and premature-termination findings of \Cref{fig:pref_assumptions} are detected by a deterministic script and use no judge.

\textbf{Independent judges.} Each pipeline is rerun with two additional models (GPT-5.6-Luna and Gemini-3.5-Flash) on a stratified subset. Three-judge Fleiss' $\kappa$ is 0.64 for the reasoning-strategy judge and 0.82 for the intention-action judge. The lower agreement on the reasoning-strategy judgements comes from the ``strongly stable'' cell, where the boundary is genuinely ambiguous because the model sometimes gives a valid argument that is not the textbook one, and the judges score these borderline cases inconsistently. Pooling hides where the disagreement sits, so we report this per notion: all three judges agree on 0.95 of the strict-core traces and 0.94 of the super-stable traces, against 0.43 for strongly stable.

\textbf{Human validation.} Human annotation is also used to validate the judge responses. For 36 responses, one author independently chose which concept the model prioritized, using the same options and instructions the judge saw. Against the paper's judge, Gemini-2.5-Flash, human-vs-judge Cohen's kappa is 0.86, in line with the inter-judge agreement. The alternate judges agree a little less, 0.72 and 0.76, mostly on the incomplete house-allocation cell, where the original judge and the human read an implicit assign-everyone justification as max-cardinality intent while the other two judges do not. The human therefore agrees most with the exact judge behind the reported numbers, and the one divergent cell is the one we flag as judge-sensitive in \Cref{app:intention_action} rather than pool with the rest.

\textbf{Feasibility oracle.} Correctness is decided by deterministic checker functions that verify every model output against the target solution concept. Because these checkers are our own implementation, we cross-checked them with an independent brute-force solver that enumerates all allocations and matchings for small instances and decides existence directly from each concept's definition. Across 1,800 random instances (strict core, super-stable, and strongly-stable, with n up to 5), the oracle's existence verdicts, the solutions it returns, and the blocking-pair verifier agree with the brute-force solver in every case. We release this validation harness alongside the code.

\end{document}